\documentclass[twoside,11pt]{article}

\usepackage{bm}
\usepackage{dsfont}
\usepackage{amsmath}
\usepackage{color}
\usepackage{graphicx}
\usepackage{url}
\usepackage{rotating}
\usepackage[nottoc]{tocbibind}
\usepackage{multirow}
\usepackage{subcaption}
\usepackage{float} 
\usepackage{lastpage}

\usepackage{jmlr2e}

\usepackage{hyperref}
\hypersetup{colorlinks=true}

\firstpageno{1}

\begin{document}

	\jmlrheading{21}{2020}{1-\pageref{LastPage}}{2/20; Revised
		11/20}{12/20}{20-107}{Carlos Villacampa-Calvo, Bryan Zald\'ivar, Eduardo C. Garrido-Merch\'an and Daniel Hern\'andez-Lobato}
	\ShortHeadings{Multi-class Gaussian Process Classification with Noisy Inputs}{Villacampa-Calvo, Zald\'ivar, Garrido-Merch\'an and Hern\'andez-Lobato}
	
	\title{Multi-class Gaussian Process Classification\\ with Noisy Inputs}

	\author{\name Carlos Villacampa-Calvo \email carlos.villacampa@uam.es \\
		\addr Computer Science Department, Universidad Aut\'onoma de Madrid, 28049, Madrid, Spain\\
		\name Bryan Zald\'ivar \email bryan.zaldivarm@uam.es \\
		\addr Theoretical Physics Department, Universidad Aut\'onoma de Madrid, 28049, Madrid, Spain\\
		\addr Instituto de F\'isica Te\'orica, 28049, Madrid, Spain \\
		\name Eduardo C. Garrido-Merch\'an\email eduardo.garrido@uam.es \\
		\addr Computer Science Department, Universidad Aut\'onoma de Madrid, 28049, Madrid, Spain\\
		\name Daniel Hern\'andez-Lobato \email daniel.hernandez@uam.es \\
		\addr Computer Science Department, Universidad Aut\'onoma de Madrid, 28049, Madrid, Spain\\
	}
	
	\editor{Philipp Hennig}
	
	\maketitle
	
	\begin{abstract}%
		It is a common practice in the machine learning community to assume that the
		observed data are noise-free in the input attributes. Nevertheless, scenarios with input noise
		are common in real problems, as measurements are never perfectly accurate. If this input noise is not
		taken into account, a supervised machine learning method is expected to perform sub-optimally.
		In this paper, we focus on multi-class classification problems and use Gaussian processes (GPs) as the
		underlying classifier. Motivated by a data set coming from the astrophysics domain, we hypothesize that
		the observed data may contain noise in the inputs. Therefore, we devise several multi-class GP
		classifiers that can account for input noise. Such classifiers can be efficiently trained using
		variational inference to approximate the posterior distribution of the latent variables of the
		model. Moreover, in some situations, the amount of noise can be known before-hand. If this is the case, it
		can be readily introduced in the proposed methods. This prior information is expected to lead to better
		performance results. We have evaluated the proposed methods by carrying out several experiments,
		involving synthetic and real data. These include several data sets from the UCI repository, the 
		MNIST data set and a data set coming from astrophysics.
		The results obtained show that, although the classification error is similar across methods, the 
		predictive distribution of the proposed methods is better, in terms of the test log-likelihood, than the 
		predictive distribution of a classifier based on GPs that ignores input noise.
	\end{abstract}
	
	\begin{keywords}
		Gaussian processes, Multi-class classification, Input dependent noise
	\end{keywords}
	
	\section{Introduction}
	
	Multi-class classification problems involve predicting a class label $y$ that can take values
	in a discrete set of $C$ labels $\{1,\ldots,C\}$ with $C>2$ \citep{Murphy2012}. For this task, one should 
	use the information contained in the input attributes $\mathbf{x}\in\mathds{R}^d$, with $d$ the dimensionality of the data. That is, we assume input attributes in the real line.
	In order to infer the relation between $y$ and $\mathbf{x}$ it is assumed that one can use some training data in the form of $N$ pairs $(\mathbf{x},y)$, namely, $\{(\mathbf{x}_i,y_i)\}_{i=1}^N$. Multi-class classification problems arise in a huge variety of fields, from industry to science. Most of the times, however, it is common to have data sets whose inputs $\mathbf{x}$ are the result of experimental measurements. These measurements are unavoidably contaminated with noise, as a consequence of measurement error. Furthermore, in some situations, the errors in the measurement of the explaining attributes can be well determined. See \citet{barford1985experimental} for further details. 
	Incorporating this inductive bias, or prior knowledge about the particular characteristics of the inputs of the classification problem is expected to lead to better results when training a classifier to predict $y$ given $\mathbf{x}$. 
	Conversely, ignoring errors or noise in the input measurements is expected to lead to sub-optimal results when 
	this is indeed the case. In this research work we have empirically validated this hypothesis on 
	several commonly used data sets extracted from the UCI repository \citep{bay2000uci}, as well as on the well-known 
	MNIST data set \citep{lecun1998gradient}. Even for these commonly used data sets in the machine learning community, we 
	find better prediction results by assuming the presence (and learning the amount) of noise in some of the corresponding 
	input variables, as we will show later. 
	
	While inference tasks on data with noisy attributes have been considered since long time in the context of regression---see for example \citet{NR}, or more recently 
	\citet{mchutchon_gaussian_2011}, in the context of Gaussian processes---the specific 
	case of multi-class classification has received much less attention from the literature, 
	with a few exceptions \citep{saez2014analyzing}. Taking into account the presence of noise in the 
	input is, as we show below, potentially essential to better modeling the conditional distribution  
	$p(y|\mathbf{x})$ giving rise to the observed labels. Considering input noise is also expected to have an 
	important impact on the classification of points that are not far from the decision boundaries, since those 
	are regions of the input space in which the data is more susceptible of being misclassified (at least in the case 
	of additive noise with finite variance). Needless to say, both improving the estimated underlying 
	predictive distribution and the better confidence in the classification of \emph{difficult} 
	points are two desirable properties for real-world problems in generic science applications, as for example in the 
	case of the medical domain or in astrophysics applications \citep{gal2016uncertainty}.
	
	Gaussian Processes (GPs) are machine learning methods that are inherently specified within a Bayesian 
	framework \citep{rasmussen2005book}. Therefore, they can deliver probabilistic outputs that allow to 
	extract uncertainty in the predictions made.  This uncertainty is specified in terms of a predictive distribution 
	for the target variable and may arise both from the modeling of intrinsic noise and also due to the lack 
	of observed data. GPs are also non-parametric models. Therefore, their expressiveness 
	grows as the number of data points in the training set, of size $N$, increases. GPs are, however, expensive to train since their complexity is in $\mathcal{O}(N^3)$. Specifically, they require the inversion of a covariance matrix of size $N\times N$. These methods also suffer from the difficulty of requiring approximate techniques to compute the posterior distribution of the latent variables of the model in the case of classification tasks. This posterior distribution is required to compute the predictive distribution for the target variable. Nevertheless, in spite of this difficulties, GPs have been successfully used to address multi-class classification problems and have been shown to be competitive with other approaches such as support vector 
	machines or neural networks \citep{Kim2006,HernandezRobust2011,henao2012predictive,hensman2015,villacampa2017}. 
	
	To alleviate the problems of scalability of GPs several approximations have been proposed in the literature. Among them,
	the most popular ones are based on using inducing points representations \citep{snelson2006sparse,titsias2009variational}.
	These techniques consist in introducing a set of $M \ll N$ inducing points that are carefully chosen to approximate the
	full GP model. Specifically, the locations of the inducing points are optimized during training alongside with any other GP hyper-parameter, by maximizing an estimate of the marginal likelihood \citep{hensman2015,villacampa2017}.
	The complexity of these approximations is in $\mathcal{O}(NM^2)$, which is significantly better 
	than $\mathcal{O}(N^3)$ when $M \ll N$. Importantly, sparse approximations based on inducing points can be combined 
	with stochastic optimization techniques, which allow to scale GPs to very large data sets \citep{HensmanMG15,villacampa2017}. 
	Nevertheless, in spite of this, to our knowledge, all current methods for multi-class GPs classification assume noiseless input attributes.  In this paper we extend the framework of multi-class GP classification to account for this type of noise.
	
	Our work is motivated by a concrete multi-class classification problem and from a data set coming from astrophysics, dealing 
	with measurements, by the Fermi-LAT instrument operated by NASA, of point-like sources of photons in the gamma ray energy 
	range all over the sky (\url{https://fermi.gsfc.nasa.gov/}). The experimental measurements obtained from such a system are unavoidably contaminated with noise and, in practice, the level of noise in some of the explaining attributes is known and can be 
	well determined. As it turns out, at present, a significant fraction of those point-like 
	sources are not associated to or labeled as known astrophysical gamma rays sources as for example pulsars, blazars 
	or quasars \citep{Fermi-LAT:2019yla}. It is thus of paramount importance for the physics related to those measurements to 
	know whether those point-like sources belong to the standard astrophysical classes, or instead whether they are part of 
	other more exotic kind of sources. As a study-case exercise, though, we train a fully supervised classifier using  exclusively the
	sources which do have labels already. Further details of the data set are given when presenting our experimental results.
	These show that a GP multi-class classifier which considers input noise can obtain better 
	predictive distributions in terms of the test log-likelihood in this data set.
	
	In this paper we focus on multi-class classification problems. The reason for this 
	is partially that this is precisely the setting of the data set coming from astrophysics that motivated 
	this work. Besides this, multi-class classification using GPs is relevant since it has systematically 
	received less attention from the machine learning community than other settings such as regression or 
	binary classification \citep{kuss205,titsias2009variational,hensman2013}. 
	This is probably related to the fact that addressing multi-class problems with 
	GPs is more challenging. More precisely, in a multi-class setting there is one latent function per class label 
	that has to be modeled. By contrast, in the regression and binary classification settings in there is only one
	latent function. Multi-class problems also involve the difficulty of dealing with complicated likelihood
	factors which often have the form of soft-max functions or Gaussian integrals that need to be 
	approximated and that may even lack a closed form expression. See, \emph{e.g.}, \citet{villacampa2017}.
	In spite of this, we also give some intuitions about how to extend the proposed approach to other
	supervised learning settings different from multi-class classification.
	
	To account for input noise in the context of multi-class GP classification, we describe three different methods. A first 
	approach is based on considering the actual noiseless input attributes (which we denote by $\mathbf{x}$) 
	as a latent variable, to then perform approximate Bayesian inference and compute their posterior distribution.
	The required computations are approximated using variational inference \citep{blei2017variational} combined with Monte Carlo methods and stochastic optimization. A second approach considers a first order Taylor expansion of the predictive mean of the GPs contained in the multi-class GP classifier (typically, one different GP per each potential class label), following \citet{mchutchon_gaussian_2011}. Under this linear approximation
	the input noise is simply translated into output noise, which is incorporated in the modeling process. The variance
	of the output noise is determined by the slope of the GP predictive mean, which can be obtained using automatic 
	differentiation tools. Variational inference is also used in this second method to approximate the required computations.
	The two methods described to account for input noise in the context of multi-class GP classification are validated 
	on several experiments, including synthetic data, data sets from the UCI repository, MNIST, and the data set related 
	to astrophysics that motivated this work described above. These experiments give empirical claims supporting our 
	hypothesis that our methods can effectively deal with noise in the inputs. In particular, we have consistently 
	observed that the predictive distribution of the methods proposed is better, in terms of the test log-likelihood, 
	than the one of a standard multi-class GP classifier that ignores input noise. The prediction error of 
	the proposed method is, however, similar. A better predictive distribution means that in general the uncertainty 
	about the potential class label of a data instance is better modeled. We illustrate that this is indeed the case
	in a set of active learning experiments. In these experiments one chooses the most informative points from 
	a validation set to be labeled and included in the training set with the goal of reducing the prediction error.
	The most informative points are those for which the model prediction is most uncertain.
	Our results confirm that by taking into account input noise a better prediction error is obtained in these experiments.
	
	The rest of the manuscript is organized as follows: We first introduce the fundamentals about multi-class GP classification 
	and sparse GPs in Section \ref{sec:gps}. Section \ref{sec:nimgp_gen} describes the proposed models and methods to account for input noise in the observed attributes. Related work about input noise in GPs and other machine learning methods 
	is described in Section \ref{sec:related_work}. Section \ref{sec:experiments} illustrates the empirical performance of 
	the proposed methods. Finally, Section \ref{sec:conclusions} gives the conclusions of this work.
	
	\section{Multi-class Gaussian Process Classification}
	\label{sec:gps}
	
	In this section we describe how Gaussian processes (GPs) can be used to address multi-class 
	classification problems. We consider first a noiseless input setting. Next, in the following section, we describe
	how noisy inputs can be incorporated into the model. Assume a data set consisting of 
	$N$ instances with $\mathbf{X} = (\mathbf{x}_1, \ldots, \mathbf{x}_N)^\text{T}$ the 
	observed explaining attributes and $\mathbf{y} = (y_1,\ldots, y_N)^\text{T}$ the target class labels, where 
	$y_i \in \{1,\ldots C\}$ and $C>2$ is the number of classes. The task of interest is to make predictions about the label $y^*$ of a new instance $\mathbf{x}^\star$ given the observed data $\mathbf{X}$ and $\mathbf{y}$.
	
	Following the representation introduced by \cite{Kim2006} for multi-class classification with GPs, we assume that each class label has been obtained with the labeling rule
	\begin{align}
		\label{eq:labelling_rule}
		y_i & =  \underset{c}{\text{arg max}} \quad f^c(\mathbf{x}_i)\,,
	\end{align}
	where  $f^c(\cdot)$, for $c=1,\ldots,C$, are different latent functions, each one of them corresponding to 
	a different class label. Therefore the class label has been obtained simply by considering the latent function with the 
	largest value at the data point $\mathbf{x}_i$. Let $\mathbf{f}_i=(f^1(\mathbf{x}_i),\ldots,f^C(\mathbf{x}_i))^\text{T}$.
	Under this labeling rule the likelihood of the value of each latent function at a training point is given by
	\begin{align*}
		p(y_i|\mathbf{f}_i) &= \prod_{c\neq y_i} \Theta\left(f^{y_i}(\mathbf{x}_i) - f^c(\mathbf{x}_i)\right)\,,
	\end{align*}
	where $\Theta(\cdot)$ is a Heaviside step function. Other likelihood functions such as the soft-max likelihood arise 
	simply by considering and marginalizing Gumbel noise around the latent functions $f^c(\cdot)$ \citep{maddison2014}. 
	Here we instead consider Gaussian noise around each $f^c$, as described later on. 
	To account for labeling errors, we consider that the actual class label $y_i$ associated to $\mathbf{x}_i$ 
	could have been flipped with probability $\epsilon$ to some other class label 
	\citep{HernandezRobust2011,hensman2015}. Under this setting, the likelihood becomes
	\begin{align}
		\label{eq:likelihood_function}
		p(y_i|\mathbf{f}_i) &= (1-\epsilon) \prod_{c\neq y_i} \Theta\left(f^{y_i}(\mathbf{x}_i) - f^c(\mathbf{x}_i)\right) + 
		\frac{\epsilon}{C-1} \left[1 - \prod_{c\neq y_i} \Theta\left(f^{y_i}(\mathbf{x}_i) - f^c(\mathbf{x}_i)\right)\right]\,.
	\end{align}
	In order to address multi-class classification with GPs, a GP prior is assumed for 
	each latent function $f^c(\cdot)$ \citep{rasmussen2005book}. 
	That is, $p(f^c) \sim \mathcal{GP}(0, k_{\theta}(\cdot, \cdot))$, where 
	$k_{\theta_c}(\cdot, \cdot)$ is a covariance function, with hyper-parameters $\theta_c$. 
	Popular examples of covariance functions include the squared exponential covariance function. Namely,
	\begin{align*}
		k_{\theta_c}(\mathbf{x},\mathbf{x}') &= \sigma^2 \exp \left \{ - \frac{1}{2} \sum_{j=1}^d\frac{(x_j - x_j')^2}{\ell_j}\right\}  + I[\mathbf{x} =\mathbf{x}'] \sigma_0^2\,,
	\end{align*}
	where $I[\cdot]$ is an indicator function and $\theta_c=\{\sigma^2, \sigma_0, \{\ell_j\}_{j=1}^d\}$ are the hyper-parameters. 
	More precisely, $\sigma^2$ is the amplitude parameter, $\ell_j$ are the length-scales and $\sigma^2_0$ is the level of additive Gaussian noise around 
	$f^c$. See \citet{rasmussen2005book} for further details. In practice, the hyper-parameters will be different for each 
	latent function $f^c(\cdot)$. We have ignored here their dependence on the latent function $f^c(\cdot)$ for the sake of readability.
	
	In order to make predictions about the potential class label of a new data point $\mathbf{x}^\star$, 
	one would like to compute the posterior distribution of $\mathbf{f} = \{\mathbf{f}_i\}_{i=1}^N$. This distribution summarizes which 
	values of the latent functions are compatible with the observed data. The posterior distribution can be computed using Bayes' rule as follows:
	\begin{align}
		p(\mathbf{f}|\mathbf{y}) =  \frac{p(\mathbf{y}|\mathbf{f}) p(\mathbf{f})}{p(\mathbf{y})} = \frac{\left[\prod_{i=1}^N p(y_i|\mathbf{f}_i)\right]
			\left[ \prod_{c=1}^C p(\mathbf{f}^c) \right]}{p(\mathbf{y})}\,.
		\label{eq:posterior_non_sparse}
	\end{align}
	where $\mathbf{f}^c=(f_c(\mathbf{x}_1),\ldots,f_c(\mathbf{x}_N))^\text{T}$ 
	and $p(\mathbf{f}^c)=\mathcal{N}(\mathbf{f}^c|\mathbf{0}, \mathbf{K}^c)$ is
	a multi-variate Gaussian distribution with zero mean and covariance matrix 
	$\mathbf{K}^c$, with $K^c_{i,j} = k_{\theta_c}(\mathbf{x}_i,\mathbf{x}_j)$.
	The denominator in the previous expression, $p(\mathbf{y})=\int p(\mathbf{y}|\mathbf{f})p(\mathbf{f})d\mathbf{f}$,  is just a normalization constant and is known as 
	the marginal likelihood. It can be maximized to obtain good values for the model hyper-parameters $\theta_c$, for $c=1,\ldots,C$. 
	Note that in this setting, we assume independence among the latent functions $f^c(\cdot)$, 
	since the prior factorizes as $ p(\mathbf{f})=\prod_{c=1}^C p(\mathbf{f}^c)$.
	
	A practical problem, however, is that the non-Gaussian likelihood in (\ref{eq:likelihood_function}) 
	makes infeasible the exact computation of the marginal likelihood, so one has to make use of 
	approximate inference methods to approximate the posterior in (\ref{eq:posterior_non_sparse}). 
	One of the most widely used methods for approximate inference is variational inference (VI), which will 
	be explained in detail in the following sections \citep{titsias2009variational,HensmanMG15,hensman2015}.
	In VI (\ref{eq:posterior_non_sparse}) is approximated by a Gaussian distribution $q$ whose parameters are 
	obtained by minimizing the Kullback-Leibler (KL) divergence between $q$ and the exact posterior.
	The main advantage of VI is that it transforms the approximate inference problem into an optimization problem
	that can be solved using stochastic optimization techniques, and can hence scale to large data sets. Furthermore, 
	it can be easily implemented in modern frameworks for machine learning such as Tensorflow \citep{tensorflow2015}.
	
	Other techniques that have been used for approximate inference in the context of multi-class GP classification 
	include the Laplace approximation \citep{williams1998bayesian}. This technique finds a Gaussian approximation 
	matching the curvature of the posterior at its mode. Nevertheless, it scales poorly to large data sets, 
	since finding the model hyper-parameters by approximately maximizing the marginal 
	likelihood requires a double loop algorithm \citep{rasmussen2005book}, and there is evidence showing 
	that other methods (such as expectation propagation) give better results, at least for binary classification 
	\citep{nickisch2008approximations}. Expectation propagation (EP) has also been used for approximate inference 
	in multi-class GP classification \citep{riihimaki2013nested,villacampa2017}. This method is competitive with VI and 
	sometimes gives better predictive distributions.  Unlike VI, EP approximately minimizes the 
	KL-divergence between the posterior (\ref{eq:posterior_non_sparse}) and an approximate Gaussian 
	distribution $q$. The disadvantage is that it is based on a set of update rules 
	to find $q$ and not on solving directly an optimization problem. 
	These update rules are difficult to implement in frameworks such as Tensorflow. 
	
	The KL-divergence is generalized by a set of divergences known as the $\alpha$-divergence \citep{bui2017unifying}. The 
	use of these divergences for approximate inference in multi-class GP classification has been investigated 
	by \cite{villacampa2020}, showing that sometimes one can obtain better results than those of VI or EP.
	In this work we focus on VI for approximate inference because of its simplicity and ease of 
	implementation. The potential use of $\alpha$-divergences for approximate inference is left as future work.
	
	\subsection{Sparse Gaussian Processes}
	\label{sec:sparse_gps}
	
	A difficulty of the method described so far is that, even if the likelihood was 
	Gaussian and exact inference was tractable, the cost of computing the posterior 
	distribution would be in $\mathcal{O}(N^3)$, where $N$ is the training set size. 
	The reason for this cost is the need of inverting the prior covariance 
	matrices $\mathbf{K}^c$. See \citet{rasmussen2005book} for further details.
	As a consequence, multi-class GPs classification would only be applicable on a data set of at most a few thousand data points.
	
	A popular and successful approach to reduce the previous cost, is to consider an approximation based on sparse Gaussian processes \citep{titsias2009variational}. 
	Under a sparse setting, a set of $M$ pseudo-inputs or inducing points is introduced associated to each latent function $f^c(\cdot)$. 
	Namely, $\mathbf{Z}^c=(\mathbf{z}_1^c,\ldots,\mathbf{z}_M^c)$. These points $\mathbf{Z}^c$ will lie in the same space 
	as the training data. Namely, $\mathds{R}^d$,  and their locations will be
	specified during training, simply by maximizing an estimate of the marginal 
	likelihood. Associated to these inducing points $\mathbf{Z}^c$ we 
	will consider some inducing outputs $\mathbf{u}^c$, where $u_j^c = f^c(\mathbf{z}_j^c)$. 
	The process values at each $\mathbf{x}_i$,
	\emph{i.e.}, $\mathbf{f}^c=(f^c(\mathbf{x}_1),\ldots,f^c(\mathbf{x}_N))^\text{T}$, 
	are characterized by $\mathbf{Z}^c$ and $\mathbf{u}^c$ and 
	can then be obtained from the predictive distribution of a GP as follows:
	\begin{align}
		p(\mathbf{f}^c|\mathbf{u}^c) = \mathcal{N}\left(\mathbf{f}^c| \mathbf{K}^c_{\mathbf{X},
			\mathbf{Z}^c} (\mathbf{K}^c_{\mathbf{Z}^c,\mathbf{Z}^c})^{-1} \mathbf{u}^c, 
		\mathbf{K}^c_{\mathbf{X},\mathbf{X}} - \mathbf{K}^c_{\mathbf{X},\mathbf{Z}^c} 
		(\mathbf{K}^c_{\mathbf{Z}^c,\mathbf{Z}^c})^{-1} \mathbf{K}_{\mathbf{Z}^c,\mathbf{X}}^c \right)\,,
		\label{eq:conditional}
	\end{align}
	where $\mathbf{K}^c_{\mathbf{X},\mathbf{Z}^c}$ is a $N\times M$ matrix of covariances of $f^c(\cdot)$ between the process values at the observed data points $\mathbf{X}$ 
	and the inducing points $\mathbf{Z}^c$. Similarly, $\mathbf{K}^c_{\mathbf{Z}^c,\mathbf{Z}^c}$ is the $M \times M$ covariance matrix. Each entry in this matrix contains the 
	covariances among the process values at the inducing points $\mathbf{Z}^c$. Under the sparse approximation, the prior for each $\mathbf{u}^c$ is simply the Gaussian process prior. Namely,
	\begin{align*}
		p(\mathbf{u}^c) = \mathcal{N}\left(\mathbf{u}^c|\mathbf{0}, \mathbf{K}^c_{\mathbf{Z}^c,\mathbf{Z}^c}\right)\,.
	\end{align*}
	Importantly, now one only has to invert the matrices $\mathbf{K}^c_{\mathbf{Z}^c,\mathbf{Z}^c}$ of size $M\times M$ and compute the product
	$\mathbf{K}^c_{\mathbf{X}, \mathbf{Z}^c} (\mathbf{K}^c_{\mathbf{Z}^c,\mathbf{Z}^c})^{-1}$. Therefore,
	the training cost will be in $\mathcal{O}(M^2N)$, which is significantly better if $M \ll N$.
	
	In practice, the inducing point values $\mathbf{u}^c$ will be unknown and they are treated 
	as latent variables of the model. An approximate Gaussian posterior distribution 
	will be specified for them. Namely, $q(\mathbf{u}^c)=\mathcal{N}(\mathbf{u}^c|\mathbf{m}_c,\mathbf{S}_c)$ for $c=1,\ldots,C$.
	This uncertainty about $\mathbf{u}^c$ can be readily introduced in (\ref{eq:conditional}) simply 
	by marginalizing these latent variables. The result
	is a Gaussian distribution for $\mathbf{f}^c$ with extra variance due to the randomness of $\mathbf{u}^c$. That is,
	\begin{align}
		p(\mathbf{f}^c|\mathbf{y}) & \approx \int p(\mathbf{f}^c|\mathbf{u}^c) q(\mathbf{u}^c) = \mathcal{N}(\mathbf{f}^c|\bm{\mu}_c, \bm{\Sigma}_c)\,,
		\label{eq:marginalizing_u}
	\end{align}
	where
	\begin{align*}
		\bm{\mu}_c &= \mathbf{K}^c_{\mathbf{X}, \mathbf{Z}^c} (\mathbf{K}^c_{\mathbf{Z}^c,\mathbf{Z}^c})^{-1} \mathbf{m}_c\,,  \\
		\bm{\Sigma}_c &= \mathbf{K}^c_{\mathbf{X},\mathbf{X}} - \mathbf{K}^c_{\mathbf{X},\mathbf{Z}^c}
		(\mathbf{K}^c_{\mathbf{Z}^c,\mathbf{Z}^c})^{-1} (\mathbf{K}^c_{\mathbf{Z}^c,\mathbf{Z}^c} + 
		\mathbf{S}_c) (\mathbf{K}^c_{\mathbf{Z}^c,\mathbf{Z}^c})^{-1} \mathbf{K}_{\mathbf{Z}^c,\mathbf{X}}^c\,.
	\end{align*}
	In the following sections we describe how to compute the parameters of each approximate 
	distribution $q(\mathbf{u}^c)$, for $c=1,\ldots,C$, using variational inference. 
	
	\section{Multi-class GP Classification with Input Noise}
	\label{sec:nimgp_gen}
	
	In this section we describe the proposed approaches for dealing with noisy inputs 
	in the context of multi-class GP classification.
	Let us consider that $\tilde{\mathbf{X}}$ is the matrix of noisy observations in which the data patterns are contaminated 
	with additive Gaussian noise with some mean and some variance. Again, consider that $\mathbf{X}$ is the matrix of noiseless inputs. That is,
	\begin{align}
		\tilde{\mathbf{x}}_i & = \mathbf{x}_i + \bm{\epsilon}_i\,, &  \bm{\epsilon}_i  & \sim \mathcal{N}(\mathbf{0}, \mathbf{V}_i)\,,
		\label{eq:noise_model}
	\end{align}
	where $\tilde{\mathbf{x}}_i \in \mathds{R}^d$ is a particular observation and $\mathbf{V}_i$ is a $d\times d$ diagonal matrix.
	That is, we assume independent additive Gaussian noise for the inputs. For the moment, we consider that the variance of 
	the additive noise $\mathbf{V}_i$ is known beforehand. Later on, we describe how this parameter can be inferred from the training data.
	Recall that we assume input attributes in the real line.
	
	\subsection{Modeling the Input Noise Using Latent Variables}
	\label{sec:nimgp}
	
	A first approach for taking into account noisy inputs $\tilde{\mathbf{X}}$ in 
	the context of GP multi-class classification is based on making approximate Bayesian inference
	about the actual noiseless inputs $\mathbf{x}_i$, for $i=1,\ldots,N$. Importantly, these 
	variables will be latent. The observed variables
	will be the ones contaminated with Gaussian noise $\tilde{\mathbf{x}}_i$, for $i=1,\ldots,N$. With this 
	goal, note that the assumption made in (\ref{eq:noise_model}) 
	about the generation of $\tilde{\mathbf{x}}_i$ provides a likelihood function for the actual observation. That is,
	\begin{align*}
		p(\tilde{\mathbf{x}}_i|\mathbf{x}_i) = \mathcal{N}(\tilde{\mathbf{x}}_i|\mathbf{x}_i, \mathbf{V}_i)\,.
	\end{align*}
	In order to make inference about the noiseless observation $\mathbf{x}_i$, we need to specify a prior distribution for that variable.
	In practice, however, the actual prior distribution for $\mathbf{x}_i$ is specific of each classification problem and unknown, in general.
	Therefore, we set the prior for $\mathbf{x}_i$ to be a multi-variate Gaussian with a broad variance. 
	Namely,
	\begin{align*}
		p(\mathbf{x}_i) = \mathcal{N}(\mathbf{x}_i|\mathbf{0},\mathbf{I}s)\,,
	\end{align*}
	where $\mathbf{I}$ is the identity matrix and $s$ is chosen to have a large value, 
	\emph{i.e.}, $s=1,000$, in order to make it similar to a non-informative uniform distribution. 
	This prior has shown good results in our experiments.
	
	\subsubsection{Joint and Posterior Distribution}
	
	The first step towards making inference about $\mathbf{x}_i$ is to describe the joint distribution of 
	all the variables of the model (observed and latent). This distribution is given by
	\begin{align*}
		p(\mathbf{X},\tilde{\mathbf{X}},\mathbf{y},\mathbf{F},\mathbf{U}) &= 
		\left[ \prod_{i=1}^N p(y_i|\mathbf{f}_i) \right] \left[\prod_{c=1}^C  p(\mathbf{f}^c|\mathbf{u}^c) p(\mathbf{u}^c) \right]
		\left[ \prod_{i=1}^N p(\tilde{\mathbf{x}}_i|\mathbf{x}_i) p(\mathbf{x}_i) \right]
		\,,
	\end{align*}
	where $\tilde{\mathbf{X}} = (\tilde{\mathbf{x}}_1,\ldots,\tilde{\mathbf{x}}_N)^\text{T}$ 
	is the matrix of noisy observations, $\mathbf{X}=(\mathbf{x}_1,\ldots,\mathbf{x}_N)^\text{T}$ 
	is the matrix of actual noiseless inputs, $\mathbf{y}$ is the vector of observed labels, $\mathbf{F}=(\mathbf{f}_1,\ldots,\mathbf{f}_N)^\text{T}$ is 
	the matrix of process values at the actual noiseless inputs (\emph{i.e.}, at each $\mathbf{x}_i$ instead of at each $\tilde{\mathbf{x}}_i$), 
	and $\mathbf{U}=(\mathbf{u}^1,\ldots,\mathbf{u}^C)^\text{T}$ is the matrix of process values at the inducing points.
	
	The posterior distribution of the latent variables, \emph{i.e.}, $\mathbf{X}$, $\mathbf{F}$ and $\mathbf{U}$ is 
	obtained using Bayes' rule:
	\begin{align}
		p(\mathbf{X},\mathbf{F},\mathbf{U}|\mathbf{y},\tilde{\mathbf{X}}) & = \frac{p(\tilde{\mathbf{X}},\mathbf{X}, \mathbf{y},\mathbf{F},\mathbf{U})}{p(\mathbf{y},\tilde{\mathbf{X}})}\,.
		\label{eq:posterior}
	\end{align}
	Again, as in the case of standard multi-class classification where there is noise in the inputs, 
	computing this posterior distribution is intractable and approximate inference will be required. To approximate this distribution we employ variational inference,
	as described in the next section.
	
	\subsubsection{Approximate Inference Using Variational Inference}
	\label{sect:vi}
	
	We will use variational inference (VI) as the approximate inference method \citep{jordan99introduction}. The posterior approximation that will 
	target the exact posterior (\ref{eq:posterior}) is specified to be
	\begin{align}
		q(\mathbf{X},\mathbf{F},\mathbf{U}) & = \left[ \prod_{c=1}^C p(\mathbf{f}^c|\mathbf{u}^c) q(\mathbf{u}^c) \right]
		\left[ \prod_{i=1}^N q(\mathbf{x}_i)\right]\,,
		\label{eq:post_approx_orig}
	\end{align}
	where
	\begin{align}
		q(\mathbf{u}^c) & = \mathcal{N}(\mathbf{u}^c|\mathbf{m}_c,\mathbf{S}_c)\,, & 
		q(\mathbf{x}_i) & = \mathcal{N}(\mathbf{x}_i|\bm{\mu}_i^x,\mathbf{V}_i^x)\,, 
		\label{eq:post_approx}
	\end{align}
	with $\mathbf{V}_i^x$ a diagonal matrix. This posterior approximation assumes independence 
	among the different GPs of the model and the actual noiseless inputs $\mathbf{X}$.
	
	To enforce that $q$ looks similar to the target posterior distribution, VI minimizes the Kullback-Leibler (KL) divergence
	between $q$ and the exact posterior $p$, given by the distribution in (\ref{eq:posterior}). This is done indirectly by maximizing 
	the evidence lower bound $\mathcal{L}$. See \citet{jordan99introduction} for further details. The evidence lower bound (ELBO) is given by
	\begin{align}
		\mathcal{L} & = \mathds{E}_q\left[ \log \frac{p(\mathbf{X},\tilde{\mathbf{X}},\mathbf{y},\mathbf{F},\mathbf{U})}{q(\mathbf{X},\mathbf{F},\mathbf{U})} \right]
		\nonumber \\
		& = \sum_{i=1}^N \mathds{E}_q\left[ \log p(y_i|\mathbf{f}_i)\right] + \sum_{i=1}^N \mathds{E}_q[\log p(\tilde{\mathbf{x}}_i|\mathbf{x}_i)]\nonumber \\
		& \quad - \sum_{c=1}^C \text{KL}(q(\mathbf{u}^c)|p(\mathbf{u}^c))  - \sum_{i=1}^N \text{KL}(q(\mathbf{x}_i)|p(\mathbf{x}_i))\,,
		\label{eq:elbo}
	\end{align}
	where $\text{KL}(\cdot|\cdot)$ is the Kullback-Leibler divergence and where we have used the fact that the factors of the form $p(\mathbf{f}^c|\mathbf{u}^c)$
	described in (\ref{eq:conditional}) and present in both the joint distribution and in $q$ will cancel.
	
	One problem that arises when computing the previous expression is that the first expectation in (\ref{eq:elbo}), 
	\emph{i.e.} $\mathds{E}_q\left[ \log p(y_i|\mathbf{f}_i)\right]$, does not have a closed 
	form solution. It can, however, be computed using a one dimensional quadrature and
	Monte Carlo methods combined with the reparametrization trick \citep{KingmaW13,hensman2015}. 
	Concerning the other factors, the second expectation, 
	$\sum_{i=1}^N \mathds{E}_q[\log p(\mathbf{x}_i|\tilde{\mathbf{x}}_i)]$, is the expectation of the logarithm of a Gaussian distribution, 
	so it can be computed in a closed form. We can also evaluate 
	analytically the KL divergences in the lower bound, 
	$\sum_{k=1}^K \text{KL}(q(\mathbf{u}^k)|p(\mathbf{u}^k))  - \sum_{i=1}^N \text{KL}(q(\tilde{\mathbf{x}}_i)|p(\tilde{\mathbf{x}}_i))$, since they involve Gaussian distributions. 
	
	Importantly, the ELBO in (\ref{eq:elbo}) is expressed as a sum across the observed data points. This means that
	this objective is suitable for being optimized using mini-batches. The required gradients can be obtained using automatic differentiation
	techniques such as those implemented in frameworks such as Tensorflow \citep{tensorflow2015}. 
	Appendix \ref{sec:appendix_a} describes the details about how to obtain an unbiased noisy estimate of 
	the ELBO, $\mathcal{L}$. 
	
	One last remark is that it is possible to show that 
	\begin{align*}
		\log p(\mathbf{y},\tilde{\mathbf{X}}) = \mathcal{L} + \text{KL}[q||p]\,,
	\end{align*}
	where $\text{KL}[q||p]$ is the Kullback-Leibler (KL) divergence between $q$ and 
	the target posterior distribution $p$ in (\ref{eq:posterior}) \citep{jordan99introduction}. 
	After maximizing the ELBO, $\mathcal{L}$, it is expected that the KL term is fairly small and 
	hence $\log p(\mathbf{y},\tilde{\mathbf{X}}) \approx \mathcal{L}$. Therefore, $\mathcal{L}$ can be maximized 
	to find good values for the model hyper-parameters. This is expected to maximize 
	$\log p(\mathbf{y},\tilde{\mathbf{X}})$, which will correspond to a type-II maximum
	likelihood approach \citep{rasmussen2005book}. The locations of the inducing points 
	$\mathbf{Z}^c$, for $c=1,\ldots,C$, are found by maximizing $\mathcal{L}$, 
	an estimate of the marginal likelihood, as in \citet{hensman2015,villacampa2017}. Note, however, that these correspond 
	to parameters of the posterior approximation $q$, defined in (\ref{eq:post_approx_orig}), 
	and not of the described probabilistic model \citep{titsias2009variational}.
	
	\subsubsection{Predictions}
	
	After the maximization of the lower bound (\ref{eq:elbo}), the approximate 
	distribution $q$ is fitted to the actual posterior. 
	The predictive distribution for the class label $y_\star$ of a new instance $\mathbf{x}_\star$ can 
	be approximated by replacing the exact posterior by the posterior approximation in the exact predictive distribution. Namely,
	\begin{align}
		p(y_\star|\mathbf{x}_\star) \approx \int p(y_\star|\mathbf{f}_\star) 
		\left[\prod_{c=1}^C p(f_\star^c|\mathbf{u}^c) q(\mathbf{u}^c) d \mathbf{u}^c \right] 
		p(\mathbf{x}_\star|\tilde{\mathbf{x}}_\star) d \mathbf{x}_\star d \mathbf{f}_\star \,,
		\label{eq:pred}
	\end{align}
	where $p(\mathbf{x}_\star|\tilde{\mathbf{x}}_\star)$ is the posterior distribution of the actual 
	attributes of the new instance given the observed attributes $\tilde{\mathbf{x}}_\star$.
	This posterior is the normalized product of the prior times the likelihood. Therefore, it can be computed in 
	closed form and is a Gaussian. Namely,
	\begin{align*}
		p(\mathbf{x}_\star|\tilde{\mathbf{x}}_\star) & = \frac{p(\tilde{\mathbf{x}}_\star|\mathbf{x}_\star)
			p(\mathbf{x}_\star)}{p(\tilde{\mathbf{x}}_\star)}=\mathcal{N}(\mathbf{x}_\star|\bm{\mu}_\star^x, \mathbf{V}_\star^x)\,, 
	\end{align*}
	where
	$\mathbf{V}_\star^x = (\mathbf{V}_\star^{-1}+\mathbf{I}s^{-1})^{-1}$ and $\bm{\mu}_\star^x = \mathbf{V}_\star^x(\mathbf{V}_\star^{-1}\tilde{\mathbf{x}}_\star)$, and where $\mathbf{V}_\star$ is a diagonal matrix with the variances of 
	the Gaussian noise around $\mathbf{x}_\star$.
	Note that if $s$ is fairly large, as it is in our case, then essentially $\mathbf{V}_\star^x \approx \mathbf{V}_\star$
	
	In general, the integral in (\ref{eq:pred}) is intractable. However, we can generate samples of $\mathbf{x}_\star$ 
	simply by drawing from $p(\mathbf{x}_\star|\tilde{\mathbf{x}}_\star)$ to then compute a Monte Carlo approximation:
	\begin{align}
		p(y_\star|\mathbf{x}_\star) \approx \frac{1}{S}\sum_{s=1}^S\int  p(y_\star|\mathbf{f}_\star^s) 
		\left[\prod_{c=1}^C p(f_{s,\star}^c|\mathbf{u}^c) q(\mathbf{u}^c) d \mathbf{u}^c \right]d \mathbf{f}_\star^s\,,
		\label{eq:montecarlo_predictive}
	\end{align}
	where $S$ is the number of samples, 
	$\mathbf{f}_\star^s=(f^1(\mathbf{x}_\star^s),\ldots, f^C(\mathbf{x}_\star^s))^\text{T}$,
	$f_{s,\star}^c=f^c(\mathbf{x}_\star^s)$ with $\mathbf{x}_\star^s$ the generated $s$ sample of $\mathbf{x}_\star$. 
	
	The only remaining thing is how to compute the integral in the right hand side of (\ref{eq:montecarlo_predictive}).
	It turns out that the integral with respect to each $\mathbf{u}^c$ can be computed analytically using (\ref{eq:marginalizing_u}).
	The integral with respect to $\mathbf{f}_\star^s$ can be approximated using a one-dimensional quadrature.
	In particular, under the likelihood function in (\ref{eq:likelihood_function}), 
	the approximation of the predictive distribution becomes
	\begin{align}
		p(y_\star|\mathbf{x}_\star) \approx \frac{\epsilon}{C-1} + (1 - \epsilon) 
		\frac{1}{S}\sum_{s=1}^S\int \mathcal{N}(f^{y_\star}_s|m_{y_\star}^s, v_{y_\star}^s)
		\prod_{c\neq y_\star}\Phi\left(\frac{f^c_s - m_c^s}{\sqrt{v_c^s}}\right) df^{y_\star}_s\,,
		\label{eq:quad}
	\end{align}
	where $f^{y_\star}_s = f^{y_\star}(\mathbf{x}_\star^s)$, $f^c_s=f^c(\mathbf{x}_\star^s)$, 
	$\Phi(\cdot)$ is the cumulative probability of a standard Gaussian distribution and
	\begin{align*}
		m_{c}^s &= \mathbf{k}^c_{\mathbf{x}_\star^s,\mathbf{Z}^c}(\mathbf{K}^c_{\mathbf{Z}^c,\mathbf{Z}^c})^{-1} \mathbf{m}_c\,, \\
		v_{c}^s &= k^c_{\mathbf{x}_\star^s,\mathbf{x}_\star^s} -
		\mathbf{k}^c_{\mathbf{x}_\star^s,\mathbf{Z}^c} (\mathbf{K}^c_{\mathbf{Z}^c,\mathbf{Z}^c})^{-1} 
		\mathbf{K}_{\mathbf{Z}^c,\mathbf{x}_\star^s} + 
		\mathbf{k}^c_{\mathbf{x}_\star^s,\mathbf{Z}^c} (\mathbf{K}^c_{\mathbf{Z}^c,\mathbf{Z}^c})^{-1}
		\mathbf{S}_c(\mathbf{K}^c_{\mathbf{Z}^c,\mathbf{Z}^c})^{-1} \mathbf{k}^c_{\mathbf{Z}^c,\mathbf{x}_\star^s}\,,
	\end{align*}
	for $c=1,\ldots,C$ with $k^c_{\mathbf{x}_\star^s,\mathbf{x}_\star^s}$ the variance of $f^c(\cdot)$ at $\mathbf{x}_\star$,
	$\mathbf{k}^c_{\mathbf{x}_\star^s,{\mathbf{Z}}^c}$ the matrix of covariances between the values of $f^c(\cdot)$ at
	$\mathbf{x}_\star^s$ and $\mathbf{Z}^c$, and $\mathbf{K}^c_{\mathbf{Z}^c,\mathbf{Z}^c}$ the covariances of $f^c(\cdot)$ among
	the inducing points $\mathbf{Z}^c$. Note that the integral over $f^{y_\star}_s$ in (\ref{eq:quad}) 
	has no closed form solution but it can be computed using one-dimensional quadrature. 
	
	\subsection{Amortized Approximate Inference}
	\label{sect:avi}
	
	A limitation of the method described is that the approximate posterior distribution over the noiseless inputs,
	\emph{i.e.}, $\prod_{i=1}^N q(\mathbf{x}_i)$, demands storing in memory a number of parameters that is in $\mathcal{O}(N)$, where $N$ is the
	number of observed instances. Of course, in the big data regime, \emph{i.e.}, when $N$ is in the order of thousands or even millions 
	the memory resources needed to store those parameters can be too high. To alleviate this problem, we propose to use amortized approximate inference to reduce the number of parameters that need to be store in memory \citep{KingmaW13,shu2018amortized}.
	
	Amortized variational inference assumes that the parameters of each distribution $q(\mathbf{x}_i)$, \emph{i.e.}, the mean and the diagonal
	covariance matrix, can be obtained simply as a non-linear function that depends on the observed data instance $(\tilde{\mathbf{x}}_i, y_i)$. 
	This non-linear function is set to be a neural network whose parameters are optimized during training. That is,
	\begin{align*}
		q(\mathbf{x}_i)=\mathcal{N}(\mathbf{x}_i|\bm{\mu}_\theta(\tilde{\mathbf{x}}_i,y_i),\mathbf{V}_\theta(\tilde{\mathbf{x}}_i,y_i))\,,
	\end{align*}
	where both $\bm{\mu}_\theta(\tilde{\mathbf{x}}_i,y_i)$ and $\mathbf{V}_\theta(\tilde{\mathbf{x}}_i,y_i)$ are obtained as the output of a neural network
	with parameters $\theta$ (we use a one-hot encoding for the label $y_i$). Therefore, one only has to store in memory the neural network
	which has a fixed number of parameters. This number of parameters does not depend on $N$.  
	The neural network can be adjusted simply by maximizing w.r.t $\theta$ the evidence lower bound described in (\ref{eq:elbo}).  
	The computational cost of the method is not changed, since the cost for the feed-forward pass of the network to obtain 
	$\bm{\mu}_\theta(\tilde{\mathbf{x}}_i,y_i)$ and $\mathbf{V}_\theta(\tilde{\mathbf{x}}_i,y_i)$ is constant.
	
	Amortized approximate inference introduces the inductive bias that points that are located in similar regions of the input space
	should have similar parameters in the corresponding posterior approximation $q(\mathbf{x}_i)$. Of course, this has the benefit
	property of reducing the number of parameters of the approximate distribution $q$. A second advantage is, however, that the neural network
	can provide a beneficial regularization that is eventually translated into better generalization results \citep{shu2018amortized}. 
	More precisely, our experiments show that amortized variational inference sometimes provides better results than using 
	an approximate distribution $q$ that has separate parameters per each data point $\tilde{\mathbf{x}}_i$.
	
	Besides using this neural network to compute the parameters of $q(\mathbf{x}_i)$, the model is not changed significantly. 
	Prediction is done in the same way, and hyper-parameter optimization is also carried out by maximizing the evidence lower bound. 
	
	\subsection{First Order Approximation}
	\label{sect:foa}
	
	In this section we describe an alternative method to account for input noise in the context of multi-class classification with GPs.
	This method is inspired on work already done for regression problems \citep{mchutchon_gaussian_2011}.
	Consider the relation between the noisy and the noiseless input measurements given by (\ref{eq:noise_model}).
	Now, consider a Taylor expansion of a latent function $f(\cdot)$ around the noiseless measurement $\mathbf{x}_i$. Namely,
	\begin{align}
		f(\mathbf{x}_i + \bm{\epsilon}_i) &= f(\mathbf{x}_i) +
		\bm{\epsilon}_i^\text{T} \frac{\partial f(\mathbf{x}_i)}{\partial \mathbf{x}_i} + \cdots
		\approx
		f(\tilde{\mathbf{x}}_i) +
		\bm{\epsilon}_i^\text{T} \frac{\partial f(\tilde{\mathbf{x}}_i)}{\partial \tilde{\mathbf{x}}_i} + \cdots
		\label{eq:taylor}
	\end{align}
	where $\tilde{\mathbf{x}}_i$ is the noisy observation.
	Since we do not have access to the noiseless input vector $\mathbf{x}_i$, we simply approximate it
	with the noisy one $\tilde{\mathbf{x}}_i$. Note that this last expression involves the derivatives of the GP.
	Although they can be showed to be again GPs (see \citet{rasmussen2005book} for further details), we here
	decide to approximate these derivatives with the derivatives of the mean of GP, as in \citet{mchutchon_gaussian_2011}.
	This approximation corresponds to ignoring the uncertainty about the derivative. Let $\partial_{\overline{f}}(\tilde{\mathbf{x}}_i)$
	denote the d-dimensional vector corresponding to the derivative of the mean of the GP with respect to each input
	dimension at $\tilde{\mathbf{x}}_i$. If we expand the right hand side of (\ref{eq:taylor}) up to the first order terms, we get a linear model.
	Namely,
	\begin{align*}
		f(\mathbf{x}_i + \bm{\epsilon}_i)  &= f(\tilde{\mathbf{x}}_i) +
		\bm{\epsilon}_i^\text{T} \partial_{\overline{f}}(\tilde{\mathbf{x}}_i)\,.
	\end{align*}
	Therefore, the input noise can be understood as output noise whose variance is proportional to square of
	the derivative of the mean value of $f$ at $\tilde{\mathbf{x}}_i$.
	
	The model just described can be combined with the framework for sparse GPs described in Section
	\ref{sec:sparse_gps} to give an alternative posterior predictive distribution
	for $\mathbf{f}^c$ to the one described in (\ref{eq:marginalizing_u}). That is,
	\begin{align}
		p(\mathbf{f}^c|\mathbf{u}^c) & =
		\mathcal{N}\left(\mathbf{f}^c| \bm{\mu}_c, \bm{\Sigma}_c \right)\,,
		\label{eq:modified_conditional}
	\end{align}
	with
	\begin{align*}
		\bm{\mu}_c &= \mathbf{K}^c_{\tilde{\mathbf{X}},\mathbf{Z}^c}
		(\mathbf{K}^c_{\mathbf{Z}^c,\mathbf{Z}^c})^{-1} \mathbf{m}_c\,, \\
		\bm{\Sigma}_c &= 
		\mathbf{K}^c_{\tilde{\mathbf{X}},\tilde{\mathbf{X}}} -
		\mathbf{K}^c_{\tilde{\mathbf{X}},\mathbf{Z}^c}
		(\mathbf{K}^c_{\mathbf{Z}^c,\mathbf{Z}^c})^{-1} 
		\left((\mathbf{K}^c_{\mathbf{Z}^c,\mathbf{Z}^c})^{-1} - \mathbf{S}_c\right)
		(\mathbf{K}^c_{\mathbf{Z}^c,\mathbf{Z}^c})^{-1} 
		\mathbf{K}_{\mathbf{Z}^c,\tilde{\mathbf{X}}} + \bm{\Delta}\,,
	\end{align*}
	where $\mathbf{m}_c$ and $\mathbf{S}_c$ are the parameters of $q(\mathbf{u}^c)$, 
	the covariance matrices are evaluated at the noisy measurements $\tilde{\mathbf{X}}$ and
	\begin{align}
		\Delta_{i,i} & = \partial_{\overline{f}^c}(\tilde{\mathbf{x}}_i)^\text{T} \mathbf{V}_i  \partial_{\overline{f}^c}(\tilde{\mathbf{x}}_i)\,.
		\label{eq:fo_noise_influence}
	\end{align}
	Therefore, $\bm{\Delta}$ is a diagonal matrix whose entries account for the extra output noise that results
	from the corresponding input noise.  This makes sense, since the input noise is expected
	to have a small effect in those regions of the input space in which the latent function is expected to be
	constant. Importantly, in the sparse setting
	\begin{align*}
		\partial_{\overline{f}^c}(\tilde{\mathbf{x}}_i) &= \frac{\partial \bm{\mu}_c}{\partial \tilde{\mathbf{x}}_i} = \frac{\partial \mathbf{k}^c_{\tilde{\mathbf{x}}_i,\mathbf{Z}^c}
			(\mathbf{K}^c_{\mathbf{Z}^c,\mathbf{Z}^c})^{-1} \mathbf{m}_c}{\partial \tilde{\mathbf{x}}_i}\,.
	\end{align*}
	This partial derivative can be easily obtained automatically in modern frameworks for implementing multi-class GP
	classifiers such as Tensorflow \citep{tensorflow2015}.
	The expression in (\ref{eq:modified_conditional}) can replace (\ref{eq:marginalizing_u}) in a standard
	sparse multi-class GP classifier to account for input noise. One only has to provide the corresponding 
	input noise variances $\mathbf{V}_i$, for $i=1,\ldots,N$. Approximate inference in such a model can be carried out
	using variational inference, as described in \citet{hensman2015}.
	
\begin{figure}[t]
		\begin{center}
			\includegraphics[width=0.49\textwidth]{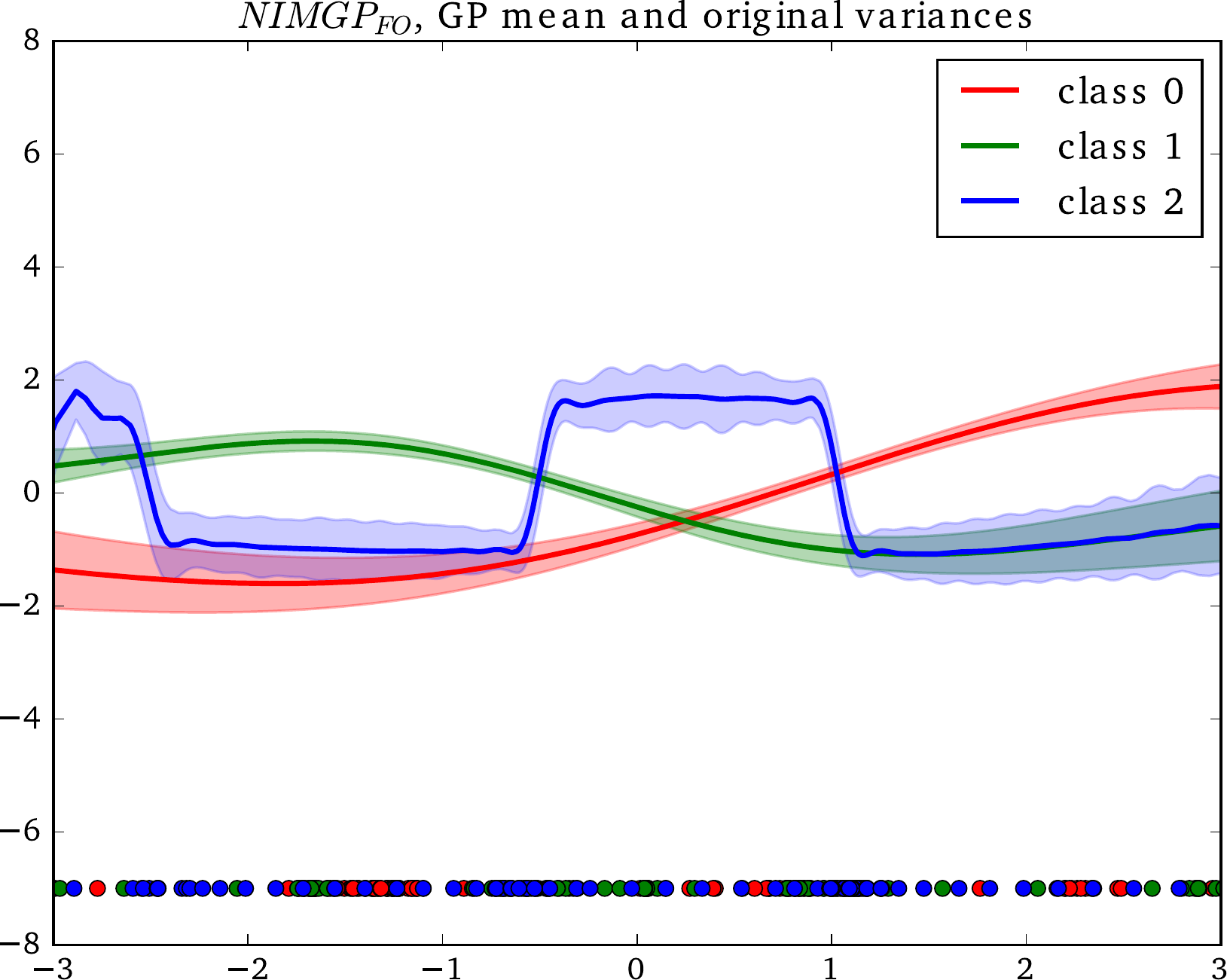}
			\includegraphics[width=0.49\textwidth]{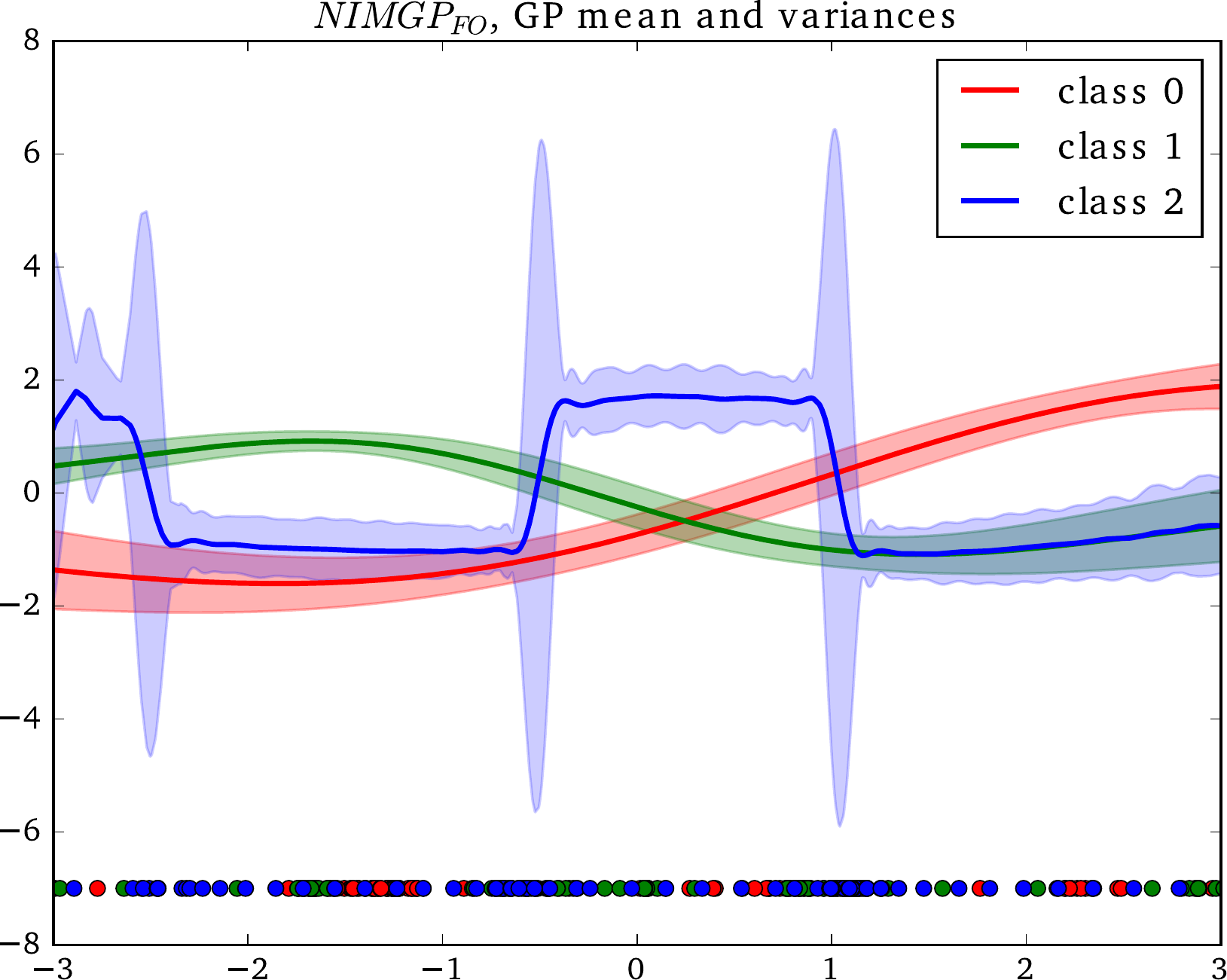}
			
			\caption{\small For the $\text{NIMGP}_\text{FO}$ model, the predictive mean GP and variances (right) and original variances (left), for each 
				latent function, for a classification problem with three classes shown in red, green and blue. The learned locations of the inducing 
				points, $\mathbf{Z}^c$, are shown in a row at the bottom of each figure. Best seen in color.}
			\label{fig:pred_dist_GPs}
		\end{center}
	\end{figure}

	Figure \ref{fig:pred_dist_GPs} shows the predictive distribution of the model described, which we refer to as $\text{NIMGP}_\text{FO}$, 
	for each latent function in a three class toy classification problem described in Section \ref{sec:toy}. The figure on the left
	shows the predictive distribution obtained when the extra term in the predictive variance $\bm{\Delta}$ that depends on the slope of the 
	mean is ignored in (\ref{eq:modified_conditional}). The figure on the right  shows the resulting predictive distribution
	when that term is taken into account. We can observe that the produced effect is to increase the variance of the predictive
	distribution by some amount that is proportional to the squared value of slope of the predictive mean. This is particularly
	noticeable in the case of the latent function corresponding to class number $2$. A bigger variance in the predictive
	distribution for the latent function will correspond to less confidence in the predictive distribution for the class label.
	See Section \ref{sec:toy} for further details.
	
	Figure \ref{fig:pred_dist_GPs} also shows the learned locations of the inducing points for each latent function (displayed at the
	bottom of each image). We observe that they tend to be placed uniformly in the case of the latent functions corresponding
	to class labels 0 and 1. However, in the case of the latent function corresponding to class label 2, they concentrate in specific
	regions of the input space. Namely, in those regions in which the latent function changes abruptly. 
	
	The inductive bias of a machine learning algorithm is a set of assumptions that the chosen methodology uses to predict outputs given, in this case, uncertain inputs. The stochastic estimate of the lower bound, in contrast w.r.t the linear approximation of $\text{NIMGP}_\text{FO}$, is unbiased, as it is shown in Appendix A. In the $\text{NIMGP}_\text{FO}$ method, the input noise is converted to output noise through a linear approximation based in a second-order Taylor expansion, that assumes that the input noise distribution fits a quadratic approximation and regularity assumptions that need not be necessarily true.
	
	\subsection{Learning the Level of Noise in the Inputs}
	\label{sec:learning_noise}
	
	The previous sections assumed that the variance of Gaussian noise associated to each input dimension, \emph{i.e.}, the diagonal matrix $\mathbf{V}_i$,
	is known before-hand. This is the case of many practical problems in which the error associated to the measurement instrument that is used to obtain
	the observed attributes $\tilde{\mathbf{x}}_i$ is well-known. However, in certain situations it can be the case that the variance of the error is unknown.
	In this case, it may still be possible to infer this level of error from the observed data.
	
	Typically, in this case, one will assume that the level of noise is the same across all observed data instances. That is, 
	$\mathbf{V}_i=\mathbf{V}$. This has the advantage of reducing the number of parameters that have to be inferred. 
	To estimate $\mathbf{V}$ one can simply treat this parameter as a hyper-parameter of the model. Its value
	can be estimated simply by type-II maximum likelihood, as any other hyper-parameter of the GP \citep{rasmussen2005book}.
	Under this setting, one simply maximizes the marginal likelihood of the model, \emph{i.e.}, the denominator 
	in Bayes' theorem w.r.t the parameter of interest. This is precisely the approach followed in 
	\citet{mchutchon_gaussian_2011} for regression problems.
	
	Because evaluating the marginal likelihood is infeasible in the models described so far, 
	one has to use an approximation. This approximation can be the
	evidence lower bound described in (\ref{eq:elbo}), which will be similar to the 
	marginal likelihood if the approximate distribution $q$ is an accurate posterior approximation.
	The maximization of (\ref{eq:elbo}) w.r.t $\mathbf{V}$ can be simply done with no 
	extra computational cost using again stochastic optimization techniques.
	
	In general, we will assume that $\mathbf{V}_i$ is known for each data instance. 
	If that is not the case, we will infer that level of noise using the method described in this section.
	
	\subsection{Other Problems Different from Multi-class Classification}
	
	The methods described so far focus on multi-class problems as we are initially motivated by 
	a real-world application from the field of astrophysics. This setup is 
	challenging since there is a latent function for each potential class label. This 
	also results in complex likelihood factors and the need to approximate the first expectation 
	in (\ref{eq:elbo}) using one dimensional quadrature and Monte Carlo methods. 
	Of course, addressing regression problems is straight-forward. In this case, the likelihood
	factors are Gaussian and some of the required computations to evaluate the (\ref{eq:elbo}) are 
	tractable (the only expectation that needs to be approximated is that with respect to each $q(\mathbf{x}_i)$). 
	A single GP is used in this case. There has been plenty of research work considering this setting in 
	the literature and we refer the reader to Section \ref{sec:related_work}.
	
	We describe how the proposed method can be used in problems involving other non-Gaussian likelihood functions. 
	For example, in problems in which the task of interest is to predict count data, 
	we may want to use the likelihood factors that are Poisson distributed 
	\citep{diggle1998model}. That is, $p(y_i|f_i) = \text{Pois}(y_i|\lambda_i)$, where $\lambda_i$ is the 
	rate parameter that may non-linearly depend on $\mathbf{x}_i$ through a GP, 
	\emph{e.g.}, $\lambda_i = \exp(f_i)$, with $f_i$ the process value at $\mathbf{x}_i$. 
	Similarly, in binary classification, where $y_i \in \{-1,1\}$, 
	the likelihood factors can be of the form $p(y_i|f_i) = \Phi(f_iy_i)$, where $\Phi(\cdot)$ is the sigmoid 
	or the probit function \citep{rasmussen2005book}. Multi-label data can also be considered by 
	using similar sigmoid likelihood factors, several GPs and a linear transformation to account for 
	each potential label \citep{panos2019}. A robustified GP for regression can be considered 
	by using Student-t factors in the likelihood, as
	an alternative to the standard Gaussian factors \citep{neal1997}. 
	Namely, $p(y_i|f_i) = \text{Student}(y_i|f_i,\sigma,\mu)$, where $f_i$, $\sigma$ and $\nu$ 
	are the location, scale and degrees of freedom parameters, respectively. 
	Other problems may involve ordinal regression, where the target variables are discrete but ordered 
	\citep{chu2005gaussian}. In that case the likelihood function may involve two probit functions
	to account for the fact that the process value $f_i$ associated to $\mathbf{x}_i$ should lie in a 
	particular interval of the real line. 
	
	All these problems are characterized by non-Gaussian likelihood factors involving a single GP. 
	In this case, the expectation of $\mathds{E}_q[\log p(y_i|f_i)]$ with respect to $q(f_i)$
	required to evaluate (\ref{eq:elbo}) is intractable. However, it can be easily approximated also using a 
	one-dimensional quadrature. The expectation with respect to each $q(\mathbf{x}_i)$ can be approximated using 
	Monte Carlo methods, as in the case of multi-class classification. The evaluation of the proposed 
	method in the described settings is left for future work since the focus of this paper is multi-class classification.
	
	\subsection{Summary of the Proposed Methods to Deal with Input Noise}
	
	Below we briefly describe the different methods explained so far to 
	deal with input noise in the context of multi-class GP classification:
	
	\begin{itemize}
		
		\item {\bf NIMGP}: This is the method described in Section \ref{sec:nimgp}
		which uses latent variables to model the original noiseless inputs.
		It relies on a Gaussian distribution for the actual observation with the mean being a random variable 
		representing the noiseless input. We assume a non-informative Gaussian prior for the noiseless input. 
		The joint posterior distribution that involves non-Gaussian likelihood factors for the process values is 
		approximated using variational inference. Quadrature and Monte Carlo methods are combined with the 
		reparametrization trick to obtain a noisy estimate of the lower bound which is optimized using stochastic methods.  
		
		\item {\bf $\text{NIMGP}_\text{NN}$:} A limitation of the previous method is the need of 
		storing parameters for each data instance. This is a disadvantage in big data applications. 
		To circumvent this problem, amortized approximate inference is employed in this method,
		as explained in Section \ref{sect:avi}. We use a neural network to compute the parameters of 
		posterior approximation for the noiseless attributes of each data instance. 
		This network reduces the number of parameters of the approximate distribution and also regularizes the model. 
		Approximate inference is carried out as in NIMGP.
		
		\item {\bf $\text{NIMGP}_\text{FO}$}: This is the method described in Section 
		\ref{sect:foa}. In this method we adapt for classification with sparse GPs an 
		already proposed method for regression that accounts for noisy inputs using GPs. 
		This method is based on propagating the noise found 
		in the inputs to the variance of the GPs predictive distribution. For this, a 
		local linear approximation of the GP is used at each input 
		location. This allows the input noise to be recast as output noise proportional to the squared gradient of the GP predictive mean.
		
	\end{itemize}
	
	A summary of the parameters employed by each method is provided in Appendix \ref{sec:appendix_b_params}.
	
	\section{Related Work}
	\label{sec:related_work}
	
	In the literature there are some works dealing with GPs and input points corrupted with  
	noise in the context of regression problems \citep{mchutchon_gaussian_2011}. 
	For this, a local linear approximation of the GP at each input point is performed. When this is done, the input 
	noise is translated into output noise proportional to the squared gradient of the GP posterior mean. 
	Therefore, the mentioned paper simplifies the problem of modeling input noise by assuming that the 
	input measurements are deterministic and by inflating the corresponding output variance to compensate
	for the extra noise. When this operation is performed, it leads to output noise variance varying across 
	the input space. This property is defined as \emph{heteroscedasticity}. The model presented in that paper can hence be described as a heteroscedastic GP model. 
	We have extended the linearization approach of \citet{mchutchon_gaussian_2011} 
	to address multi-class classification problems as they only considered regression problems. 
	Furthermore, we have compared such a method, $\text{NIMGP}_\text{FO}$, with another approach that uses latent
	variables to model the noiseless inputs, NIMGP, and that, in principle, does not rely on a 
	linear approximation of the GP. We show that on synthetic data NIMGP performs better than 
	$\text{NIMGP}_\text{FO}$ which indicates that the linear approximation used in 
	$\text{NIMGP}_\text{FO}$ may be inaccurate. In fact, the GP is non-linear in practice. 
	Note that we also consider sparse GPs instead of standard GPs, which make our approach more scalable 
	to data sets with a larger number of instances. These differences are common between our proposed 
	methods and most of the techniques described in this section.
	
	As described in the previous paragraph, one of our proposed methods can be understood as a 
	GP model in which the level of output noise depends on the input location, \emph{i.e.}, the data can be 
	considered to be heteroscedastic. Several works have tried to address such a problem in the context 
	of GPs and regression tasks. In particular, \cite{goldberg1998regression} introduce a second GP to deal 
	with the output noise level as a function of the input location. This approach uses a Markov chain Monte 
	Carlo method to approximate the posterior noise variance, which is time-consuming. A more efficient MAP 
	estimation approach is described by \cite{le2005heteroscedastic}. This other method is applicable in models with likelihood factors in the exponential family and it leads to a convex optimization problem solvable by the 
	Newton method. An interesting extension to the work of \cite{goldberg1998regression} tries 
	to circumvent its computational limitations by replacing the Monte Carlo method with an approximative 
	most likely noise approach \citep{kersting2007most}. This other work learns both the 
	hidden noise variances and the kernel parameters simultaneously. Furthermore, it significantly 
	reduces the computational cost. 
	
	Other related work concerning heteroscedasticity in the context of regression problems is the one of 
	\cite{lazaro2011variational}. This approach also relies on variational inference for approximate inference 
	in the context of GPs. These authors explicitly take into account the input noise, and model it with GP priors.  
	By using an exponential transformation of the noise process, they specify the variance of the output noise.
	Exact inference in the heteroscedastic GP is intractable and the computations need to be approximated using variational inference.
	Variational inference in such a model has equivalent cost to an analytically tractable homoscedastic GP. 
	Importantly, this work focuses exclusively on regression and ignores classification tasks. 
	Furthermore, no GP sparse approximation is considered by these authors. Therefore,
	the problems addressed cannot contain more than a few thousand instances. 
	
	Copula processes are another alternative to deal with input noise in the context of regression tasks and GPs \citep{wilson2010copula}.
	In this case, approximate inference is carried out using the Laplace approximation and Markov chain Monte 
	Carlo methods. There also exists in the literature an approach for online heteroscedastic GP regression that tackles the 
	incorporation of new measurements in constant run-time and makes the computation cheaper by considering online sparse GPs
	\citep{bijl2017system}. This approach has proven to be effective in a practical applications considering, \emph{e.g.}, 
	system identification. 
	
	The work of \cite{mchutchon_gaussian_2011} has been employed in several practical applications involving machine learning
	regression problems. For example, in a problem concerning driving assistant systems \citep{armand2013modelling}, where the velocity 
	profile that the driver follows is modeled as the vehicle decelerates towards a stop intersection. Another example application can be 
	found in the context of  Bayesian optimization \citep{nogueira2016unscented, oliveira2017bayesian}, where a GP models an objective function 
	that is being optimized. In this scenario, the input space is contaminated with i.i.d Gaussian noise. Two real applications are considered: 
	safe robot grasping and safe navigation under localization uncertainty \citep{nogueira2016unscented,oliveira2017bayesian}.
	
	Another approach that considers input noise in the context of regression problems in arbitrary models is that of \cite{bocsi2013hessian}.
	This approach corrects the bias caused by the integration of the noise. The correction is proportional to the Hessian of the learned model 
	and to the variance of the input noise. The advantage of the method is that it works for arbitrary regression models and the disadvantage 
	is that it does not improve prediction for high-dimensional problems, where the data are implicitly scarce, and the estimated Hessian 
	is considerably flattened. 
	
	Input dependent noise has also been taken into account in the context of binary classification with GPs by \cite{hernandez_lobato2014}. 
	In particular, these authors describe the use of an extra GP to model the variance of additive Gaussian noise around a latent function 
	that is used for binary classification. This latent function is modeled again using a GP. Importantly, in this work the level of noise is 
	expected to depend on privileged information. These are extra input attributes that are only available at training time, but not 
	at test time. The goal is to exploit that privileged information to obtain a better classifier during the training phase. Approximate 
	inference is done in this case using expectation propagation instead of variational inference \citep{Minka01}.
	The experiments carried out show that privileged information is indeed useful to obtain better classifiers based on GPs.
	
	Importantly, in all these related works involving output noise that depends on the input, the domain of application is different 
	from ours since only regression or binary classification problems are addressed. Furthermore, it is assumed a 
	level of correlation between the location of the input features and the amount of extra output noise. This relation 
	between input features and the extra output noise can be arbitrarily complicated, since it is most of the times modeled via another GP.  In $\text{NIMGP}_\text{FO}$ the level of extra output noise not only depends on the input location of the
	instance, but also on the variance of the input noise associated to that particular instance. Moreover, 
	the expression for the dependence of the output noise with respect to the input is fixed (it cannot be learned from the 
	data) and is given by the second derivative of the GP. See Eq. (\ref{eq:fo_noise_influence}) for further details. 
	This expression is just obtained as a consequence of modeling in an approximate way the input noise.
	
	Random inputs have also been considered in the context of GPs for un-supervised learning in the Bayesian GP 
	latent variable model (GP-LVM) \citep{titsias2010bayesian,damianou2016variational}. In this model the 
	observed data is assumed to be obtained using a non-linear mapping applied to a set of latent variables. 
	The non-linear mapping is modeled using GPs and the latent variables are estimated using 
	a Bayesian approach. This involves setting a prior distribution for their potential values 
	and using variational inference (VI) to compute an approximate posterior distribution. 
	This framework is very similar to the VI approach we describe for the proposed method NIMGP.
	The main difference, however, between NIMGP and the Bayesian GP-LVM is that the latter can 
	only address unsupervised learning problems and it requires tractable Gaussian likelihood factors.
	A generalization of the Bayesian GP-LVM to account for partially observed input attributes is described 
	by \cite{damianou2015semi}. The resulting framework also resembles the VI approach described for the 
	NIMGP method. However, in that case, the approximate distribution $q$ over the latent attributes is
	constrained to be a Gaussian with fixed mean and adjustable variance, unlike the approximate distribution
	in Eq. (\ref{eq:post_approx}), which includes adjustable means. Again, this model requires tractable 
	Gaussian likelihood factors and cannot be used to address multi-class classification problems, like NIMGP.
	
	A generalization of the Bayesian GP-LVM considers a concatenation of non-linear transformations of the 
	latent features using GPs. This leads to deep GPs for unsupervised learning \citep{damianou2013deep}. 
	The applicability of such a method is, however, restricted to small data sets partially because of the 
	large number of variational parameters that grows linearly with the size of the data set. This problem 
	can be alleviated by augmenting the model with a recognition system that constraints the variational 
	posterior over the latent variables of each layer \citep{dai2015variational}. The recognition system 
	is given by a neural network that amortizes the approximate distribution $q$, as in the proposed 
	method $\text{NIMGP}_\text{NN}$. The main difference is that, although the extension to 
	address regression problems is straight-forward, it is not clear how to use such a method for classification problems. 
	
	To tackle input noise in the context multi-class classification, when one does does not rely on the use of GPs, some decomposition strategies can be used. Specifically, it is possible to decompose the problem into several binary classification subproblems, reducing the 
	complexity and, hence, dividing the effects caused by the noise into each of the subproblems \citep{saez2014analyzing}. 
	There exist several of these decomposition strategies, being the one-vs-one scheme a method that can be applied for 
	well known binary classification algorithms. The results obtained by these authors show that the one-vs-one decomposition leads to better performances and more robust classifiers than other decompositions. A problem of these decompositions, however, is that they can lead to ambiguous regions in which it is not clear what class label should be predicted \citep{bishop2006}. Furthermore, they do not learn an underlying true multi-class classifier and rely on binary classifiers to solve the multi-class problem, which is expected to be sub-optimal.  
	
	Random input attributes that follow a Gaussian distribution have been considered in the context of GP regression in \citet{girard2003gaussian}. That work addresses the problem 
	of learning in such a setting by using a Gaussian approximation that matches the mean and the variance of 
	the GP predictive distribution when the input attributes are noisy. The required computations are 
	tractable for some particular covariance functions such as the squared exponential. 
	Such an approach has also been applied in the context of solving control problems using GPs \citep{deisenroth2011pilco}. In that setting, however, the likelihood is a reinforcement 
	learning cost function that has been chosen carefully so that all computations remain 
	tractable. Moment matching to account for random inputs has 
	also been used  the context of deep Gaussian processes in which the input to a hidden layer is the output of the 
	previous layer, which is random \citep{bui2016deep}. An alternative approximation based on Kalman filters that is potentially more accurate than the one obtained by moment matching has also been considered in the literature in 
	the context of sequential state estimation \citep{ko2007gp} and in the context of GPs with arbitrary 
	non-linear likelihood functions \citep{steinberg2014extended}. Nevertheless, all these works differ from NIMGP 
	in that no inference is been made about the noiseless inputs. More precisely, the mean of the random 
	input attributes is assumed to be known, unlike in our setting, in which it is a latent 
	variable. Moreover, no multi-class classification problems have been specifically addressed.
	
	Solving regression problems in the context of noisy input observations has been addressed by 
	\cite{dallaire2009learning} using the moment-matching technique described in the previous paragraph.
	Accounting for noise in the inputs significantly improves the prediction results of a standard GP.
	In that work, however, the posterior distribution for the noiseless inputs is computed
	without conditioning to the observed targets. Note, however, that the likelihood factors $p(y_i|f_i)$ 
	constrain the potential values that the noiseless inputs can take. This is information is ignored when
	such an approach is used. By contrast, NIMGP is able to consider the label information by learning a 
	specific posterior approximation for each noiseless input, as described in Eq.  (\ref{eq:post_approx_orig}). 
	This posterior approximation is inferred at the same time as the other latent variables of the model. 
	The consequence is that the approach suggested by \cite{dallaire2009learning} is expected to be suboptimal. 
	Furthermore, the variance associated to the inputs is assumed to be given. We show that this parameter 
	can be inferred from the observed data in NIMGP and $\text{NIMGP}_\text{NN}$. 
	Moreover, no classification problems are addressed by those authors.
	
	Another work that has addressed noisy inputs only for test data in the context of GPs is 
	\citet{oakley2002bayesian}. Given the distribution of a test input, the resulting predictive 
	distribution is approximated using Monte Carlo methods. More precisely, samples from the posterior 
	process are generated approximately and the quantities of interest (\emph{e.g.}, the expected value 
	at the test point) are approximated by sampling from the distribution of the test inputs. The advantage 
	of such a method is that it can take into account arbitrary distributions for the noisy inputs. 
	The limitation is, however, that only regression problems are addressed, the training 
	inputs are assumed to be noiseless, and the distribution of the noisy test inputs is assumed 
	to be known beforehand.
	
	Noisy inputs have also been considered in the context of distribution regression in which the 
	inputs of the regression problem are data sets and/or probability distributions \citep{law2018bayesian}.
	In this case, a kernel mean embedding can be used a \emph{summary} or feature representation of each data set 
	\citep{muandet2017kernel}. This feature representation can be estimated from empirical 
	data, and the variance of the estimation is expected to be reduced 
	with the size of each data set. In the referred work it is described how to
	account for uncertainty in the estimation of the kernel mean embedding in the context of regression 
	problems using a Bayesian approach. A sophisticated method based on Hamilton Monte Carlo (HMC) is 
	suggested to approximate the required computations. As described previously, this work only considers 
	regression problems, assumes that the distribution of the inputs is known before hand and, moreover, 
	it could be limited by the expensive cost of running a Markov Chain using HMC. 
	
	Noisy inputs have also been consider in the context of other methods not directly related to GPs.
	More precisely, other approaches to deal with input noise involve using robust features in the context 
	of multi-class SVMs \citep{rabaoui2008using} or enhancements of fuzzy models \citep{ge2007dependency}.
	The first work can be understood as a pre-processing step in which robust features are generated and used for training with the
	goal of reducing the effect of background noise in a sound recognition system. These robust features are, however, specific of 
	the application domain conspired. Namely, sounds recognition. They are not expected to be useful in other classification problems.
	The second work focuses exclusively on linear regression models and hence cannot address multi-class classification problems,
	as the ones considered in our work.

	\section{Experiments}
	\label{sec:experiments}
	
	In this section we carry out several experiments to evaluate the performance of the proposed method for multi-class GP classification with
	input noise. More precisely, we compare the performance of a standard multi-class Gaussian process that does not consider 
	noise in the inputs (MGP) and the three proposed methods. Namely, the approach described in Section \ref{sec:nimgp} 
	(NIMGP), the variant described in Section \ref{sect:avi} where the parameters of the Gaussian posterior approximation are computed using a neural network ($\text{NIMGP}_\text{NN}$) and the method proposed in 
	Section \ref{sect:foa} ($\text{NIMGP}_\text{FO}$), which is based on the work of \cite{mchutchon_gaussian_2011}, and uses 
	a first order approximation to account for input noise. The experiments considered 
	include both in synthetic and real data. All the experiments carried out (except those related to the MNIST data set) involve 
	100 repetitions and we report average results. These are detailed in the following sections.
	
	All the methods described have been implemented in Tensorflow \citep{tensorflow2015}. The source code to reproduce
	all the experiments carried out is available online at \url{https://github.com/cvillacampa/GPInputNoise}.
	In these experiments, for each GP,  we have employed a squared exponential covariance function with automatic relevance 
	determination \citep{rasmussen2005book}. All hyper-parameters, including the GP amplitude parameter, the 
	length-scales and the level of additive Gaussian noise have been tuned by maximizing the ELBO.
	The class noise level has been set equal to $10^{-3}$, and kept fixed during training as in \citet{hensman2015}.
	Unless indicated differently, we have set the number of inducing points for the sparse Gaussian process to the minimum of $100$ and 
	$5\%$ of the total number of points. For the optimization of the ELBO we have used the ADAM optimizer 
	with learning rate equal to $0.01$, the number of epochs has been set to  $750$ and the mini-batch 
	size to $50$ \citep{Kingma2015}. This number of epochs seems to guarantee the convergence of the optimization process. 
	All other ADAM parameters have been set equal to their default value. 
	In $\text{NIMGP}_\text{NN}$ the neural network has $50$ hidden units and one hidden layer. 
	The activation function is set to be ReLu.  Finally, the number of Monte Carlo samples used 
	to approximate the predictive distribution in NIMGP and $\text{NIMGP}_\text{NN}$ is set to $300$.
	
	\subsection{Illustrative Toy Problem}
	\label{sec:toy}
	
	Before performing a fairly complete study on more realistic examples, we show here the results 
	of the three proposed methods on a one-dimensional synthetic data set with three classes. 
	This data set is simple so that it can be analyzed in detail and the optimal predictive distribution
	provided by the Bayes classifier can be computed in closed-form.
	The data set is generated by sampling the latent functions from the 
	GP prior using specific hyper-parameter values and then applying the 
	labeling rule in (\ref{eq:labelling_rule}). The input locations have been generated randomly 
	by drawing from a uniform distribution in the $[-3,3]$ interval. Then, we add a Gaussian 
	noise to each observation $\bf{x}_i$ to generate $\tilde{\mathbf{x}}_i$, with standard 
	deviation $\mathbf{V}_i=0.1\mathbf{I}$ for $i=1,\ldots,N$. We consider $1,000$ training instances and the number 
	of inducing points $M=100$. The mini-batch size is set to $200$ points. Since in this experiment the variance 
	of the input noise is known beforehand we do not infer its value from the observed data and rather 
	specify its actual value in each method. The number of test points is $1,000$.
	In $\text{NIMGP}_\text{NN}$ the neural network is set to have 2 hidden layers with $50$ units each.
	
\begin{figure}[t]
		\begin{center}
			\begin{tabular}{cc}
				\includegraphics[width=0.48\textwidth]{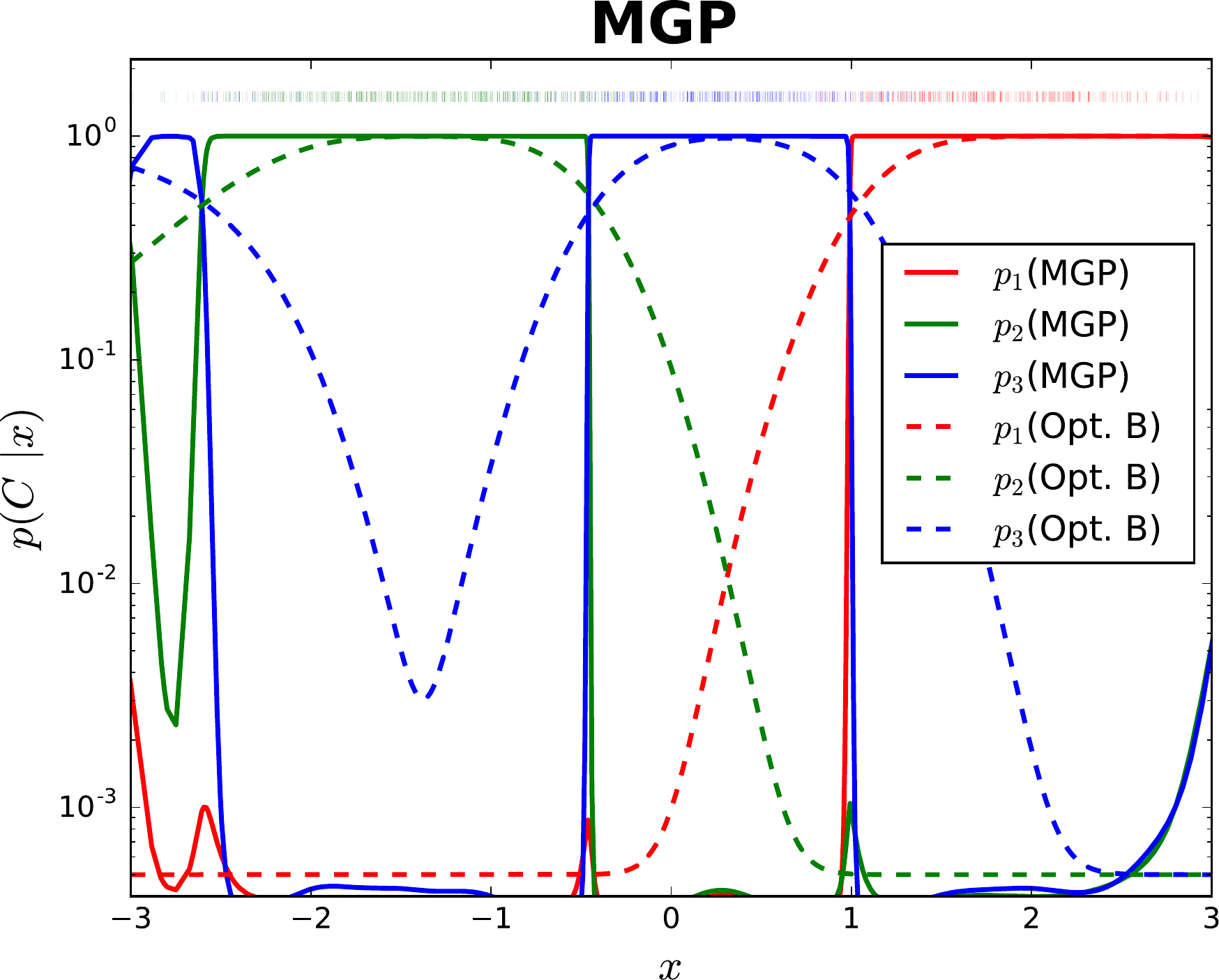} &
				\includegraphics[width=0.48\textwidth]{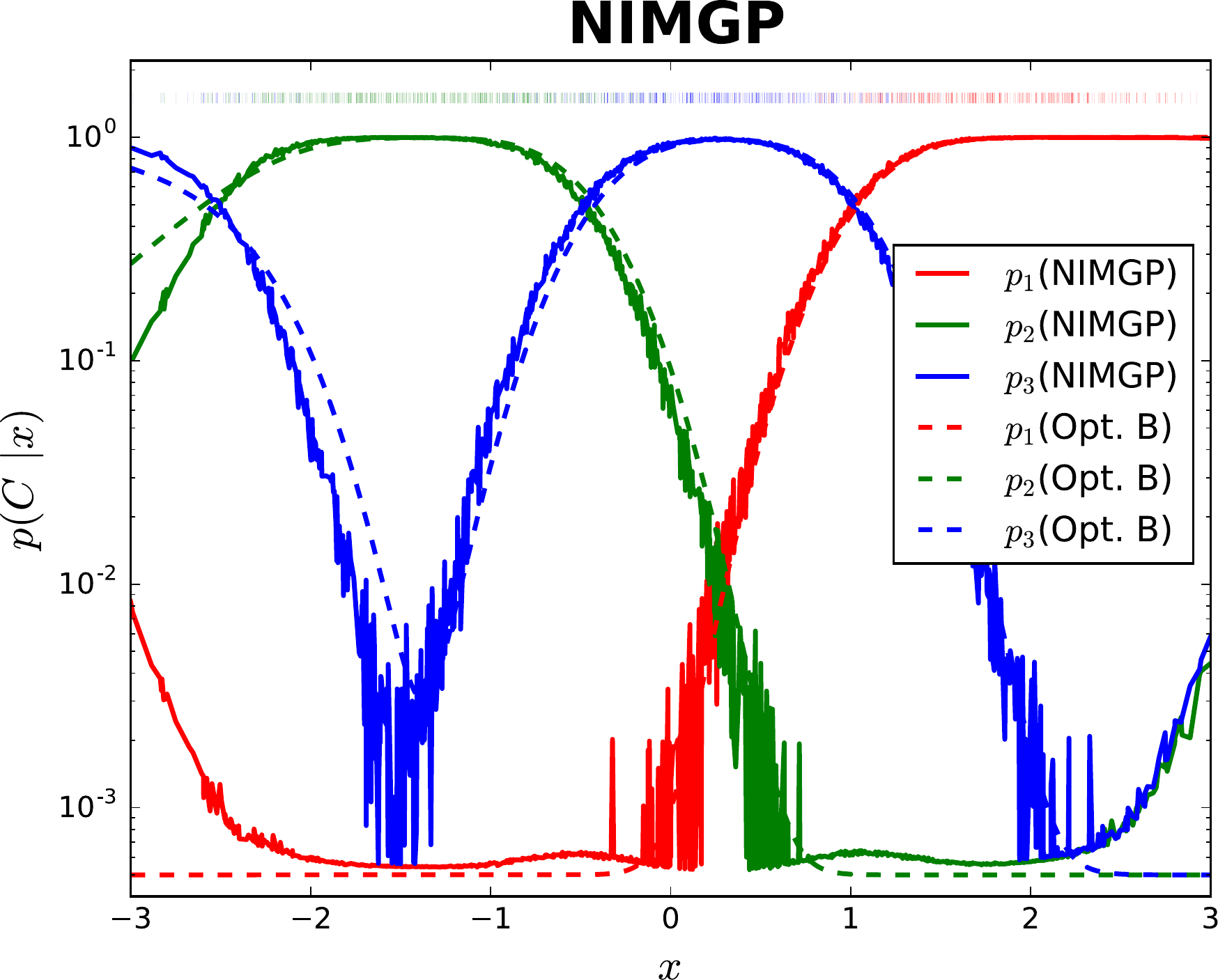} \\
				\includegraphics[width=0.48\textwidth]{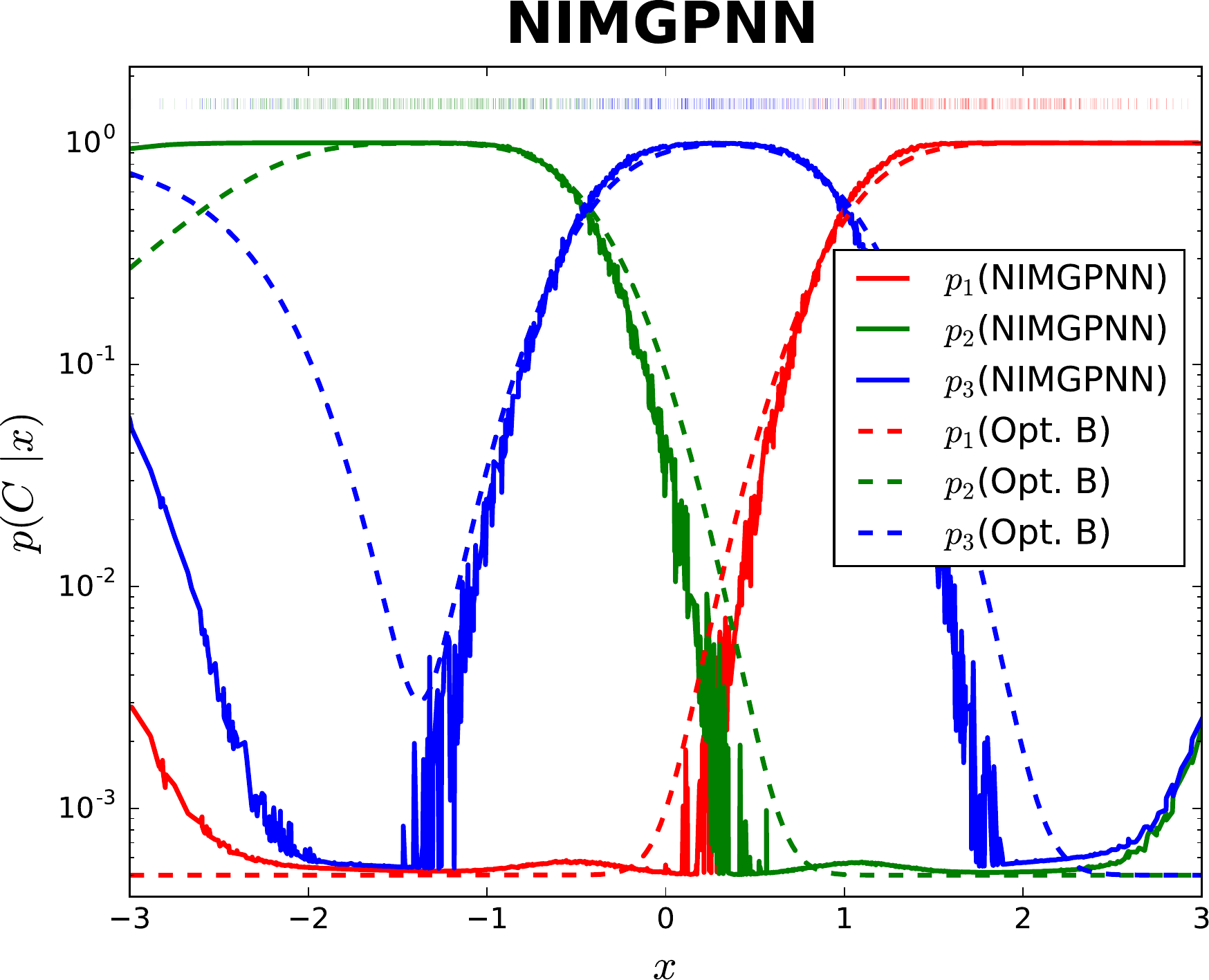} &
				\includegraphics[width=0.48\textwidth]{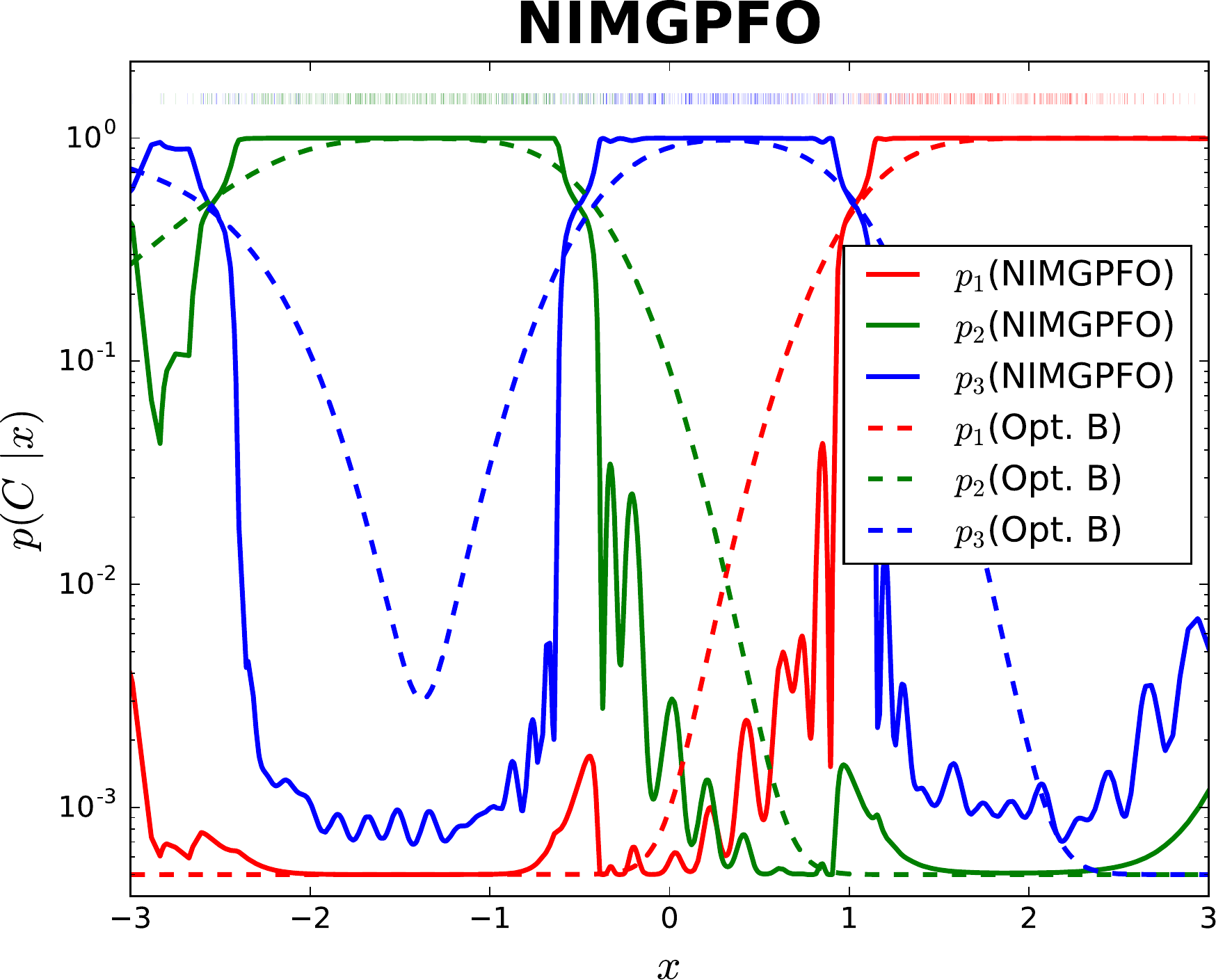} \\
			\end{tabular}
			\caption{\small Comparison of the predictive distribution computed by each method and the optimal Bayes prediction (shown as a dashed line). 
				The observed labels are shown as small vertical bars at the top of the plot in green, blue and red. 
				(Top-left) Model without input noise (MGP), (top-right) NIMGP model, (bottom-left) $\text{NIMGP}_\text{NN}$ and (bottom-right) $\text{NIMGP}_\text{FO}$.
				While the decision boundaries of each method agree with the optimal ones, NIMGP is the method that more accurately computes the predictive probabilities of each 
				class label. By contrast MGP produces predictions that are too confident.}
			\label{fig:pred_dist}
		\end{center}
	\end{figure}

	We have trained each method on this data set and compared the resulting predictive distribution with that of the 
	optimal Bayes classifier, which we can compute in closed form since we know the generating process of the labels. 
	Figure \ref{fig:pred_dist} shows, for each method, as a function of the input value $x$, the predicted distribution for each 
	of the three classes\footnote{The wiggles in the 
		prediction probability for some of the classes for the NIMGP and $\text{NIMGP}_\text{NN}$ are produced by 
		the Monte Carlo approximation of the predictive distribution. Here we use a Monte 
		Carlo average across 700 samples.}. The observed labels for each data point are 
	shown at the top of each figure as small vertical bars in green, blue and 
	red colors, depending on the class label. 
	We observe that each method produces decision boundaries that agree with the optimal ones (\emph{i.e.}, all methods predict the 
	class label that has the largest probability according to the optimal Bayes classifier).
	However, the predictive distributions produced differ from one method to another.
	More precisely, the first method, \emph{i.e.}, MGP (top-left), which ignores the input noise,
	does produce a predictive distribution that is significantly different from the optimal one, 
	especially in regions of the input space that are close to the decision boundaries. 
	This method produces a predictive distribution that is too confident.
	The closest predictive distribution to the optimal one is obtained by NIMGP (top-right), followed by $\text{NIMGP}_\text{NN}$ (bottom-left). Finally,
	$\text{NIMGP}_\text{FO}$ (bottom-right) seems to improve the results of MGP, but is far from NIMGP.
	
	Figure \ref{fig:pred_dist} shows very clearly the advantage of including the input noise when estimating the predictive 
	distribution $p(c|x)$ for each class label $c=1,2,3$, given a new test point $x$. Indeed, the proposed models 
	reproduce more closely the optimal predictive distribution of the Bayes classifier. By contrast, the model that 
	ignores the input noise, \emph{i.e.}, MGP fails to produce an accurate predictive distribution. Therefore, we expect
	to obtained better results for the proposed methods in terms of the log-likelihood of the test labels.
	
	Table \ref{tab:1d-data} shows the prediction error of each method and the corresponding negative test log-likelihood (NLL). 
	Standard deviations are estimated using the bootstrap.  Importantly,
	NLL can be understood as a measure of the quality of the predictive distribution. The smaller the better. We observe that while the 
	prediction error of each method is similar (since the decision boundaries produced are similar) the NLL of the proposed methods is 
	significantly better than the one provided by MGP, the method that ignores input noise. These results highlight the importance of accurately 
	modeling the input noise to obtain better predictive distributions, which can play a critical role when one is interested in the 
	confidence on the decisions to be made in terms of the classifier output.
	
	\begin{table}[h]
		\centering
		\begin{tabular}{c|c|c|c|c}
			\hline
			& {\bf MGP} & {\bf NIMGP} & {\bf $\text{NIMGP}_\text{NN}$} & {\bf $\text{NIMGP}_\text{FO}$} \\
			\hline
			Test error & {\bf 0.125$\pm$0.011} & $0.129 \pm 0.011$& {\bf 0.125$\pm$0.010} & $0.128\pm 0.010$\\
			Test NLL & $0.884\pm 0.079$ & {\bf 0.286$\pm$0.017} & $0.347\pm 0.030$ & $0.495\pm 0.046$ \\ 
			\hline
		\end{tabular}
		\caption{Test error and negative test log-likelihood (NLL) for the one-dimensional toy experiment.}
		\label{tab:1d-data}
	\end{table}
	
	\subsection{Synthetic Experiments}
	\label{sec:synt_exp}
	
\begin{figure}[t]
		\begin{center}
			\includegraphics[width=0.8\textwidth]{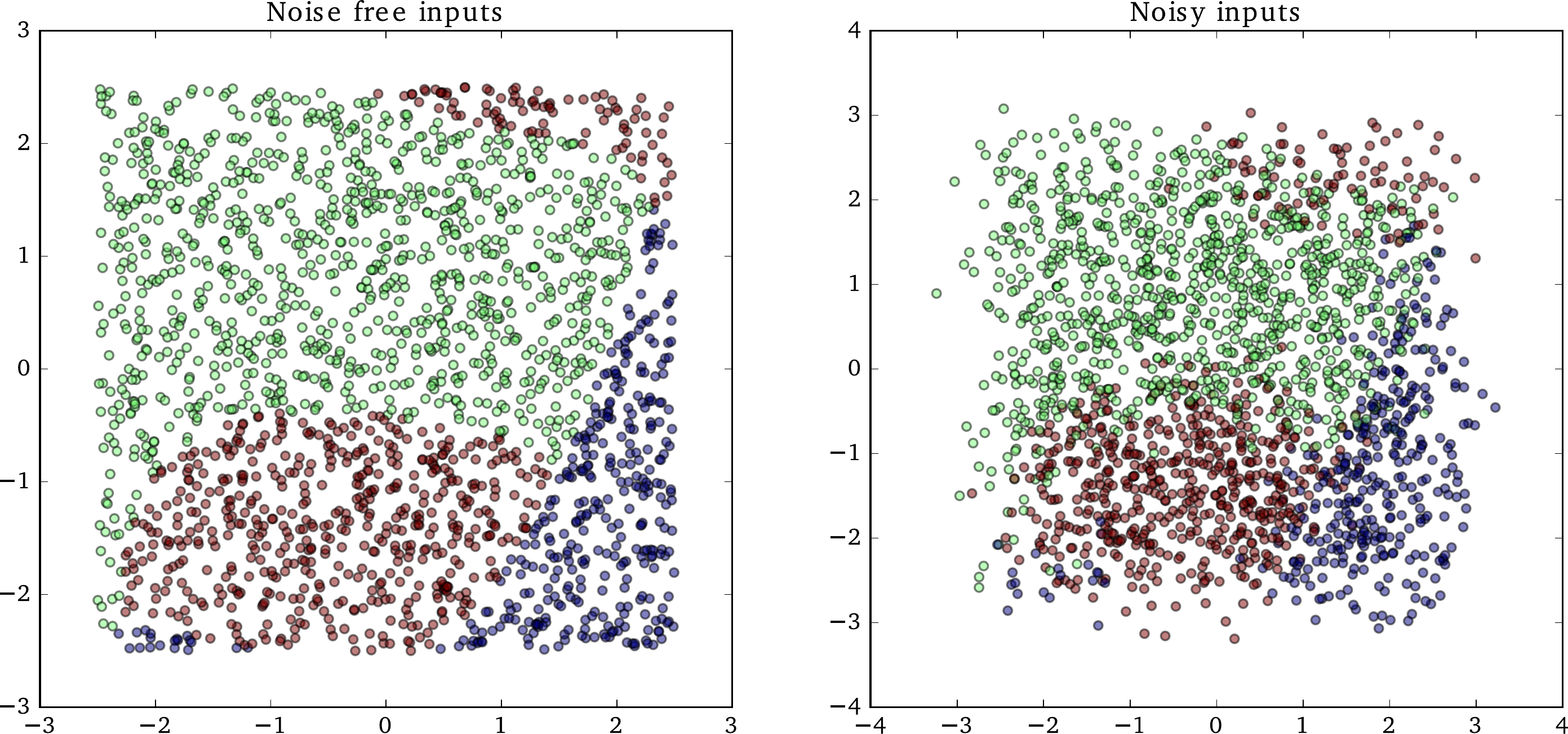} 
			\caption{\small Sample synthetic classification problem with three class labels. (left) Input data and associated class label before the noise injection in the 
				observed attributes. (right) Input data and associated class labels after the noise injection. The variance of the Gaussian noise is set equal to $0.1$. Best seen in color.}
			\label{fig:synthetic}
		\end{center}
	\end{figure}

	Next, we compare the methods on $100$ synthetic two-dimensional classification problems with three classes. As in the case of the 
	one-dimensional data set described above, these problems are generated by sampling the latent functions from a GP 
	prior with the squared exponential function and then applying the labeling rule in (\ref{eq:labelling_rule}). 
	The GP hyper-parameters employed are $\sigma^2=0.5$, $\sigma_0^2=0$, $\ell_j=2, \forall j$.
	The input vectors $\mathbf{x}_i$ are chosen uniformly in the box $[-2.5,2.5]^2$.
	Then, we add three different levels of random noise to each observation 
	$\mathbf{x}_i$, \emph{i.e.}, $\mathbf{V}_i= \{0.1\mathbf{I}, 0.25 \mathbf{I}, 0.5 \mathbf{I}\}$. 
	The interest of these experiments is to evaluate the proposed methods in a controlled setting in which the
	expected optimal model to explain the observed data is a multi-class GP classifier and we can control 
	the level of input noise in the data. In the next section we carry out experiments with real data sets.
	The number of training and test instances, inducing points, 
	mini-batch size and parameters of the ADAM optimizer are the same as in the previous experiment. 
	Again, since the level of injected noise is known in these experiments, we directly codify this information in each method.
	Figure \ref{fig:synthetic} shows a sample data set before and after the noise injection.
	
	The average results obtained on the 100 data sets, for each method, are displayed in Tables \ref{tab:toy_ll} and \ref{tab:toy_err}.
	We report results when the variance of the input noise is given beforehand (given), 
	and when it is learned from the observed data (learned). 
	In this last case, the lower bound given by Eq. (\ref{eq:elbo}) is maximized to estimate the 
	input noise variance, as described in Section \ref{sec:learning_noise}.
	We assume the same variance of the Gaussian input noise for each attribute.
	
	In the first scenario (given), we observe that, in terms of the negative test log-likelihood, all the 
	proposed methods improve over MGP, \emph{i.e.}, the standard GP multi-class classifier that 
	ignores input noise. Among the proposed methods, the best performing one is NIMGP,
	closely followed by $\text{NIMGP}_\text{NN}$. Therefore, these experiments highlight the 
	benefits of using a neural network to compute the parameters of the posterior approximation 
	$q(\mathbf{x}_i)$. Specifically, there is no performance degradation. 
	The method that is based on the first order approximation, 
	\emph{i.e.}, $\text{NIMGP}_\text{FO}$, also improves over MGP, but the differences are smaller.
	In terms of the prediction error the differences are much smaller. However, in spite of this, the proposed methods
	improve the results of MGP. Among them, again NIMGP is the best method, except when $\mathbf{V}_i=0.5\mathbf{I}$. In that case,
	$\text{NIMGP}_\text{FO}$ performs best, but the differences are very small. Summing up, the results obtained agree with
	the ones obtained in the one-dimensional problem. Namely, the proposed approaches significantly improve the quality of the
	predictive distribution in terms of the neg. test log-likelihood. The prediction error is, however, similar.
	
\begin{figure}[tb]
		\begin{center}
			\begin{tabular}{|l|c|c|c|}
				\hline
				& {\bf  NIMGP } & {\bf $\text{NIMGP}_\text{NN}$ } & {\bf $\text{NIMGP}_\text{FO}$ } \\
				\hline
				\rotatebox{90}{\bf\hspace{0.45cm} Noise 0.1}  &
				\includegraphics[width=0.2\textwidth]{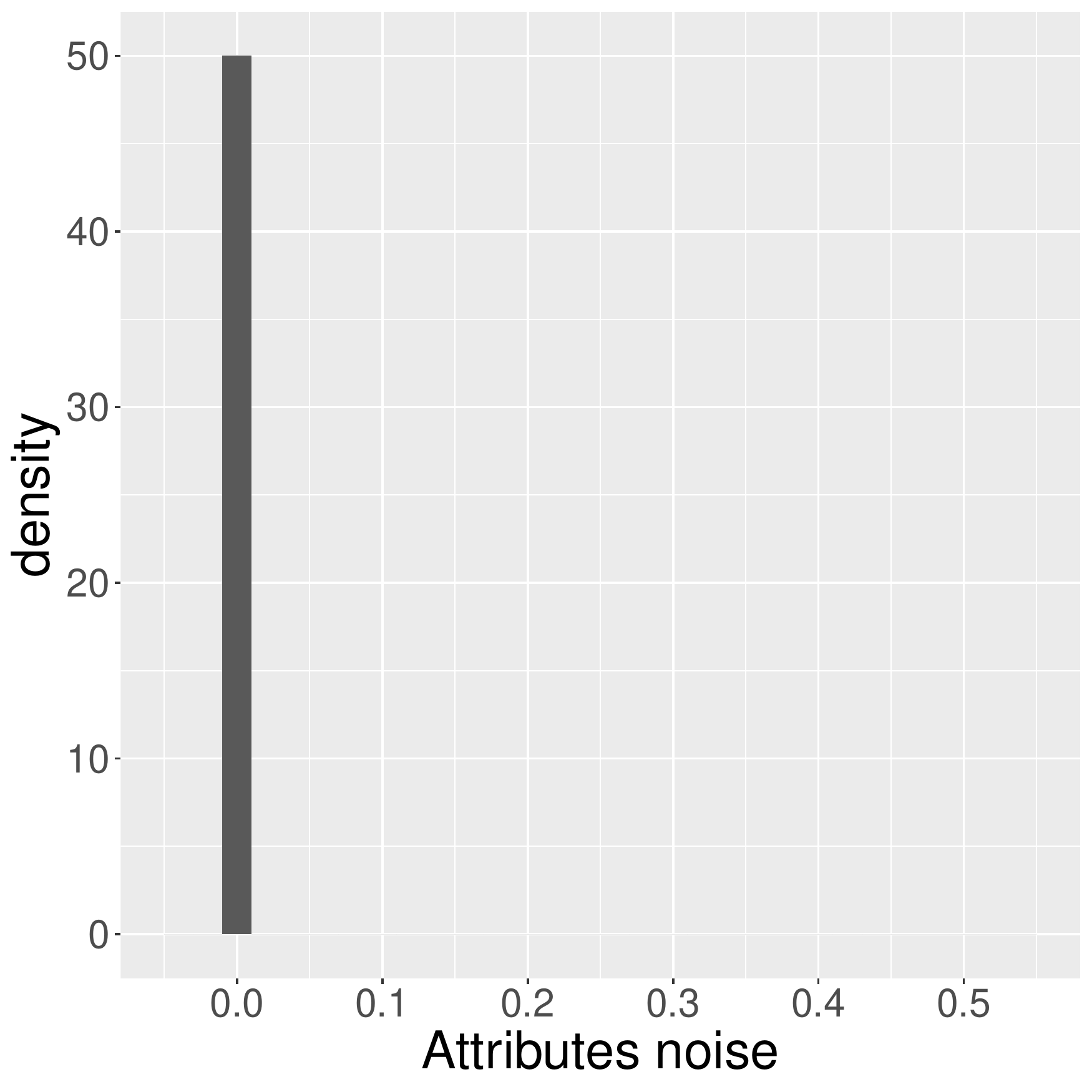} &
				\includegraphics[width=0.2\textwidth]{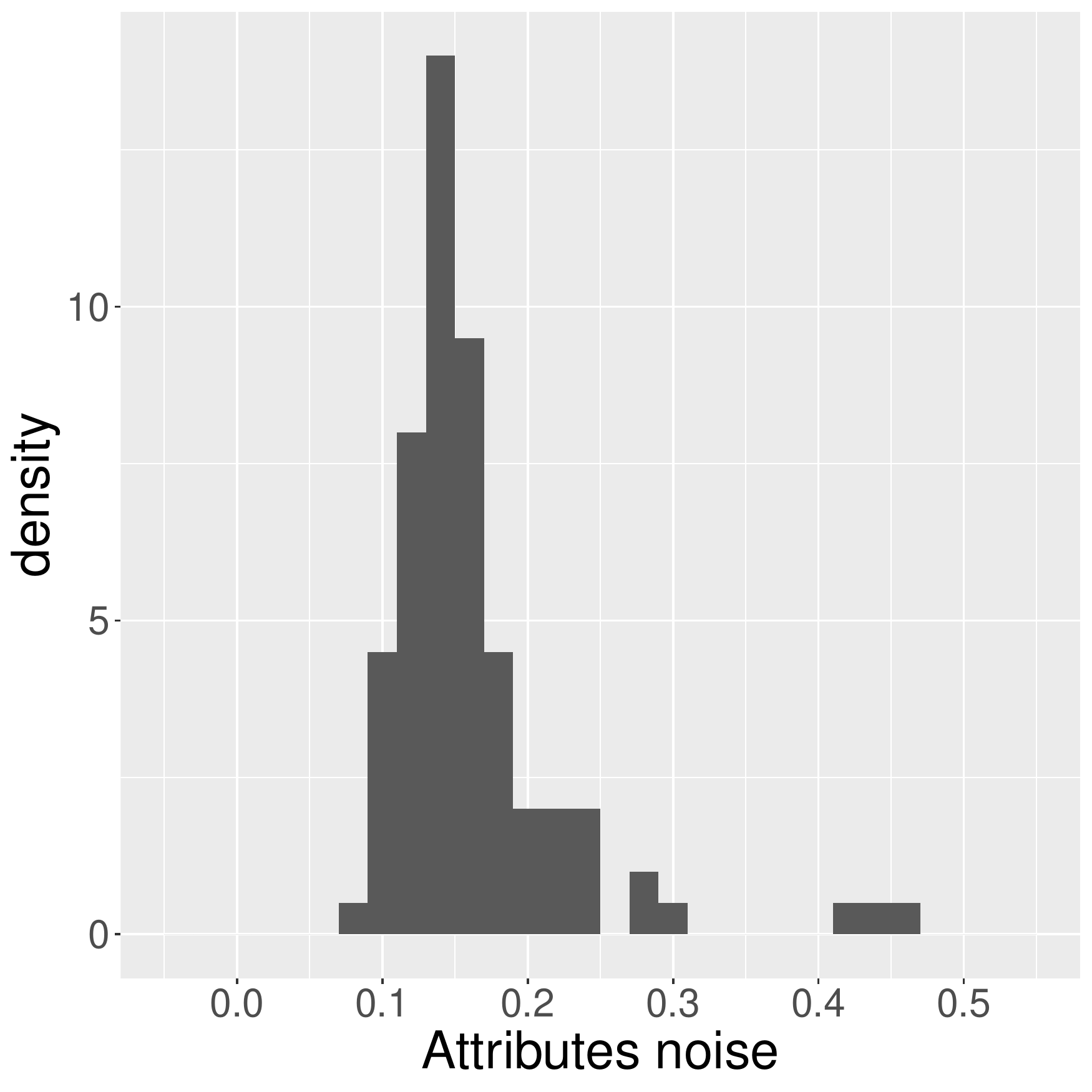} &
				\includegraphics[width=0.2\textwidth]{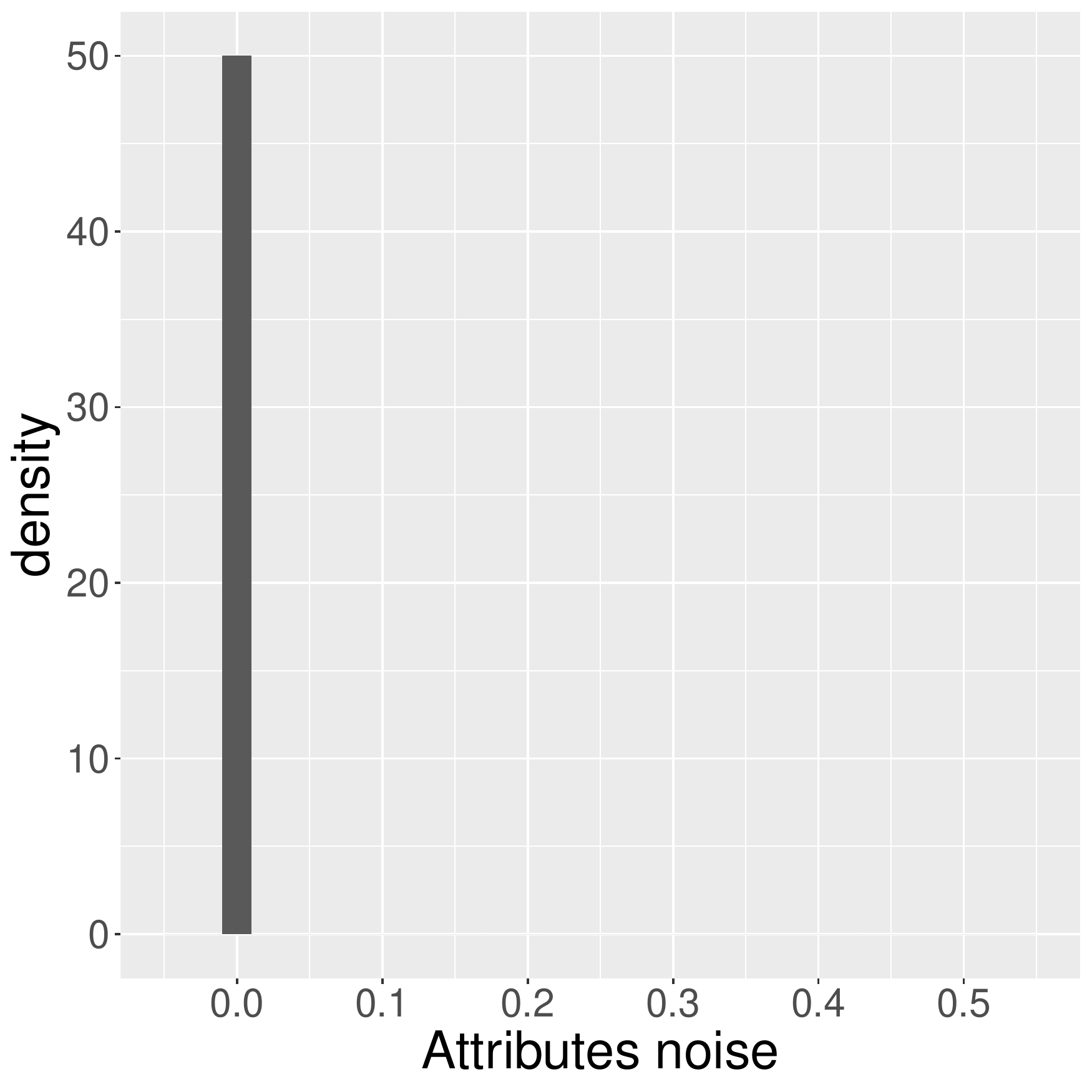} \\
				\hline
				\rotatebox{90}{\bf\hspace{0.35cm} Noise 0.25}  &
				\includegraphics[width=0.2\textwidth]{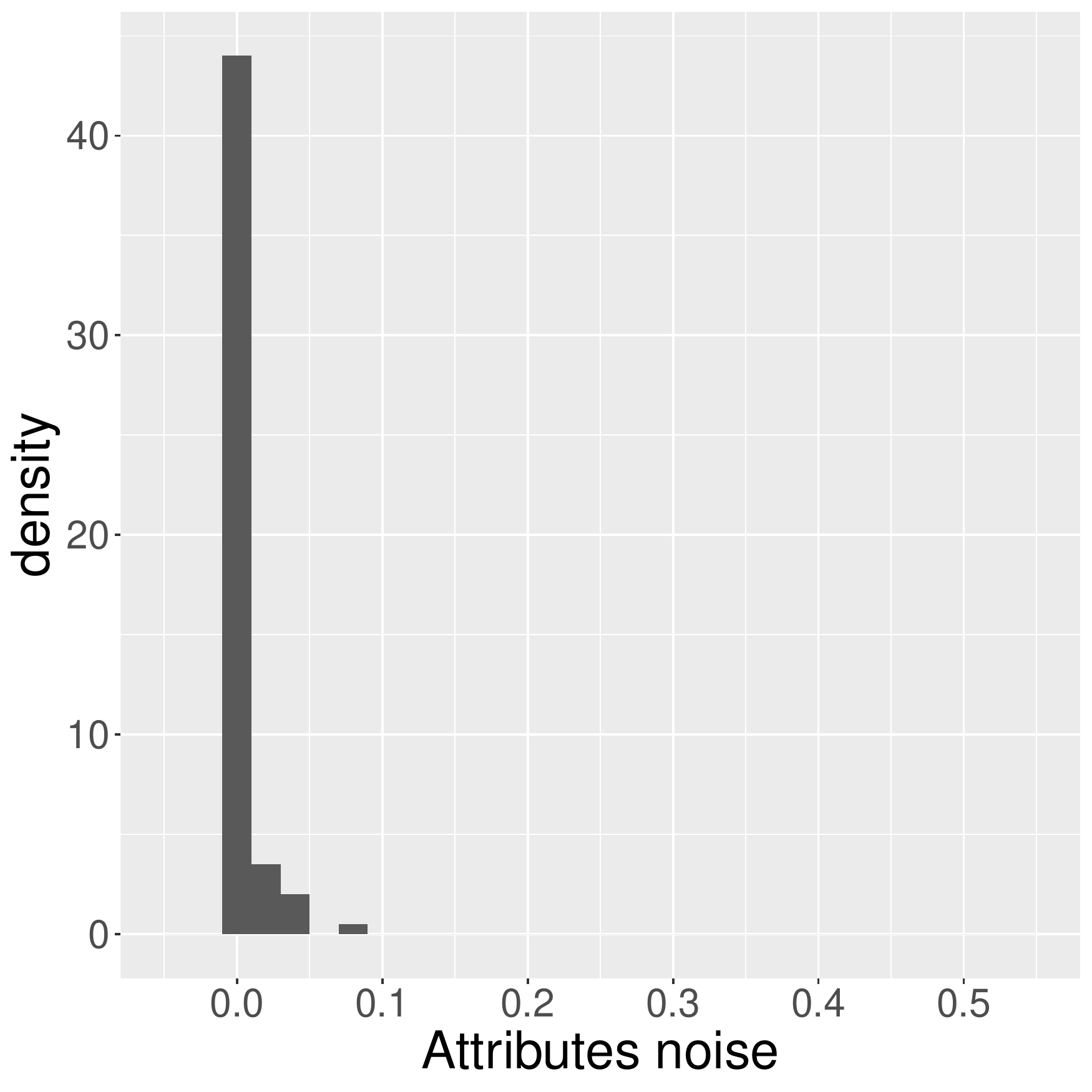} &
				\includegraphics[width=0.2\textwidth]{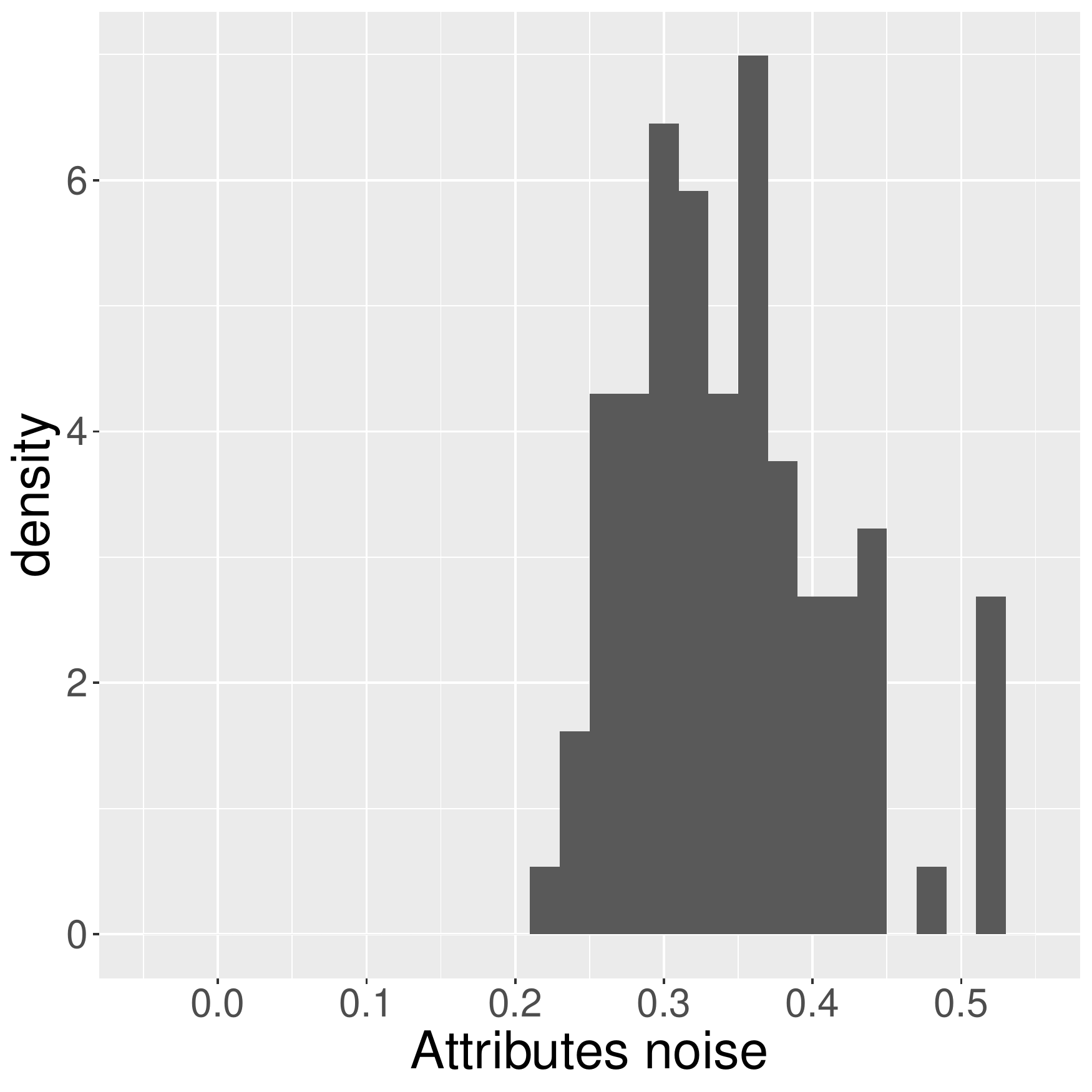} &
				\includegraphics[width=0.2\textwidth]{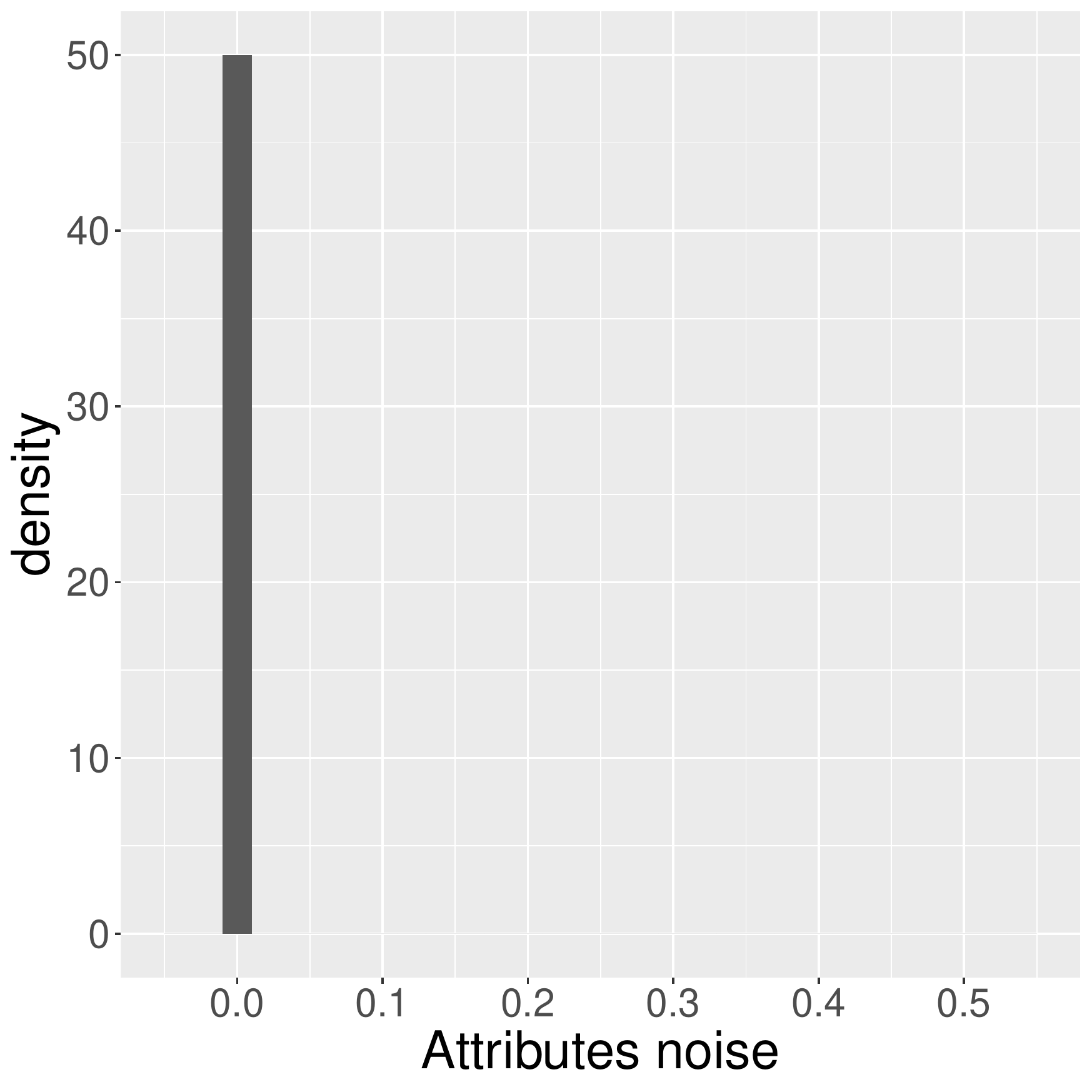}\\
				\hline
				\rotatebox{90}{\bf\hspace{0.45cm} Noise 0.5}  &
				\includegraphics[width=0.2\textwidth]{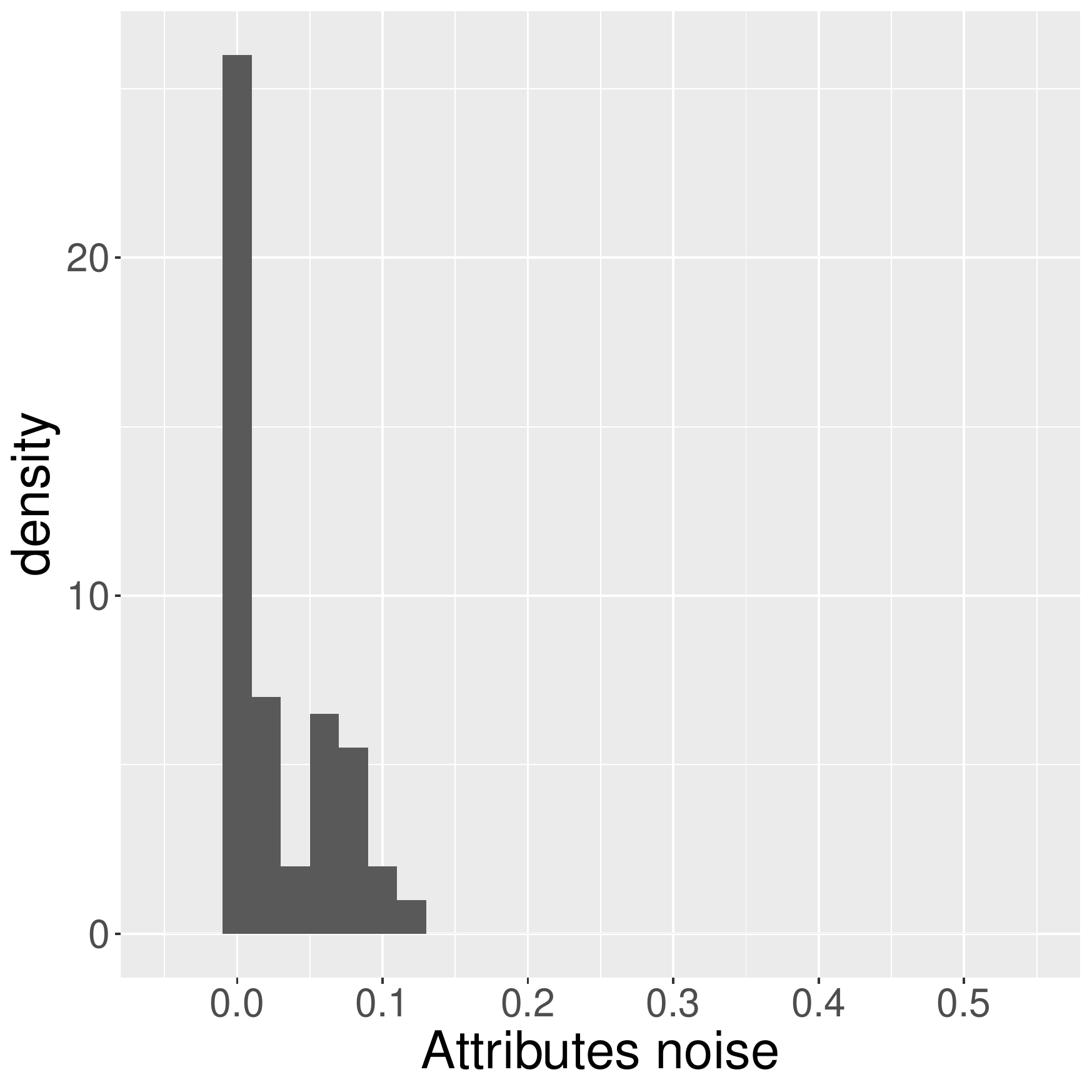} &
				\includegraphics[width=0.2\textwidth]{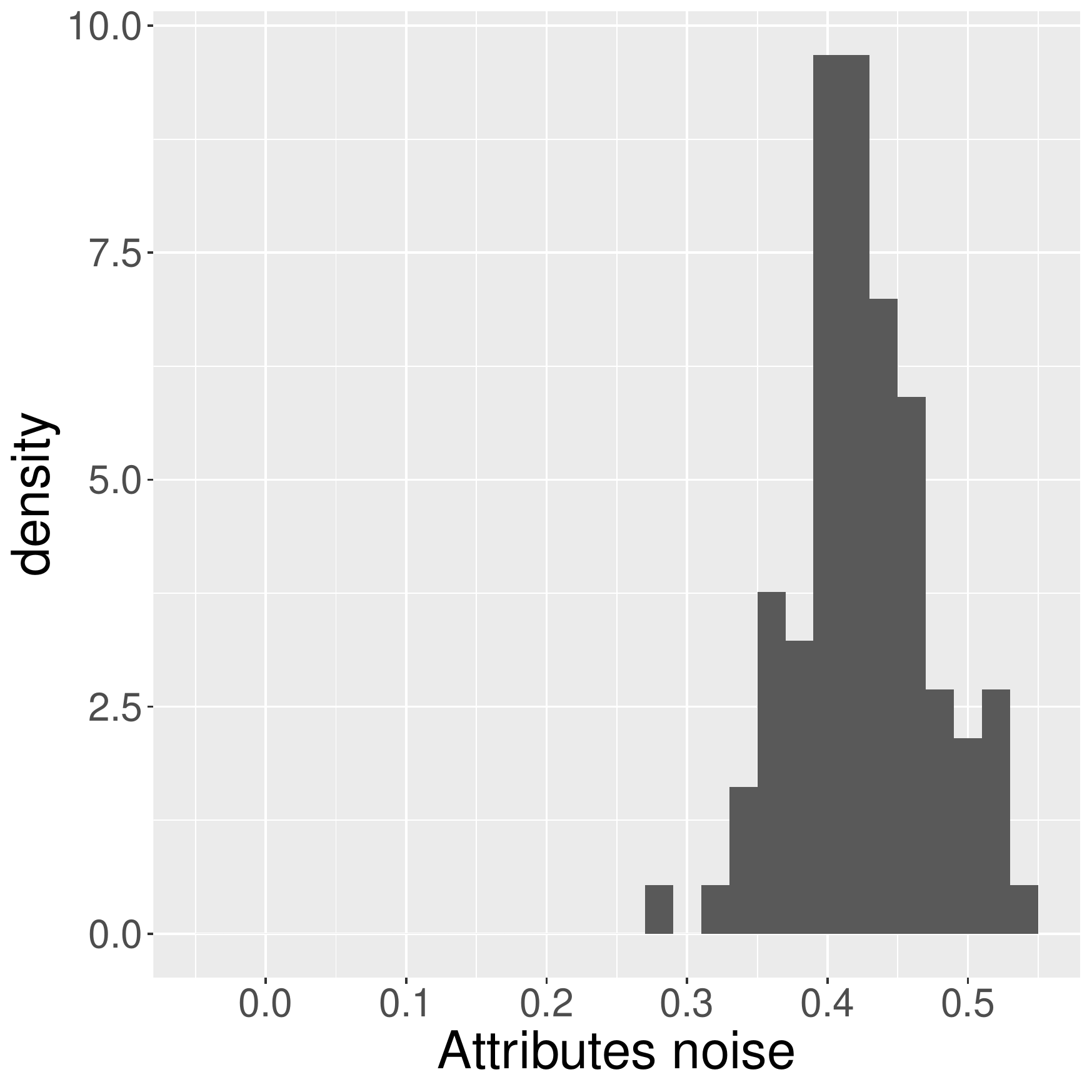} &
				\includegraphics[width=0.2\textwidth]{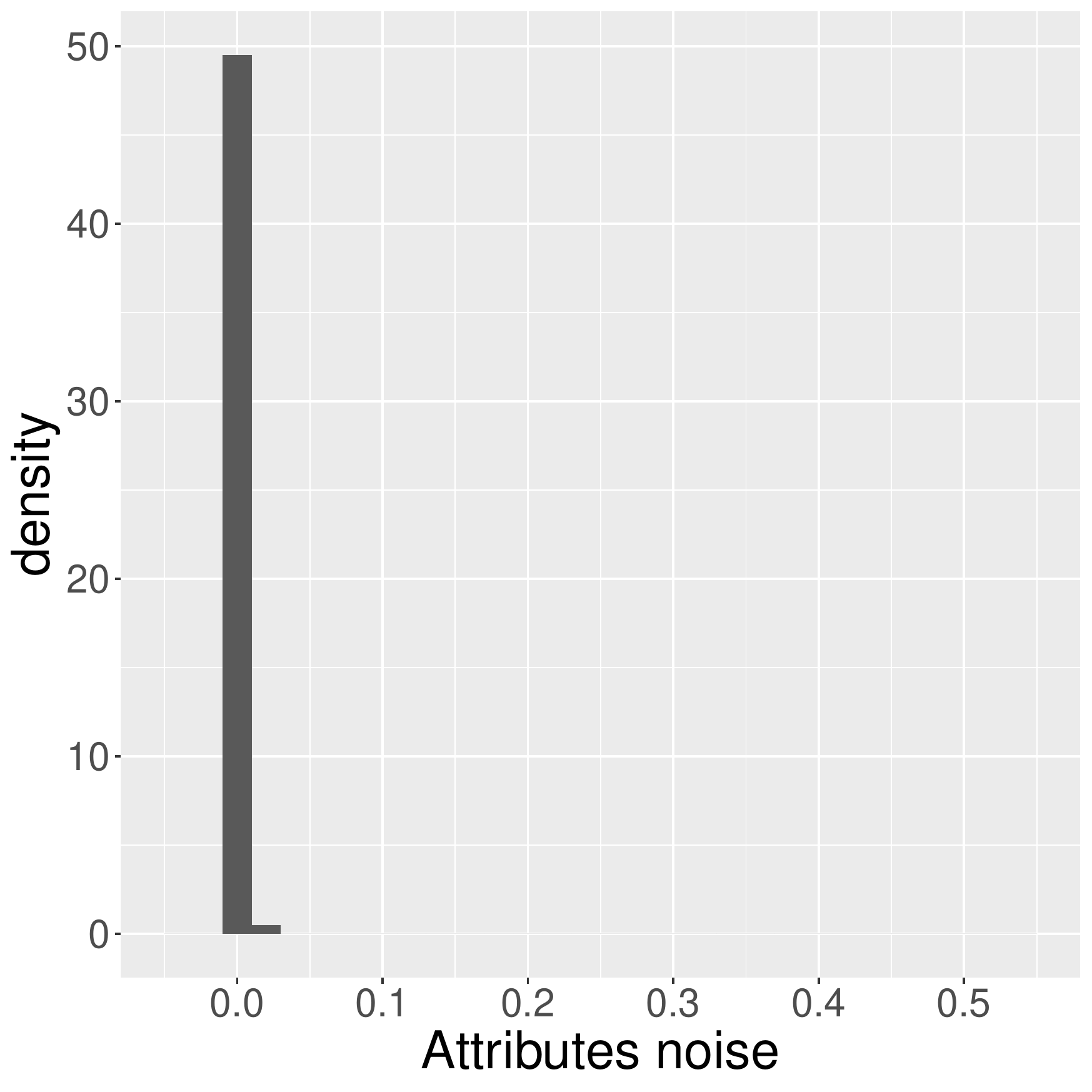} \\
				\hline
				
			\end{tabular}
		\end{center}
		\caption{Histogram of the estimated values (across the 100 experiments) for the variance of the input 
			noise, for each method. Note that $\text{NIMGP}_\text{NN}$ is the only method producing
			sensible estimates of the input noise variance in each case. The other methods most of the 
			times estimate values for the noise variance that are almost zero.}
		\label{fig:hist_estimations_synthetic}
	\end{figure}

	In the second scenario, \emph{i.e.}, when the variance of the input noise is learned from the data, 
	we observe that $\text{NIMGP}_\text{FO}$ performs more or less similarly to MGP, and that 
	NIMGP and $\text{NIMGP}_\text{NN}$ obtain better results in terms of the test log-likelihood, 
	particularly in the case of the second method. In terms of the prediction error 
	all the methods give similar results, as in the previous scenario. 
	The reason for these results is that $\text{NIMGP}_\text{NN}$ seems to be the only 
	method that can provide sensible estimates of the input noise variance from the observed data. 
	This is confirmed by Figure \ref{fig:hist_estimations_synthetic}, which shows a histogram of 
	the input noise variances inferred by each method across the 100 repetitions of the experiments.  
	Note that NIMGP and $\text{NIMGP}_\text{FO}$ seem to always 
	estimate a variance level that is close to zero. 
	By contrast, $\text{NIMGP}_\text{NN}$ estimates an input noise variance that is very similar to the 
	actual one. This is an unexpected observation. We believe that the better results obtained by 
	$\text{NIMGP}_\text{NN}$ over NIMGP can be a consequence of the regularization properties 
	introduced by the neural network employed in this method. This is confirmed by the experiments carried 
	out in the next section. The bad results of $\text{NIMGP}_\text{FO}$ could be related to the 
	linear approximations introduced in this method.
	
	\begin{table}[t]
		\centering
		\begin{tabular}{lc|r@{$\pm$}lr@{$\pm$}lr@{$\pm$}lr@{$\pm$}l}
			\hline 
			\multicolumn{2}{c}{\bf Noise} & \multicolumn{2}{c}{ {\bf MGP} }&\multicolumn{2}{c}{ {\bf NIMGP} }&\multicolumn{2}{c}{\bf  $\text{NIMGP}_\text{NN}$ }&\multicolumn{2}{c}{\bf  $\text{NIMGP}_\text{FO}$ }\\
			\hline
			\multirow{3}{*}{\rotatebox{90}{Given}} & 
			0.1 &  0.758 & 0.022 &  \bf  0.0256 & \bf 0.007 & 0.0265 & 0.008 & 0.321 & 0.009\\
			&  0.25 & 1.14 & 0.033 & \bf  0.0369 & \bf 0.011 & 0.0388 & 0.012 & 0.53 & 0.015\\
			&  0.5 &  1.537 & 0.041 & \bf  0.0493 & \bf 0.012  & 0.0526 & 0.014 & 0.77 & 0.021 \\
			\hline
			\multirow{3}{*}{\rotatebox{90}{Learned}} & 
			0.1 &  0.758  &  0.022 &  0.61  &  0.016 &  \bf 0.28  & \bf 0.0087 &  0.72  &  0.021\\
			&  0.25 &  1.14  &  0.033 &  0.87  &  0.021 &  \bf 0.4  &  \bf 0.012 &  1.1  &  0.032\\
			&  0.5 &  1.537  &  0.04 &  0.93  &  0.017 &  \bf 0.53  &  \bf 0.014 &  1.5  &  0.041\\
			\hline
		\end{tabular}
		\caption{Average Neg. Test Log Likelihood for each method on the synthetic problems when the variance of the input noise
			is given and learned from the data.}
		\label{tab:toy_ll}
	\end{table}

	\begin{table}[t]
		\centering
		\begin{tabular}{lc|r@{$\pm$}lr@{$\pm$}lr@{$\pm$}lr@{$\pm$}l}
			\hline 
			\multicolumn{2}{c}{\bf Noise} & \multicolumn{2}{c}{\bf MGP }&\multicolumn{2}{c}{\bf  NIMGP }&\multicolumn{2}{c}{\bf $\text{NIMGP}_\text{NN}$ }&\multicolumn{2}{c}{\bf $\text{NIMGP}_\text{FO}$ }\\
			\hline
			\multirow{3}{*}{\rotatebox{90}{Given}} & 
			0.1 &  0.113 & 0.003 & \bf  0.108 &  \bf 0.003 & 0.108 & 0.003 & 0.109 & 0.003\\
			& 0.25 & 0.164 & 0.006 & \bf  0.158 & \bf 0.005 & 0.158 & 0.005 & 0.158 & 0.005\\
			& 0.5 &  0.218 & 0.006 & 0.21 & 0.058 & 0.22 & 0.063 & \bf  0.21 & \bf 0.006\\
			\hline
			\multirow{3}{*}{\rotatebox{90}{Learned}} & 
			0.1 &  0.113  &  0.003 &  \bf 0.11  &  \bf 0.003 &  \bf 0.11  &  \bf 0.003 &  0.11  &  0.003\\
			& 0.25 &  0.164  &  0.006 &  0.17  &  0.005 &  0.17  &  0.005 &  \bf 0.16  & \bf 0.005\\
			& 0.5 &  0.218  &  0.006 &  0.23  &  0.006 &  \bf 0.21  &  \bf 0.006 &  0.22  &  0.006\\
			\hline
		\end{tabular}
		\caption{Average  Test Error for each method on the synthetic problems when the variance of the input noise
			is given and learned from the data.}
		\label{tab:toy_err}
	\end{table}
	
	In Appendix \ref{sec:appendix_extra_experiments} we report extra experimental results on the synthetic data sets described
	using other configurations (\emph{i.e.}, varying the number of classes, the number of dimensions and the
	number of training instances). We also report results for two other baselines. These are (i) using 
	MGP and augmenting the training set with samples from the posterior distribution of 
	each data instance assuming a uniform prior; and (ii) using MGP but sampling the input attributes for 
	each mini-batch from the previous distribution. The results obtained for the other data configurations are similar 
	to the ones reported here. Moreover, the other baselines do not seem to improve the results of MGP at all, or 
	the improvements are significantly smaller the ones obtained by NIMGP, $\text{NIMGP}_\text{NN}$ and $\text{NIMGP}_\text{FO}$.
	
	\subsection{Experiments on data sets Extracted from the UCI Repository}
	\label{sec:exp_uci}
	
	Another set of experiments evaluates the proposed methods on $8$ 
	different multi-class data sets extracted from the UCI repository \citep{Dua2019}.
	Table \ref{tab:charac} displays the characteristics of the data sets considered. 
	For each of these data sets, we consider $100$ splits into train and
	test, containing 90\% and 10\% of the data respectively. Unlike in the previous experiments, in 
	these data sets, a GP multi-class classifier need not be optimal. Furthermore, the
	input attributes may already be contaminated with additive noise. 
	To assess the benefits of considering such noise during the learning process, we 
	have considered four different setups. In each one, we inject Gaussian noise in 
	the observed attributes with different variances. Namely, $0.0$, $0.1$, $0.25$
	and $0.5$. This will allow to evaluate the different methods in a setting in which 
	there may be or may be not input noise due to the particular characteristics of the problem addressed
	(\emph{i.e.} when the variance of the injected noise is equal to $0.0$), and also for increasing levels
	of input noise (\emph{i.e.}, when the variance of the injected noise is equal to $0.1$, $0.25$ and $0.5$). 
	This noise injection process is done after a standardization step in which the observed 
	attributes are normalized to have zero mean and unit variance in the training set.
	All methods are trained for $1,000$ epochs using a mini-batch size of $50$. In these 
	experiments, in each of the proposed methods, the level of noise is 
	learned during the training process by maximizing the ELBO. The reason for this is that
	the actual level of noise in the input attributes need not be equal to the injected 
	level of noise.
	
	\begin{table}[b]
		\begin{center}
			{\small
				\begin{tabular}{lccc}
					\hline
					{\bf data set} & {\bf \#Instances} & {\bf \#Attributes} & {\bf \#Classes} \\
					\hline
					Glass & 214 & 9 & 6 \\
					New-thyroid & 215 & 5 & 3 \\
					Satellite & 6435 & 36 & 6 \\
					Svmguide2 & 391 & 20 & 3 \\
					Vehicle & 846 & 18 & 4 \\
					Vowel & 540 & 10 & 6 \\
					Waveform & 1000 & 21 & 3 \\
					Wine & 178 & 13 & 3 \\
					\hline
				\end{tabular}
			}
		\end{center}
		\caption{Characteristics of the data sets extracted from the UCI repository.}
		\label{tab:charac}
	\end{table}
	
	Table \ref{tab:uci_ll} shows the average results obtained for each method in terms of the negative test 
	log-likelihood, for each level of noise considered. We also report the mean rank for each method across data sets and splits.
	If a method is always the best one, it will receive a mean rank equal to 1. Conversely, if a method is always worst, it will receive
	a mean rank equal to 4. Therefore, in general lower is better. We can see that on average, the proposed methods improve over MGP and 
	the method that works the best (according to the mean rank) is $\text{NIMGP}_\text{NN}$, even for the case where we do not introduce 
	noise in the inputs. This suggests that these data sets have already some noise in the inputs. Also, as 
	we increase the noise level the mean rank for MGP and NIMGP is worsen and $\text{NIMGP}_\text{NN}$ and $\text{NIMGP}_\text{FO}$ 
	both improve. The fact that $\text{NIMGP}_\text{NN}$ and $\text{NIMGP}_\text{FO}$ give better results than 
	NIMGP as we increase the level of noise indicates that the use of the neural network and the 
	first order approximation act as a regularizer with better generalization properties \citep{shu2018amortized}.
	
	Table \ref{tab:uci_err} shows the average results obtained for each method in terms of the prediction error.
	In this case, we do not observe big differences among the different methods. Moreover, the methods that take 
	into account the noise in the inputs do not improve the prediction error as we increase the noise level. 
	These small differences in terms of the error can be due to each method having similar decision boundaries, 
	even when we obtain better predictive distributions in terms of the test log-likelihood, as illustrated in Section \ref{sec:toy}. 
	
	\begin{table}[p]
		\begin{tabular}{c|lr@{$\pm$}lr@{$\pm$}lr@{$\pm$}lr@{$\pm$}l}
			\hline 
			& & \multicolumn{2}{c}{ MGP }&\multicolumn{2}{c}{ NIMGP }&\multicolumn{2}{c}{ $\text{NIMGP}_\text{NN}$ }&\multicolumn{2}{c}{ $\text{NIMGP}_\text{FO}$ }\\ 
			\hline 
			\rotatebox{90}{\hspace{-2.4cm}{Noise =  0.0 }}&  glass  & 1.63  &  0.048  & 1.28  &  0.04  & {\bf 1.17 } & {\bf  0.033 } & 1.19  &  0.033  \\ 
			&  new-thyroid  & 0.096  &  0.007  & {\bf 0.083 } & {\bf  0.006 } & 0.122  &  0.006  & 0.113  &  0.005  \\ 
			&  satellite  & 0.5  &  0.007  & 0.363  &  0.005  & {\bf 0.281 } & {\bf  0.002 } & 0.316  &  0.003  \\ 
			&  svmguide2  & 0.594  &  0.024  & 0.586  &  0.023  & {\bf 0.519 } & {\bf  0.018 } & 0.531  &  0.02  \\ 
			&  vehicle  & 0.638  &  0.019  & 0.514  &  0.019  & {\bf 0.408 } & {\bf  0.005 } & 0.497  &  0.006  \\ 
			&  vowel  & 0.415  &  0.025  & {\bf 0.278 } & {\bf  0.019 } & 0.321  &  0.012  & 0.445  &  0.015  \\ 
			&  waveform  & 0.676  &  0.017  & 0.657  &  0.015  & {\bf 0.335 } & {\bf  0.006 } & 0.451  &  0.01  \\ 
			&  wine  & {\bf 0.054 } & {\bf  0.004 } & 0.056  &  0.004  & 0.074  &  0.004  & 0.065  &  0.004  \\ 
			\hline 
			& Mean rank  &  3.05  &  0.0429 &  2.39  &  0.0377 &  \bf 2.05  &  \bf 0.0486 &  2.51  &  0.0295\\
			\hline
			\rotatebox{90}{\hspace{-2.4cm}{Noise =  0.1 }}&  glass  & 1.71  &  0.051  & 1.91  &  0.055  & {\bf 1.33 } & {\bf  0.034 } & 1.37  &  0.041  \\ 
			&  new-thyroid  & 0.278  &  0.025  & 0.303  &  0.026  & {\bf 0.19 } & {\bf  0.011 } & 0.201  &  0.013  \\ 
			&  satellite  & 0.703  &  0.008  & 0.613  &  0.007  & {\bf 0.347 } & {\bf  0.003 } & 0.421  &  0.004  \\ 
			&  svmguide2  & 0.663  &  0.025  & 0.655  &  0.024  & {\bf 0.565 } & {\bf  0.018 } & 0.596  &  0.021  \\ 
			&  vehicle  & 1.25  &  0.027  & 1.28  &  0.026  & {\bf 0.612 } & {\bf  0.007 } & 0.795  &  0.014  \\ 
			&  vowel  & 0.836  &  0.026  & 0.815  &  0.025  & {\bf 0.656 } & {\bf  0.012 } & 0.673  &  0.018  \\ 
			&  waveform  & 0.752  &  0.017  & 0.734  &  0.016  & {\bf 0.381 } & {\bf  0.006 } & 0.52  &  0.011  \\ 
			&  wine  & {\bf 0.087 } & {\bf  0.007 } & 0.089  &  0.007  & 0.097  &  0.006  & 0.093  &  0.007  \\ 
			\hline 
			& Mean rank  &  3.19  &  0.038 &  3.06  &  0.03 &  \bf 1.71  &  \bf 0.037 &  2.04  &  0.022\\
			\hline 
			\rotatebox{90}{\hspace{-2.5cm}{Noise =  0.25 }}&  glass  & 1.89  &  0.053  & 1.96  &  0.06  & {\bf 1.36 } & {\bf  0.034 } & 1.45  &  0.037  \\ 
			&  new-thyroid  & 0.445  &  0.035  & 0.472  &  0.034  & {\bf 0.271 } & {\bf  0.015 } & 0.302  &  0.02  \\ 
			&  satellite  & 0.835  &  0.009  & 0.77  &  0.008  & {\bf 0.409 } & {\bf  0.003 } & 0.472  &  0.004  \\ 
			&  svmguide2  & 0.761  &  0.025  & 0.756  &  0.025  & {\bf 0.627 } & {\bf  0.015 } & 0.638  &  0.019  \\ 
			&  vehicle  & 1.61  &  0.031  & 1.65  &  0.03  & {\bf 0.783 } & {\bf  0.008 } & 0.967  &  0.017  \\ 
			&  vowel  & 1.37  &  0.037  & 1.38  &  0.033  & 1.04  &  0.013  & {\bf 0.943 } & {\bf  0.017 } \\ 
			&  waveform  & 0.849  &  0.018  & 0.836  &  0.018  & {\bf 0.434 } & {\bf  0.006 } & 0.519  &  0.009  \\ 
			&  wine  & {\bf 0.134 } & {\bf  0.011 } & 0.136  &  0.011  & 0.15  &  0.008  & 0.141  &  0.008  \\ 
			\hline 
			& Mean rank  &  3.25  &  0.033 &  3.16  &  0.028 &  \bf 1.68  &  \bf 0.037 &  1.92  &  0.025\\
			\hline 
			\rotatebox{90}{\hspace{-2.5cm}{Noise =  0.5 }}&  glass  & 2.03  &  0.053  & 2.01  &  0.051  & {\bf 1.45 } & {\bf  0.034 } & 1.52  &  0.038  \\ 
			&  new-thyroid  & 0.565  &  0.038  & 0.623  &  0.04  & {\bf 0.369 } & {\bf  0.018 } & 0.381  &  0.021  \\ 
			&  satellite  & 0.973  &  0.009  & 0.932  &  0.009  & {\bf 0.491 } & {\bf  0.003 } & 0.531  &  0.004  \\ 
			&  svmguide2  & 0.877  &  0.025  & 0.878  &  0.025  & 0.706  &  0.013  & {\bf 0.702 } & {\bf  0.018 } \\ 
			&  vehicle  & 1.93  &  0.032  & 1.99  &  0.031  & {\bf 0.994 } & {\bf  0.006 } & 1.1  &  0.016  \\ 
			&  vowel  & 1.99  &  0.038  & 2.08  &  0.037  & 1.33  &  0.012  & {\bf 1.25 } & {\bf  0.018 } \\ 
			&  waveform  & 1.01  &  0.021  & 0.984  &  0.021  & {\bf 0.503 } & {\bf  0.006 } & 0.565  &  0.01  \\ 
			&  wine  & 0.253  &  0.017  & 0.264  &  0.017  & 0.24  &  0.01  & {\bf 0.236 } & {\bf  0.011 } \\ 
			\hline 
			& Mean rank  &  3.33  &  0.027 &  3.25  &  0.028 &  \bf 1.63  &  \bf 0.03 &  1.79  &  0.027\\
			\hline 
		\end{tabular} 
		\caption{Average neg. test log likelihood for the experiments on the UCI data sets.}
		\label{tab:uci_ll}
	\end{table} 
	
	\begin{table}[p]
		\begin{tabular}{c|lr@{$\pm$}lr@{$\pm$}lr@{$\pm$}lr@{$\pm$}l}
			\hline 
			& & \multicolumn{2}{c}{ MGP }&\multicolumn{2}{c}{ NIMGP }&\multicolumn{2}{c}{ $\text{NIMGP}_\text{NN}$ }&\multicolumn{2}{c}{ $\text{NIMGP}_\text{FO}$ }\\ 
			\hline 
			\rotatebox{90}{\hspace{-2.4cm}{Noise =  0.0 }}&  glass  & 0.346  &  0.008  & 0.387  &  0.01  & 0.375  &  0.009  & {\bf 0.345 } & {\bf  0.009 } \\ 
			&  new-thyroid  & 0.041  &  0.004  & {\bf 0.031 } & {\bf  0.004 } & 0.042  &  0.005  & 0.044  &  0.004  \\ 
			&  satellite  & 0.092  &  0.001  & {\bf 0.092 } & {\bf  0.001 } & 0.118  &  0.001  & 0.093  &  0.001  \\ 
			&  svmguide2  & 0.166  &  0.006  & {\bf 0.165 } & {\bf  0.006 } & 0.174  &  0.006  & {\bf 0.165 } & {\bf  0.006 } \\ 
			&  vehicle  & 0.16  &  0.004  & {\bf 0.155 } & {\bf  0.004 } & 0.216  &  0.004  & 0.159  &  0.004  \\ 
			&  vowel  & 0.07  &  0.004  & {\bf 0.054 } & {\bf  0.003 } & 0.096  &  0.004  & 0.068  &  0.004  \\ 
			&  waveform  & 0.155  &  0.003  & 0.153  &  0.003  & {\bf 0.14 } & {\bf  0.003 } & 0.155  &  0.003  \\ 
			&  wine  & 0.024  &  0.003  & 0.024  &  0.003  & {\bf 0.019 } & {\bf  0.003 } & 0.022  &  0.003  \\ 
			\hline 
			& Mean rank  &  2.37  &  0.0349 &  2.32  &  0.0393 &  2.92  &  0.0461 &  2.39  &  0.0326\\
			\hline 
			\rotatebox{90}{\hspace{-2.4cm}{Noise =  0.1 }}&  glass  & 0.389  &  0.009  & 0.412  &  0.01  & 0.413  &  0.009  & {\bf 0.387 } & {\bf  0.009 } \\ 
			&  new-thyroid  & 0.076  &  0.006  & 0.078  &  0.006  & {\bf 0.07 } & {\bf  0.005 } & 0.075  &  0.006  \\ 
			&  satellite  & 0.128  &  0.001  & {\bf 0.127 } & {\bf  0.001 } & 0.142  &  0.001  & 0.128  &  0.001  \\ 
			&  svmguide2  & 0.198  &  0.006  & 0.198  &  0.006  & {\bf 0.192 } & {\bf  0.006 } & 0.198  &  0.006  \\ 
			&  vehicle  & 0.291  &  0.005  & 0.295  &  0.005  & {\bf 0.291 } & {\bf  0.005 } & 0.291  &  0.005  \\ 
			&  vowel  & 0.217  &  0.005  & {\bf 0.217 } & {\bf  0.004 } & 0.246  &  0.005  & 0.218  &  0.005  \\ 
			&  waveform  & 0.162  &  0.003  & 0.163  &  0.003  & {\bf 0.154 } & {\bf  0.003 } & 0.162  &  0.003  \\ 
			&  wine  & 0.034  &  0.004  & 0.036  &  0.004  & {\bf 0.033 } & {\bf  0.004 } & 0.033  &  0.004  \\ 
			\hline 
			& Mean rank  &  2.44  &  0.036 &  2.51  &  0.04 &  2.64  &  0.0503 &  2.41  &  0.0357\\
			\hline 
			\rotatebox{90}{\hspace{-2.5cm}{Noise =  0.25 }}&  glass  & 0.433  &  0.009  & 0.456  &  0.01  & 0.467  &  0.01  & {\bf 0.432 } & {\bf  0.009 } \\ 
			&  new-thyroid  & 0.101  &  0.007  & 0.106  &  0.006  & {\bf 0.093 } & {\bf  0.006 } & 0.098  &  0.007  \\ 
			&  satellite  & 0.155  &  0.001  & {\bf 0.155 } & {\bf  0.001 } & 0.163  &  0.001  & 0.155  &  0.001  \\ 
			&  svmguide2  & 0.217  &  0.006  & 0.218  &  0.006  & 0.237  &  0.006  & {\bf 0.216 } & {\bf  0.006 } \\ 
			&  vehicle  & 0.357  &  0.006  & 0.366  &  0.006  & {\bf 0.351 } & {\bf  0.006 } & 0.36  &  0.006  \\ 
			&  vowel  & 0.351  &  0.007  & {\bf 0.345 } & {\bf  0.006 } & 0.432  &  0.008  & 0.349  &  0.007  \\ 
			&  waveform  & 0.193  &  0.003  & 0.193  &  0.004  & {\bf 0.188 } & {\bf  0.003 } & 0.193  &  0.003  \\ 
			&  wine  & 0.052  &  0.006  & 0.055  &  0.006  & {\bf 0.049 } & {\bf  0.005 } & 0.052  &  0.005  \\ 
			\hline 
			& Mean rank  &  2.39  &  0.0316 &  2.48  &  0.0377 &  2.71  &  0.0444 &  2.42  &  0.0356\\
			\hline 
			\rotatebox{90}{\hspace{-2.5cm}{Noise =  0.5 }}&  glass  & {\bf 0.47 } & {\bf  0.009 } & 0.501  &  0.01  & 0.51  &  0.011  & 0.473  &  0.01  \\ 
			&  new-thyroid  & 0.125  &  0.007  & 0.148  &  0.008  & 0.139  &  0.007  & {\bf 0.122 } & {\bf  0.007 } \\ 
			&  satellite  & 0.181  &  0.002  & {\bf 0.181 } & {\bf  0.002 } & 0.188  &  0.002  & 0.181  &  0.001  \\ 
			&  svmguide2  & 0.256  &  0.007  & {\bf 0.256 } & {\bf  0.007 } & 0.288  &  0.006  & {\bf 0.256 } & {\bf  0.007 } \\ 
			&  vehicle  & 0.428  &  0.006  & {\bf 0.422 } & {\bf  0.006 } & 0.445  &  0.005  & 0.424  &  0.006  \\ 
			&  vowel  & {\bf 0.473 } & {\bf  0.007 } & 0.478  &  0.008  & 0.565  &  0.007  & 0.475  &  0.007  \\ 
			&  waveform  & 0.225  &  0.004  & 0.225  &  0.004  & {\bf 0.222 } & {\bf  0.004 } & 0.225  &  0.004  \\ 
			&  wine  & 0.092  &  0.006  & 0.092  &  0.006  & 0.093  &  0.006  & {\bf 0.088 } & {\bf  0.007 } \\ 
			\hline 
			& Mean rank  &  2.35  &  0.0354 &  2.46  &  0.0396 &  2.91  &  0.0389 &  2.29  &  0.0353\\
			\hline 
		\end{tabular} 
		\caption{Average Test Error for experiments on UCI data sets}
		\label{tab:uci_err}
	\end{table} 
	
	\subsection{Experiments on the MNIST data set}

	In this section we consider a data set in which sparse GPs are needed in order to 
	train a multi-class classifier based on GPs. Namely, the MNIST data set \citep{lecun1998gradient}.
	This data set has 10 different class labels and $60,000$ training instances lying in a $28\times 28$ dimensional space.
	The test set has $10,000$ data instances. We consider a similar setup to the previous experiments and inject
	noise in the inputs (after a standardization to guarantee zero mean and unit standard deviation on the input
	attributes) with variances equal to $0.0$, $0.1$, $0.25$ and $0.5$. The level of noise is also
	learned during the training process by maximizing the ELBO. The mini-batch size is set to $200$ and
	the number of training epochs is set to $350$. This number of epochs seems to be large enough to guarantee the convergence of the 
	different methods evaluated. In the case of $\text{NIMGP}_\text{NN}$ the neural network 
	has 250 units and two hidden layers. We also slightly modified the neural network so that at the beginning of 
	the training process it outputs as the mean the noisy observed attributes $\tilde{\mathbf{x}}_i$ fed at 
	the input. The number of Monte Carlo samples used to approximate the predictive distribution 
	in NIMGP and $\text{NIMGP}_\text{NN}$ is set to $500$.  We use a bigger number of Monte Carlo samples in these 
	experiments because of the bigger size of the classification problem (10 classes) and the input 
	dimensionality, \emph{i.e.}, 784 dimensions. All the computations are sped-up by using a TESLA P100 GPU for training. 
	In these experiments we use a polynomial kernel with automatic relevance determination 
	since there is evidence supporting that it works better on this data set \citep{henao2012predictive}.
	
	The results obtained in these experiments, for each method, are displayed in Table \ref{tab:mnist}.
	The results obtained are similar to others reported in the literature \citep{henao2012predictive,hensman2015,villacampa2017}.
	We observe that the proposed methods always outperform the MGP, \emph{i.e.}, the multi-class GP that
	ignores the input noise. Both in terms of the test error and the negative test log-likelihood.
	In this case, however, the gains are small. We believe this is because this data set is particularly 
	challenging for GPs \citep{van2017convolutional}. 
	
	\begin{table}[t]
		\begin{center}
			\begin{tabular}{c|c|rccc}
				\hline
				{\bf Noise} & {\bf Metric} & {\bf MGP}  & {\bf NIMGP}  & {\bf $\text{NIMGP}_\text{NN}$} & {\bf $\text{NIMGP}_\text{FO}$} \\
				\hline
				\rotatebox{90}{\hspace{-.35cm}0.0} & NLL & 0.0649 & 0.0652 & {\bf 0.0630} & 0.0651 \\
				& Error &  0.0203 & 0.02 & {\bf 0.0196} & 0.0202  \\
				\hline
				\rotatebox{90}{\hspace{-.35cm}0.1} & NLL & 0.0718 & 0.0712 & {\bf 0.0704} & 0.0720 \\
				& Error &  {\bf 0.0215} & 0.0221 & {\bf 0.0215} & 0.0216  \\
				\hline
				\rotatebox{90}{\hspace{-.45cm}0.25} & NLL & 0.0878 & 0.0879 &  {\bf 0.0852} & 0.0880 \\
				& Error &  0.0265 & 0.0275 & {\bf 0.0253} & 0.0265 \\
				\hline
				\rotatebox{90}{\hspace{-.35cm}0.5} & NLL & 0.1023 & 0.1024 & {\bf  0.1005} & 0.1024 \\
				& Error &  0.029 & 0.0293 & {\bf 0.0284} & 0.0288 \\
				\hline
			\end{tabular} 
		\end{center}
		\caption{Average test error and neg. test log-likelihood (NLL) of each method on the MNIST data set.}
		\label{tab:mnist}
	\end{table}
	
\begin{figure}[tb]
		\begin{center}
			\begin{tabular}{|l|c|c|c|}
				\hline
				& {\bf  NIMGP } & {\bf $\text{NIMGP}_\text{NN}$ } & {\bf $\text{NIMGP}_\text{FO}$ } \\
				\hline
				\rotatebox{90}{\bf\hspace{0.45cm} Noise 0.1}  &
				\includegraphics[width=0.25\textwidth]{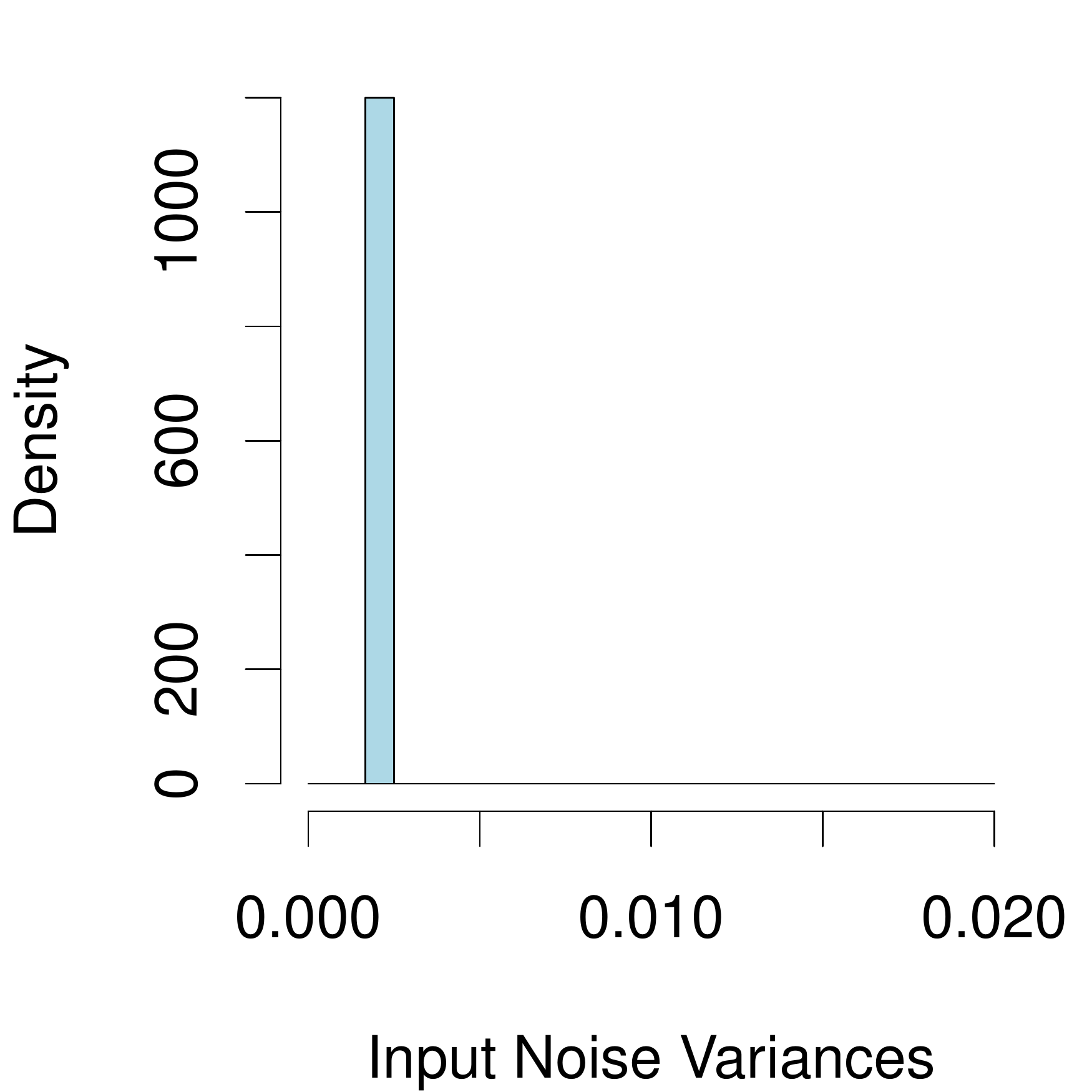} &
				\includegraphics[width=0.25\textwidth]{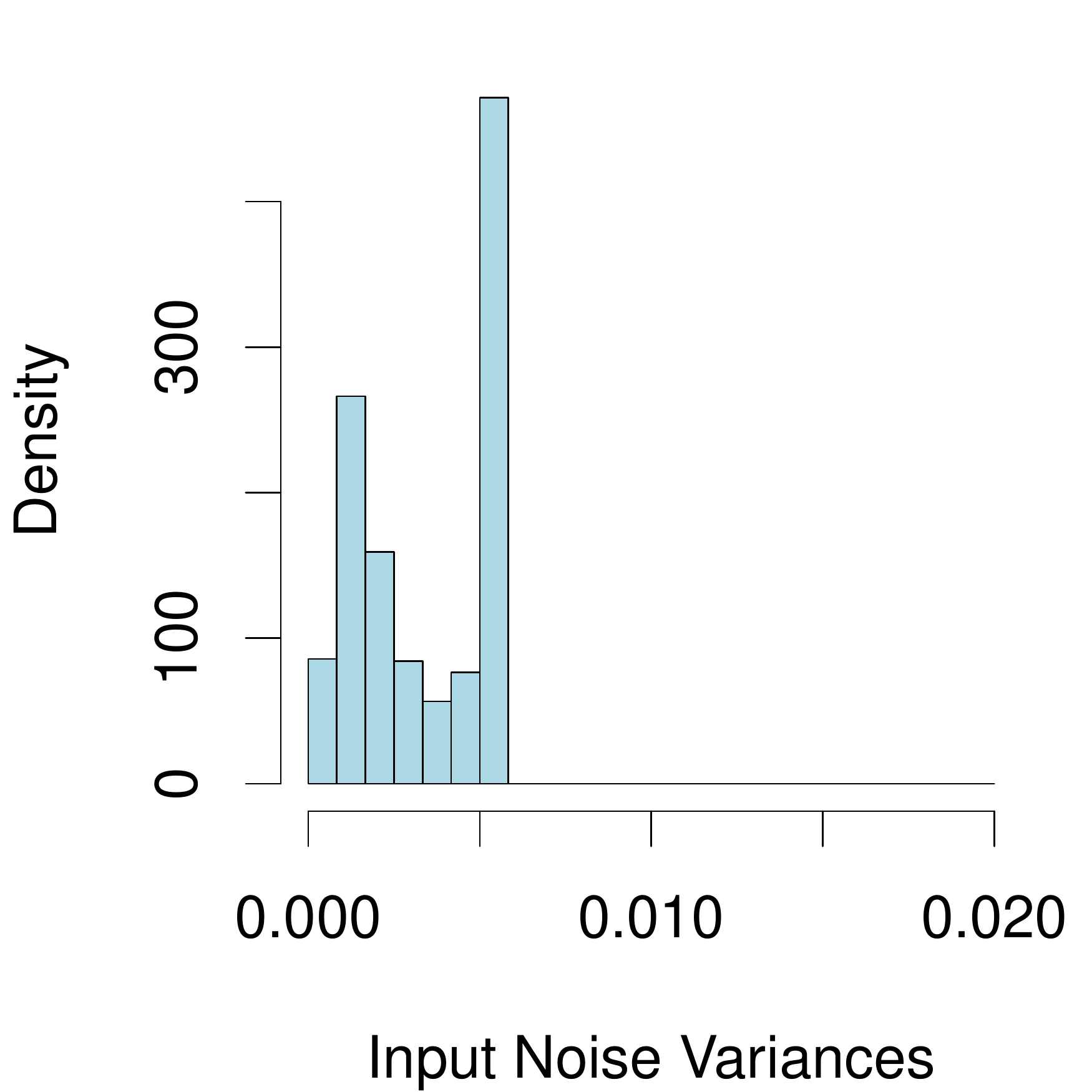} &
				\includegraphics[width=0.25\textwidth]{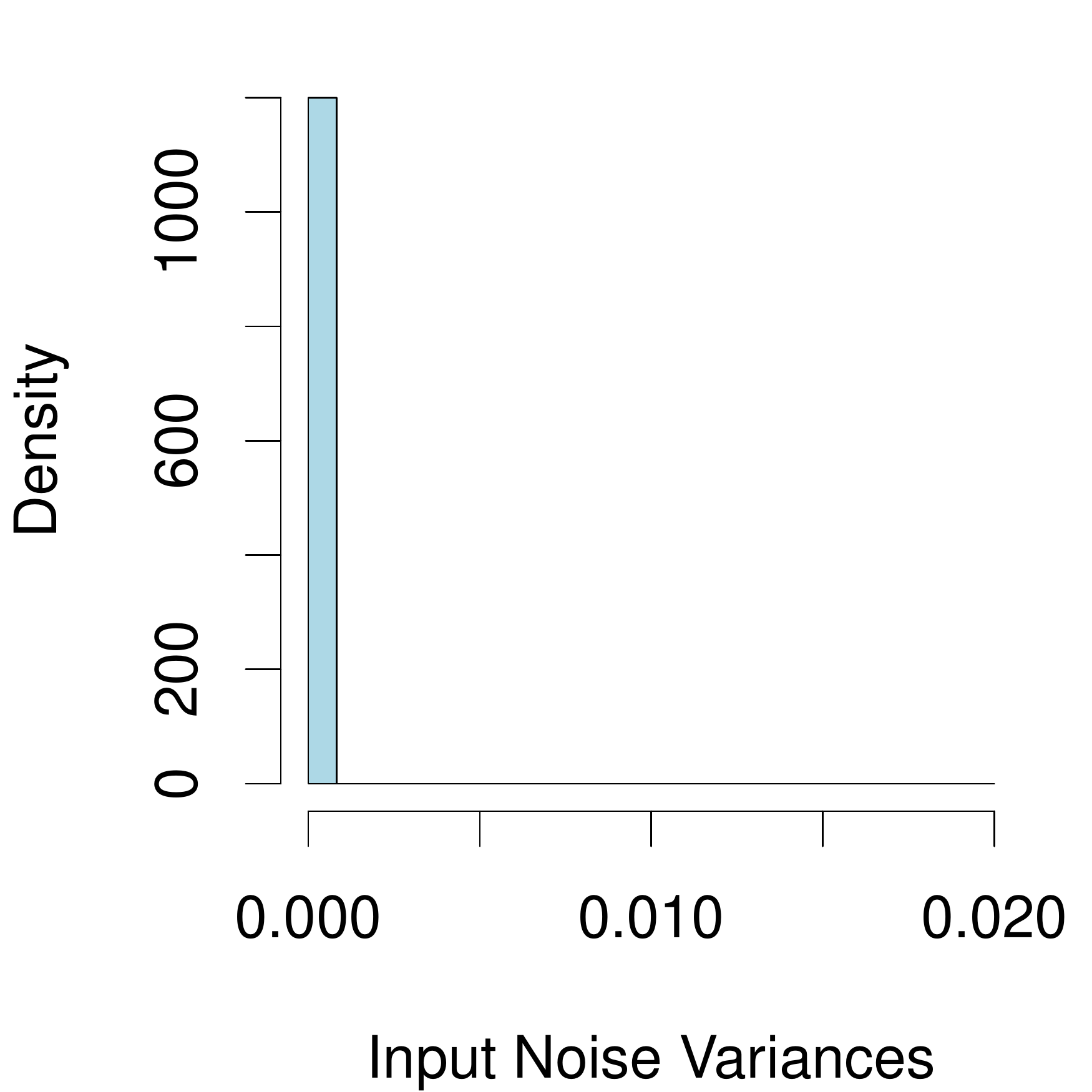} \\
				\hline
				\rotatebox{90}{\bf\hspace{0.35cm} Noise 0.25}  &
				\includegraphics[width=0.25\textwidth]{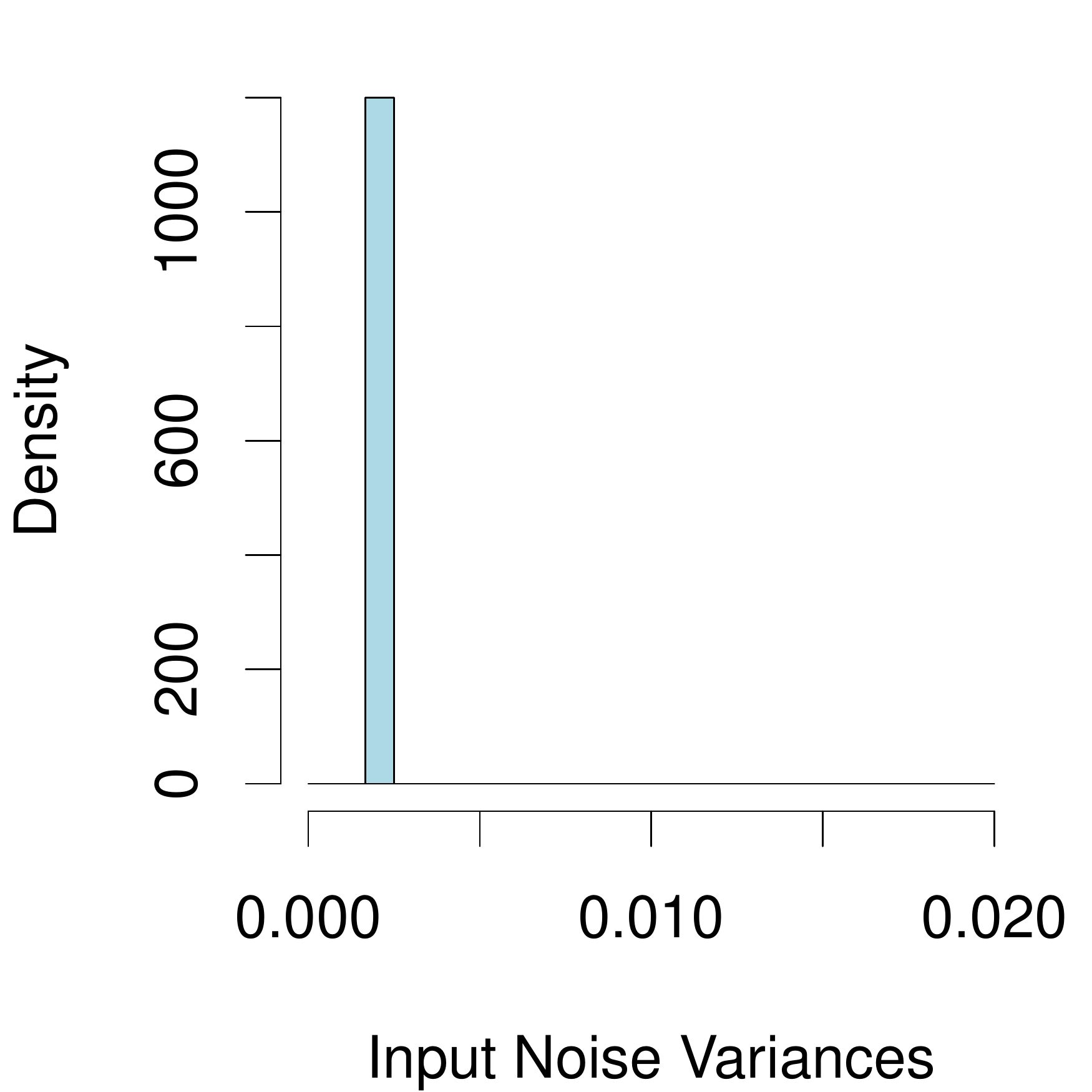} &
				\includegraphics[width=0.25\textwidth]{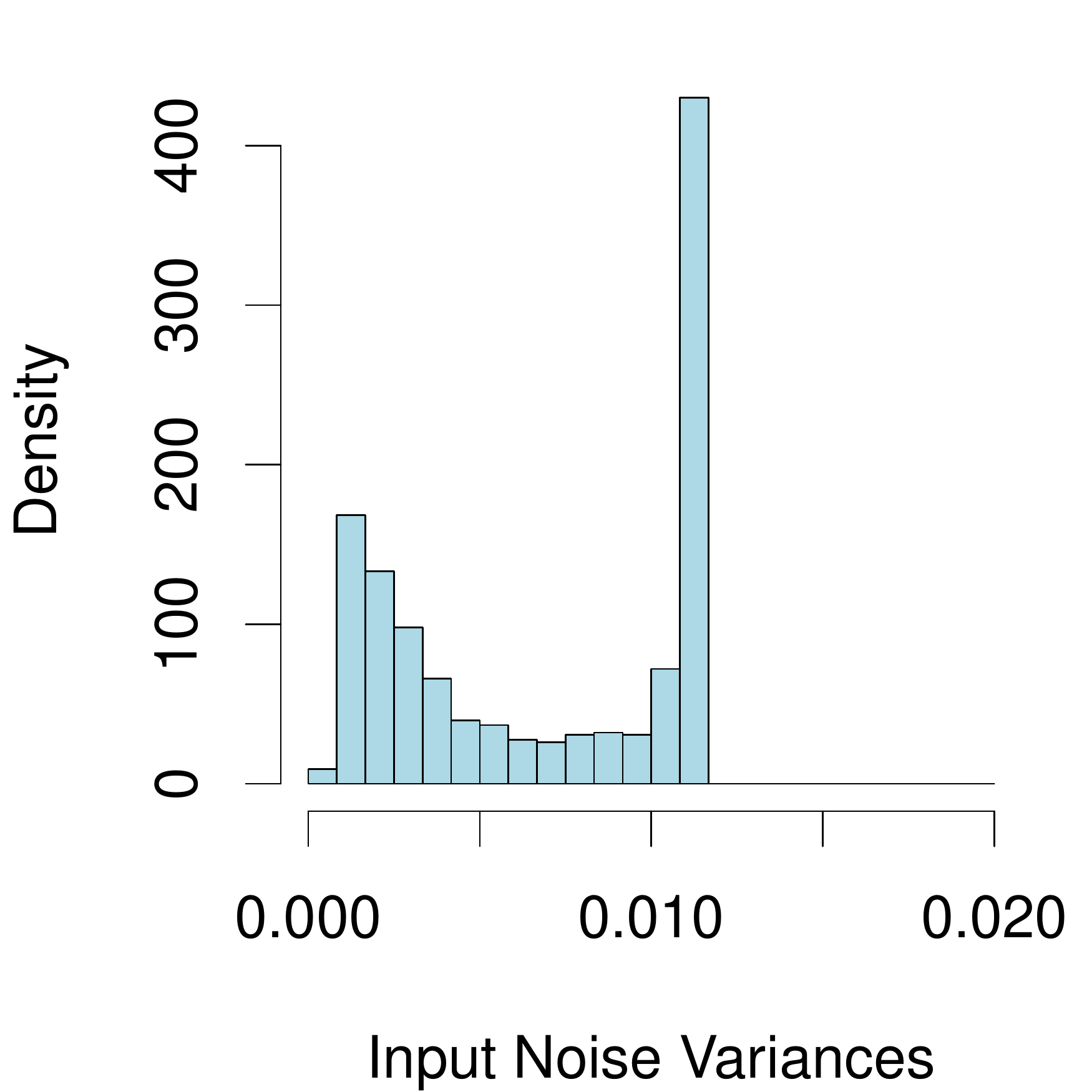} &
				\includegraphics[width=0.25\textwidth]{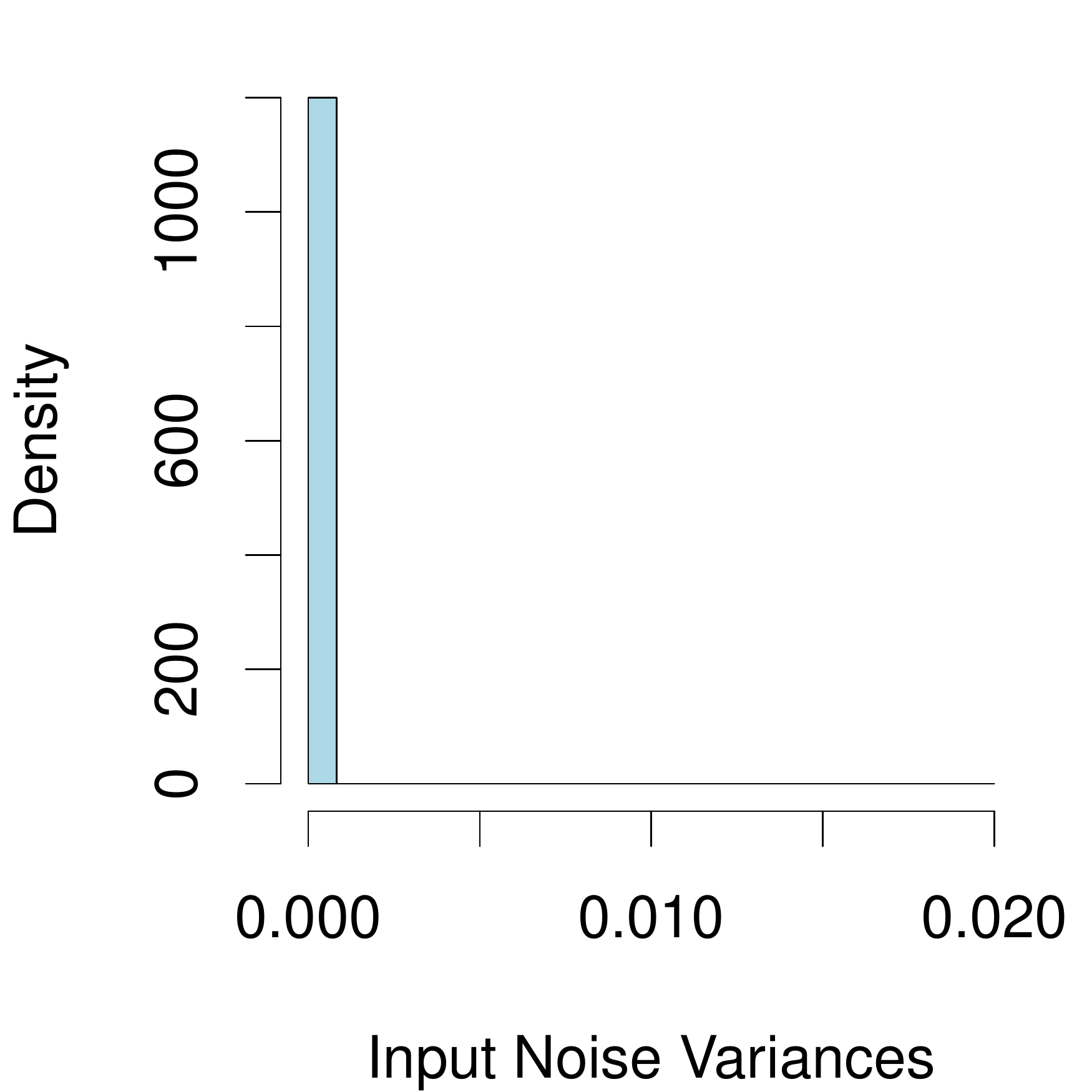} \\
				\hline
				\rotatebox{90}{\bf\hspace{0.45cm} Noise 0.5}  &
				\includegraphics[width=0.25\textwidth]{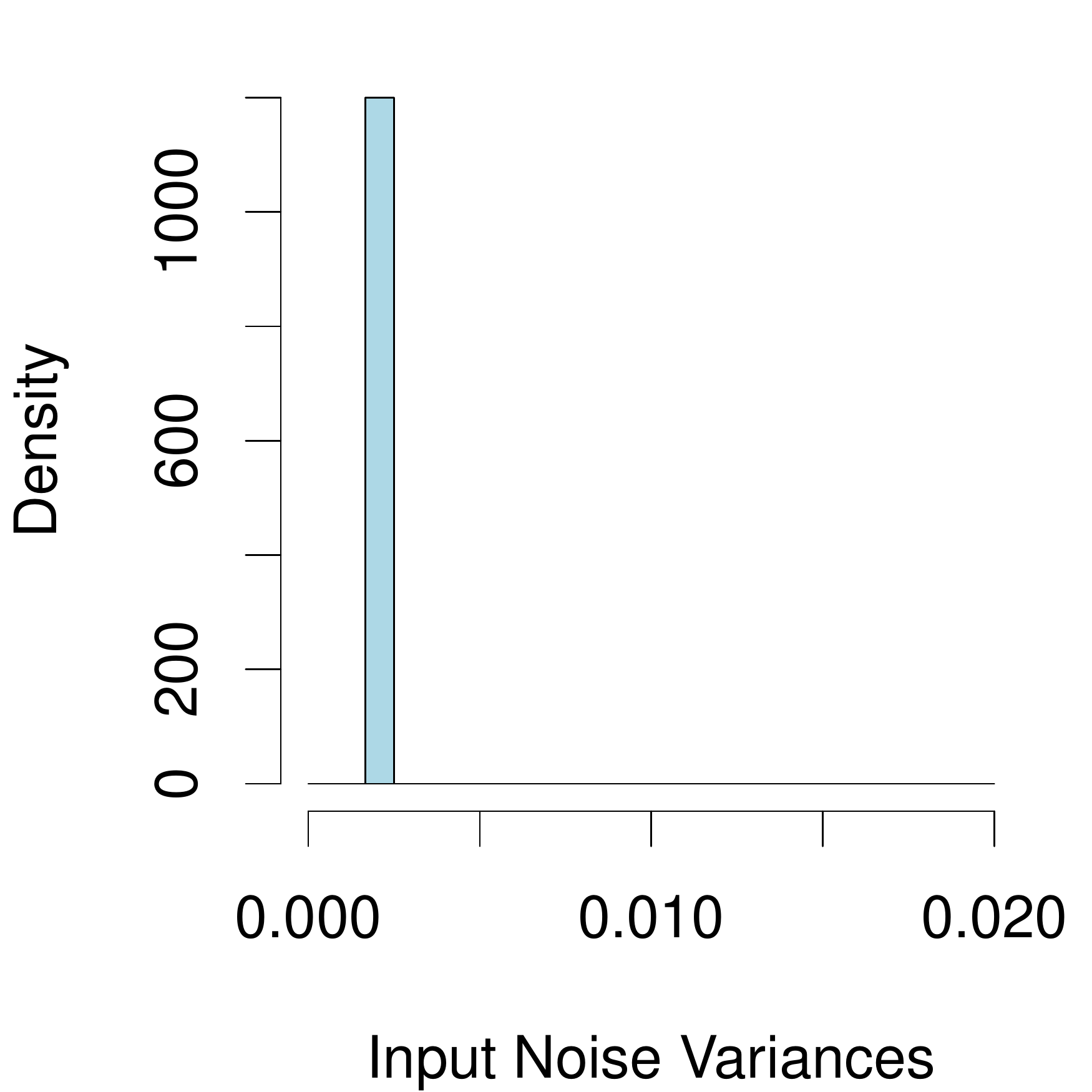} &
				\includegraphics[width=0.25\textwidth]{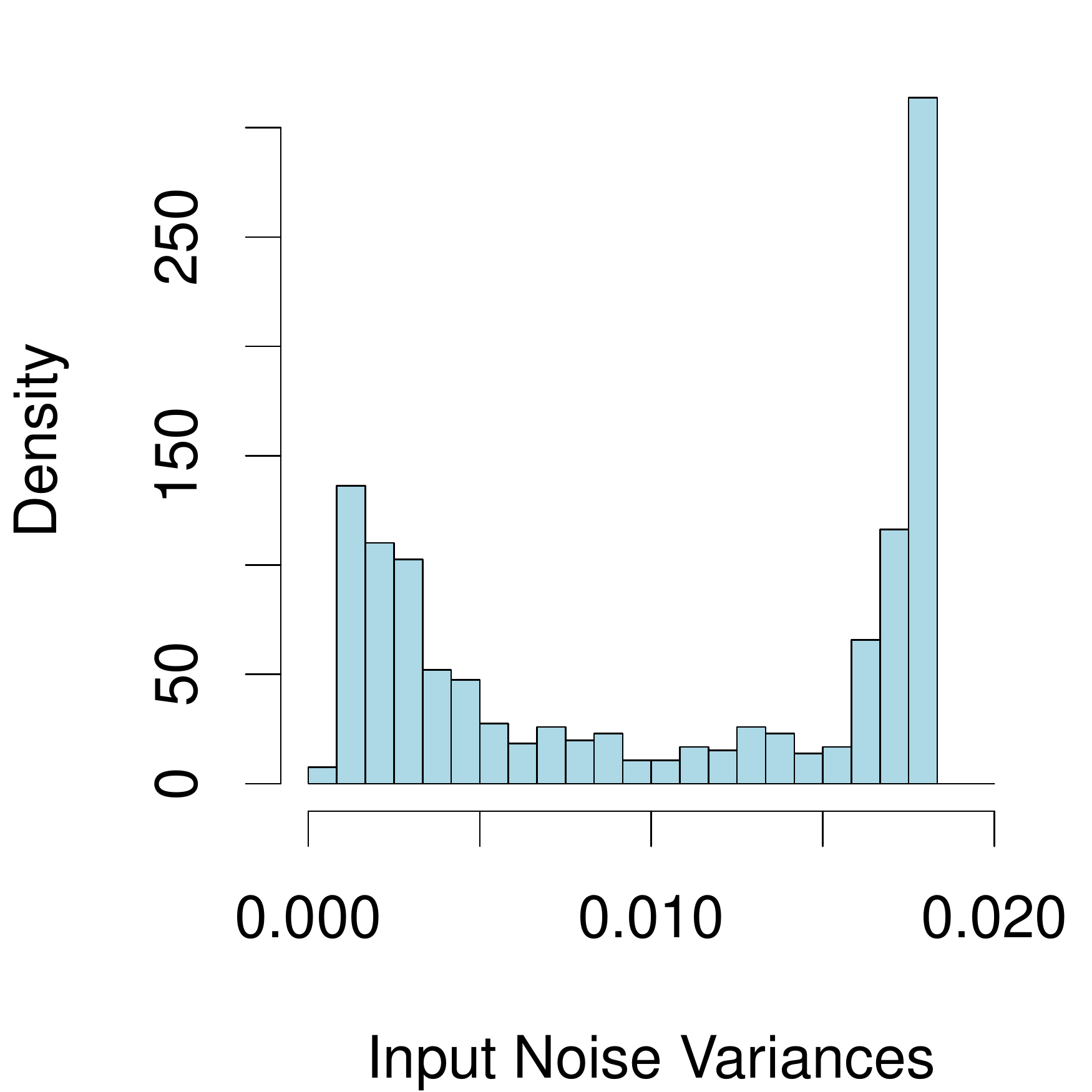} &
				\includegraphics[width=0.25\textwidth]{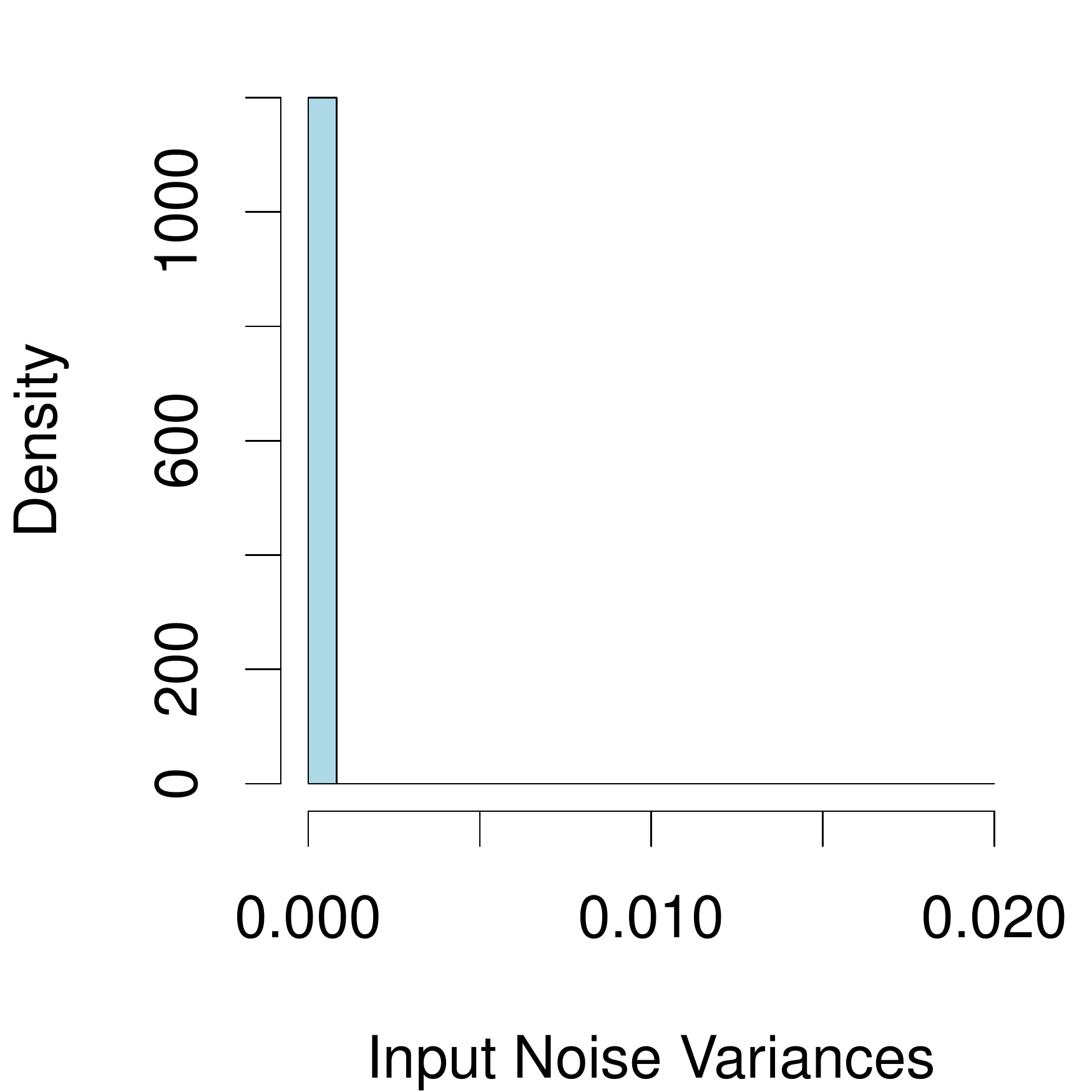} \\
				\hline
				
			\end{tabular}
		\end{center}
		\caption{Histogram of the estimated values for the variance of the input 
			noise, for each method, across the 784 dimensions of the MNIST data set. 
			Note that $\text{NIMGP}_\text{NN}$ is the only method producing
			sensible estimates of the input noise variance in each case. The other methods most of the 
			times estimate values for the noise variance that are almost zero.}
		\label{fig:hist_estimations_MNIST}
	\end{figure}

	Figure \ref{fig:hist_estimations_MNIST} shows a histogram of the estimated variances for the input noise of each
	of the 784 attributes of the MNIST data set, for each method and level of injected noise. We observe that in general
	all the methods except $\text{NIMGP}_\text{NN}$ tend to estimate input variances that are very similar for each
	attribute and that are close to zero. By contrast, $\text{NIMGP}_\text{NN}$ estimates different values for the
	input noise variance associated to each attribute. Furthermore, these values are significantly bigger than the ones
	estimated by the other methods. This may explain the better results obtained by this method on this data set. 
	However, in spite of this, the estimated variances are much smaller than the level of injected noise. This means that
	the method is underestimating the level of input noise. We believe that this can be a consequence of the difficulty
	of this data set for models based on GPs, as commented by \cite{van2017convolutional}.
	
	Table \ref{tab:mnist_time} shows the average training time employed on each epoch for each method. We observe 
	that MGP, NIMGP and $\text{NIMGP}_\text{NN}$ have similar training times. However, $\text{NIMGP}_\text{FO}$
	takes a significantly larger amount of training time on each epoch. This is due to the extra cost of 
	computing the gradients of the GP predictive mean, for each latent function.  This is an expensive operation.
	Recall that the input dimensionality of this data set is high (784 dimensions) and also the 
	number of class labels (10 class labels).

	\begin{table}[b]
		\begin{center}
			\begin{tabular}{cr@{$\pm$}lr@{$\pm$}lr@{$\pm$}lr@{$\pm$}l}
				\hline
				& \multicolumn{2}{c}{\bf MGP}  
				& \multicolumn{2}{c}{\bf NIMGP}  
				& \multicolumn{2}{c}{\bf $\text{NIMGP}_\text{NN}$} 
				& \multicolumn{2}{c}{\bf $\text{NIMGP}_\text{FO}$} \\
				\hline
				{\bf Avg. Time} & 7.19 & 0.007 & 11.93 & 0.011 & 7.62 & 0.011 & 24.01 & 0.074 \\
				\hline
			\end{tabular}
		\end{center}
		\caption{Average time per epoch, in seconds, for each method on the MNIST data set.}
		\label{tab:mnist_time}
	\end{table}

	\subsection{Active Learning Experiments}
	
	We have observed that modeling input noise can lead to better 
	prediction results in terms of the test log-likelihood 
	while the prediction error is similar in most of the cases. 
	Therefore, it seems that the main benefit is a more accurate predictive distribution.
	In this section, we consider an active learning experiment on 
	the \emph{Waveform} data set to illustrate that a better predictive distribution can be useful to 
	improve the generalization error. In active learning the task of interest is to identify
	which examples should be labeled and introduced into the training set to produce a better
	prediction model. We aim at showing that the predictive distribution can be used to 
	identify the training examples that are expected to be most useful for learning a particular task \citep{settles2009active}.
	In particular, a popular choice is to consider the examples whose associated labels the model is more uncertain 
	about \citep{seeger2008}. Therefore, a better predictive distribution should be translated
	into better active learning results. That is, the predictive performance should improve more
	as more and more examples are introduced into the training set, using the aforementioned strategy.
	
	With the goal described, we begin with an initial training set, 
	a test set to evaluate the performance, and a validation set from which we 
	will iteratively pick up a new point to be labeled and included the training set. 
	At each iteration, we will select the point for which the predictive entropy is highest. 
	In order to validate such an approach, we compare this selection criterium with an approach 
	that selects the next point at random from the validation set. We randomly split the Waveform data set 
	into 100 points for the initial training set, 500 points for testing and 400 points for validation. 
	We add 100 new points to the training set using the active learning strategies described. 
	We report results for each method. Namely, MGP, NIMGP, $\text{NIMGP}_\text{NN}$ and 
	$\text{NIMGP}_\text{FO}$ and each level of noise considered in the experiments of Section \ref{sec:exp_uci}.
	All methods are trained using ADAM with a learning 
	rate of $10^{-3}$ the first time for $1,000$ epochs, and then re-trained each 
	time a new point is added to the training set for another $1,000$ epochs, reusing 
	the solution that we have obtained in the previous iteration. The number of 
	inducing points is set to $50$. We report averages over $100$ repetitions of the experiments using
	different splits of the data. In these experiments we also infer the variance of the input noise from the observed data.
	
\begin{figure}[t]
		\begin{center}
			\includegraphics[width=\textwidth]{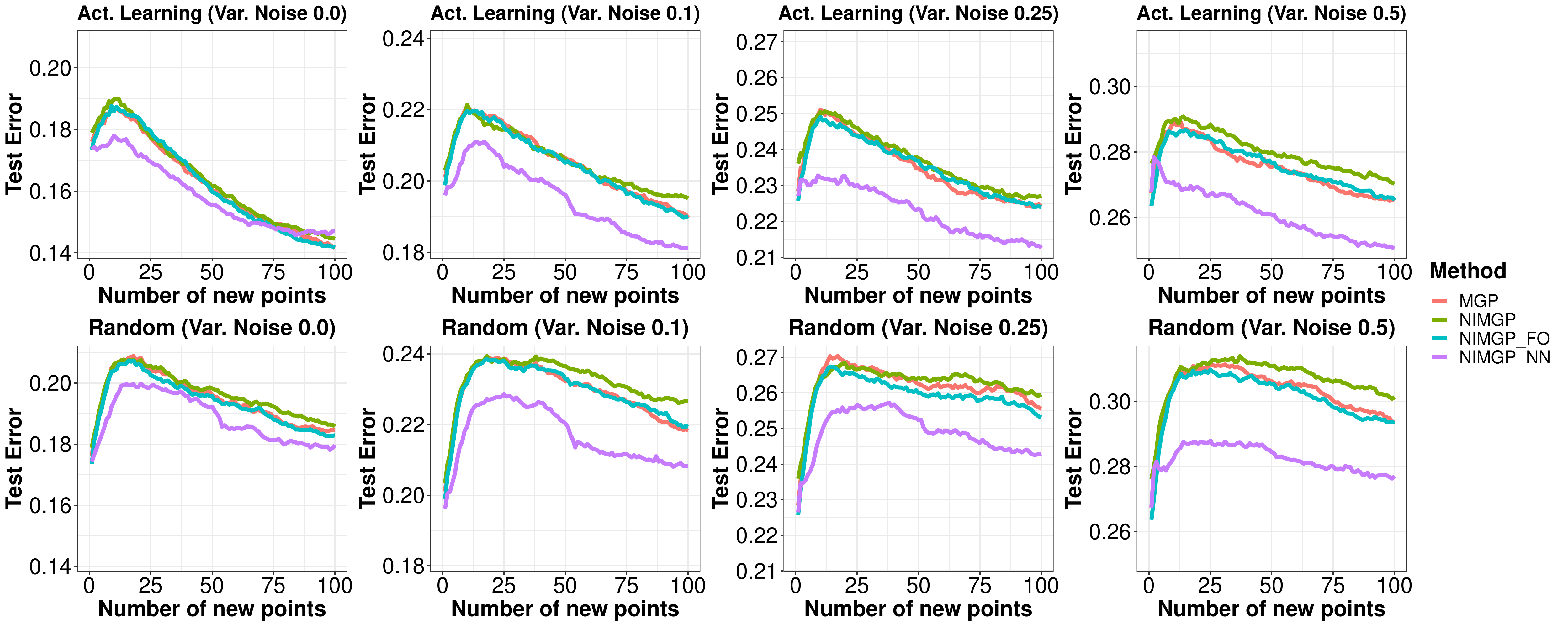}
			\caption{\small Test error on the Waveform data set for different values of the input noise variance (0.0, 0.1, 0.25 and 0.5) 
				as a function of the number of added points to the training set, 
				selected using an active learning approach (top) and or at random (bottom).}
			\label{fig:active_error}
		\end{center}
	\end{figure}

	Figure \ref{fig:active_error} shows the test error as a function of the number of new 
	added points, for each method and each input noise variance injected in the data. 
	In the top row, the new points have been added by using the active learning approach described. 
	That is, the point whose associated predictive entropy is highest, according to the predictive
	distribution of the corresponding model. In the bottom row, the new points have been added 
	at random from the validation set. We observe that the test error is always lower when the 
	active learning selection approach based on the entropy is used. This illustrates that a 
	good predictive distribution can lead to better generalization error. We also observe that
	the prediction error increases at the beginning as the number of new training points considered 
	increases. Eventually it plateaus and begins to decrease. We believe that this behavior could be related 
	to over-fitting happening as a consequence of the procedure to estimate the model hyper-parameters, which
	maximizes an estimate of the marginal likelihood in all the methods. If the number of training points
	is fairly small, this approach can produce over-fitting \citep{rasmussen2005book}. 
	As in the experiments carried out in Section \ref{sec:exp_uci}, the best performing method is $\text{NIMGP}_\text{NN}$.
	
\begin{figure}[t]
		\begin{center}
			\includegraphics[width=\textwidth]{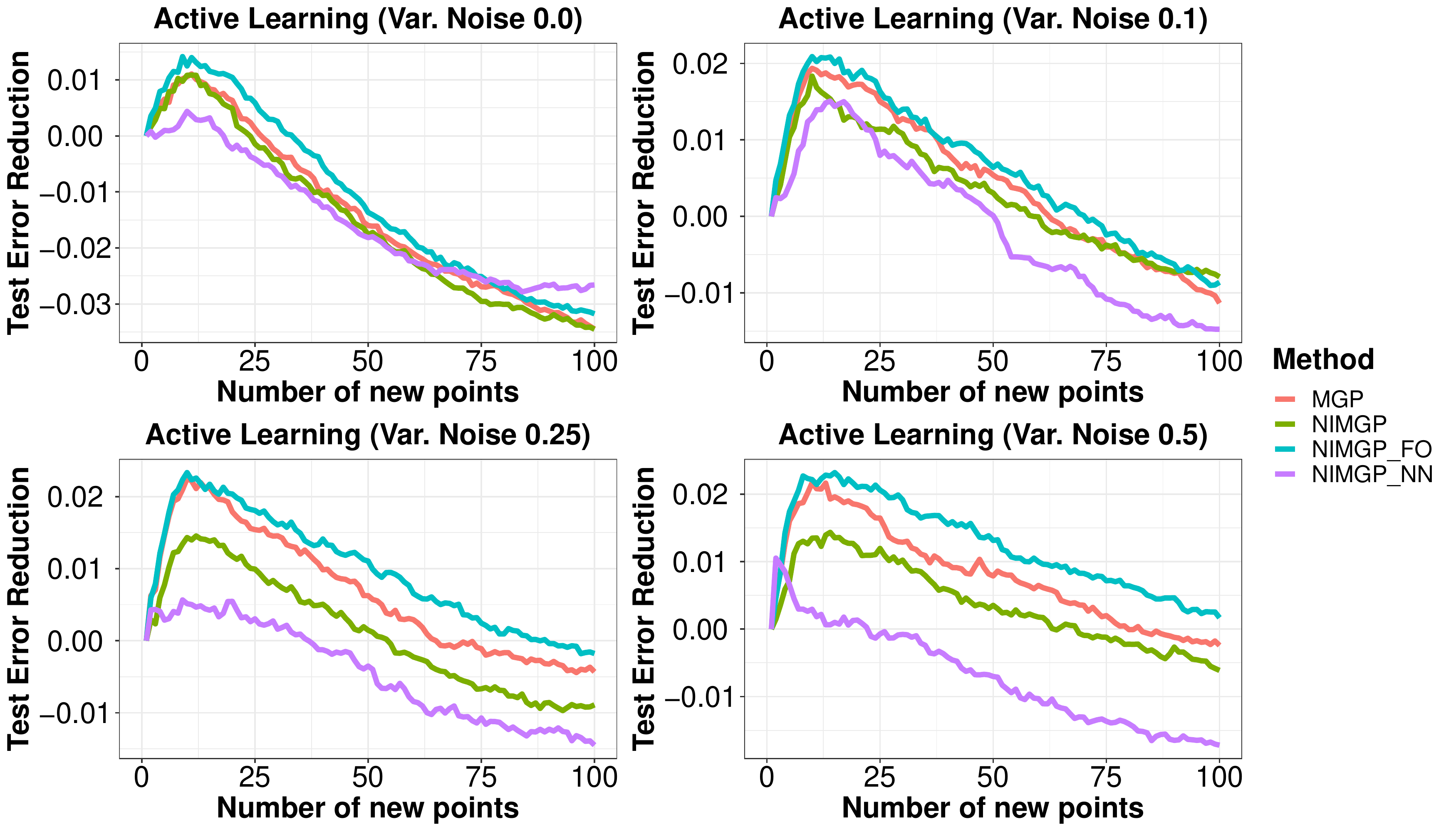}
			\caption{\small Test error reduction on the Waveform data set as a function of the number of new added 
				points to the training set, selected using an active learning approach.}
			\label{fig:active_reduction}
		\end{center}
	\end{figure}

	Figure \ref{fig:active_error} shows that an active learning strategy can be useful to produce better 
	prediction results. We now analyze in which method such a strategy leads to bigger improvements.
	For this, Figure \ref{fig:active_reduction} shows the test error reduction w.r.t. the initial test 
	error for each method and each values of the input noise variance. 
	We observe that the reduction is higher for $\text{NIMGP}_\text{NN}$ than for the other methods, 
	and the difference is bigger between the methods as we increase the variance of the input noise. 
	This suggests that the predictive distribution obtained by $\text{NIMGP}_\text{NN}$ is better. 
	Moreover, it illustrates the potential benefits of considering in the classifier that the observed 
	attributes have been contaminated with additive Gaussian noise.
	
	\subsection{Experiments on a data set Coming from Astrophysics}
	\label{sec:astrophysics}
	
	In this last experimental section we describe the results obtained by each 
	method on a three class data set coming from the astrophysics domain.
	Importantly, in this data the errors in some of the observed inputs are 
	available at training time. Therefore, it is suited to be analyzed using the methods 
	proposed in our paper. As briefly commented in the introduction, the data set 
	consists of a series of attributes measured for a set of point-like astrophysical 
	sources located all over the sky which have already been identified (distinguished 
	from the diffuse background of photon emission) by the Fermi-LAT collaboration. 
	Such catalogue of sources is fully public and can be downloaded from \citet{4FGL}, 
	while a detailed description can be found in \citet{Fermi-LAT:2019yla}. 
	
	\begin{figure}[t]
		\begin{center}
			\includegraphics[width=0.85\textwidth]{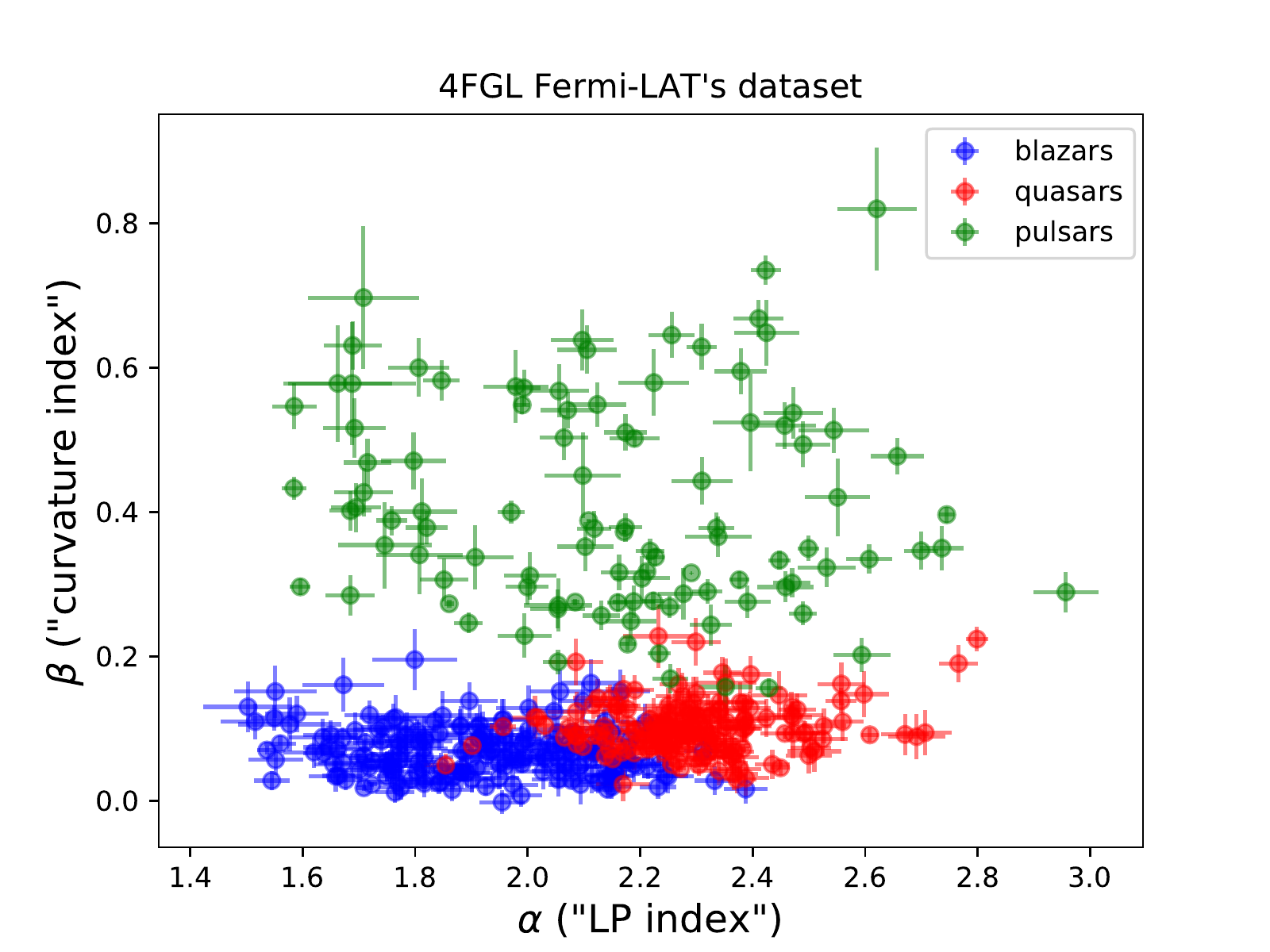}
			\caption{\small The astrophysics data set considered in this work, showing two of the six attributes 
				included in the input with their associated error bars, for the three classes of sources.}
			\label{fig:astro_data}
		\end{center}
	\end{figure}

	Among the available attributes of the sources, there is the position in the sky, 
	the flux of photons in different energy bins, the significance of detection (according 
	to a specified test statistics), the variability of the flux over some period of time, 
	characteristics of the energy spectrum, and many others. Among these, we have only taken 
	into account the following attributes: 1) the photon flux between 1 GeV and 100 GeV, 2) 
	the detection significance (in number of sigmas), 3) the curvature significance \footnote{Defined 
		as the significance, in number of sigmas, of the fit improvement between PowerLaw and 
		LogParabola fits of the energy spectrum.}, 4) the pivot energy \footnote{Defined as the 
		energy  at which the error in the differential photon flux is minimum.}, 5) the index 
	$\alpha$ \footnote{Defined as the spectral slope at pivot energy of a log-parabola fit 
		of the spectrum.} and 6) the index $\beta$ \footnote{Defined as the \emph{curvature} parameter 
		of a log-parabola fit of the spectrum.}. Of these attributes, the flux, the index $\alpha$ and the 
	index $\beta$ come with associated estimated error bars, whereas the rest, by definition, do not. 
	The choice of these attributes among all the available ones is motivated from a physics point of 
	view, where it has been shown \citep[see][]{Fermi-LAT:2019yla}  that some combinations of 
	them offer a promising discrimination power among the classes of point-like sources considered. 
	The reader may wonder about the motivation of two unusual attributes: the detection 
	significance and the curvature significance, being both goodness-of-fit quantities. 
	While the detection significance is indeed partially correlated with the photon flux 
	(which is one of the attributes), the curvature significance a priori may contribute to 
	the discrimination power, since different classes have relatively different spectral 
	shapes. We have checked that the performance of the model without these two attributes 
	is somewhat smaller (see Appendix \ref{sec:appendix_b_extra_astrophisics}), although 
	compatible at the three standard deviation level, with the performance obtained 
	with the full list of attributes.
	
	On the other hand, the catalogue offers as well several classes of sources. Here, for simplicity,
	we have chosen three classes: pulsars (in the catalogue, {\tt psr}), blazars ({\tt bll}) and 
	quasars ({\tt fsrq}) \footnote{In reality, the FSRQ objects are a type of blazars, even though this is an acronym for Flat-Spectrum Radio Quasars.}, which are also the most abundant classes among all the catalogue. The final 
	data set, after pre-processing and pre-selection\footnote{Specifically, we work with sources 
		whose significance of detection is larger than 30 sigmas, otherwise the classes are too 
		overlapped and some of the inputs have too large error bars.} contain 454 sources (points), 
	of which 184 are pulsars, 168 are blazars and 102 are quasars. Figure \ref{fig:astro_data} 
	shows two of the six attributes included in the input of the models alongside with the corresponding 
	error bars. From that figure, one can already observe a good separability of the classes, which improves 
	as the other attributes are taken into account. In general, the pulsars class has larger values 
	of $\beta$ with respect to the other two classes, whereas blazars are separable from 
	quasars in the $\alpha$ direction, but also in the pivot energy direction. 
	
	As in the previous experiments, we have estimated the prediction performance of each method on this data set. For this,
	we have generated 100 splits of the data into training and test partitions with 90\% and 10\% of the instances, respectively.
	We have evaluated the test error rate and negative test log-likelihood on each partition, for each method.
	The average results obtained are displayed in Table \ref{tab:astro-data}. As expected, we obtain a significant improvement 
	in the test log-likelihood when using the proposed methods, which improve on the baseline MGP, which does not 
	take into account the input noise. On the other hand, and consistently as well with the previous experiments, 
	the test error rate is similar for all methods, which is an indication that the decision boundaries among the 
	classes are well captured already by the MGP model.   
	
	\begin{table}[t]
		\begin{center}
			\centering
			\begin{tabular}{l|r@{$\pm$}lr@{$\pm$}lr@{$\pm$}lr@{$\pm$}l}
				\hline
				& \multicolumn{2}{c}{\bf MGP}  
				& \multicolumn{2}{c}{\bf NIMGP}  
				& \multicolumn{2}{c}{\bf $\text{NIMGP}_\text{NN}$} 
				& \multicolumn{2}{c}{\bf $\text{NIMGP}_\text{FO}$} \\
				\hline
				NLL &  0.377 & 0.0194  & \bf 0.246 & \bf 0.0097 & 0.261 & 0.011 &  0.292 & 0.0158 \\
				Test error & 0.075 & 0.0043 & 0.088 & 0.0038 &  0.082 & 0.0041 & \bf 0.071 & \bf 0.0038 \\
				\hline
			\end{tabular}
			\caption{Average test error and neg. test log-likelihood (NLL) of each method on the astrophysics data set.}
			\label{tab:astro-data}
		\end{center}
	\end{table}
	
	We show in Table \ref{tab:confusion-matrix_astro} the average confusion matrix 
	of the four different methods considered in this work. More concretely, each 
	entry in such a table represents the average  (across the different splits) of 
	the number of test samples (which consist of a total of 46 instances) that are assigned
	a particular class label (rows), for a particular actual label of the example (columns). 
	Therefore, elements outside of the diagonal are classification errors.
	As can be observed for such table, the errors are more frequent when discriminating 
	between {\tt bll} and {\tt fsrq} classes, than when classifying the {\tt psr} 
	class. This could have been guessed at some extent by looking at 
	Figure \ref{fig:astro_data}, where the {\tt psr} class (green points) is, 
	at least for the two attributes shown there, easier to discriminate than the other two.
	
	\begin{table}[tb]
		\begin{center}
			\centering
			\begin{tabular}{c|c||r@{$\pm$}l|r@{$\pm$}l|r@{$\pm$}l}
				& Model & \multicolumn{2}{c}{\tt bll} (true) & \multicolumn{2}{c}{\tt fsrq} (true) & \multicolumn{2}{c}{\tt psr} (true) \\
				\hline
				\multirow{4}{*}{{\tt bll} (pred.)}
				& {\bf MGP} & 17.66  &  0.31 &  1.31  &  0.14 &  0.01  &  0.01 \\
				& {\bf NIMGP} & 17.61  &  0.31 &  1.66  &  0.12 &  0.04  &  0.024 \\
				& $\textbf{NIMGP}_\textbf{NN}$ & 17.76  &  0.3 &  1.53  &  0.11 &  0.01  &  0.01 \\
				& $\textbf{NIMGP}_\textbf{FO}$ & 17.8  &  0.32 &  1.27  &  0.12 &  0.01  &  0.01 \\
				\hline
				\multirow{4}{*}{{\tt fsrq} (pred.)}
				& {\bf MGP} & 1.48  &  0.13 &  15.16  &  0.29 &  0.36  &  0.056 \\
				& {\bf NIMGP} & 1.53  &  0.12 &  14.8  &  0.29 &  0.47  &  0.077 \\
				& $\textbf{NIMGP}_\textbf{NN}$ & 1.38  &  0.12 &  14.95  &  0.29 &  0.46  &  0.072 \\
				& $\textbf{NIMGP}_\textbf{FO}$ & 1.34  &  0.12 &  15.2  &  0.28 &  0.37  &  0.06 \\
				\hline
				\multirow{4}{*}{{\tt psr} (pred.)}
				& {\bf MGP} & 0.00  &  0.00 &  0.29  &  0.054 &  9.73  &  0.28 \\
				& {\bf NIMGP} & 0.00  &  0.00 &  0.3  &  0.056 &  9.59  &  0.27 \\
				& $\textbf{NIMGP}_\textbf{NN}$ & 0.00  &  0.00 &  0.28  &  0.047 &  9.63  &  0.28 \\
				& $\textbf{NIMGP}_\textbf{FO}$ & 0.00  &  0.00 &  0.29  &  0.052 &  9.72  &  0.27 \\
				\hline
				
			\end{tabular}
			\caption{Average confusion matrix for the astrophysics data set, for each different method considered 
				in this work (see text for details).}
			\label{tab:confusion-matrix_astro}
		\end{center}
	\end{table}

	\section{Conclusions}
	\label{sec:conclusions}
	
	Multi-class classification problems involve estimating a predictive distribution for the class label given the 
	observed data attributes. Multi-class Gaussian process classifiers are kernel machines that can be used to address
	these problems with the benefit that they will take into account uncertainty in the estimation process.
	Often, the supervised machine learning community assumes that the observed data are noise-free in the explaining inputs.
	Notwithstanding, in some scenarios the measurement of the explaining variables is contaminated with noise. Therefore,
	input noise is often common in many problems of interest. If this input noise is not modeled correctly,
	the quality of the resulting predictive distribution can be sub-optimal.
	
	In this paper we have proposed several multi-class GP classifiers that can account for input noise. 
	All these classifiers can be efficiently trained using variational inference to approximate the posterior 
	distribution of the latent variables of the model. They also allow to specify manually, or to infer
	from the observed data, the level of input noise. Two approaches are based on introducing extra latent variables
	in the model to account for noisy inputs, one of them using a neural network to amortize variational parameters.
	The last method is, however, based on a linear approximation of the GPs of the classifier. 
	Under this approximation input noise is directly translated into output noise.
	
	The inductive bias described is expected to lead to better performance results in practical data sets. 
	To show this, we have evaluated the proposed methods on several experiments, involving synthetic and 
	real data. These include several data sets from the UCI repository, the MNIST data set and also a data set coming from 
	astrophysics. We have compared the results of the proposed methods with those of a standard multi-class GP 
	classifier that ignores input noise.  The experiments show that the predictive distribution of the proposed methods is 
	significantly better in terms of the test log-likelihood. The classification error is, however, similar.
	This means that the decision boundaries of the classifier are not significantly changed. Only the class prediction probabilities.
	
	We have also analyzed the ability of each method to infer the level of input noise from the observed data by
	approximately maximizing the marginal likelihood. Our results show that accurate estimation results can only be 
	obtained in the case of synthetic data sets, and only for the method that uses a neural network to amortize 
	variational parameters. The other methods prefer input noise variances that are close to zero. 
	This is an unexpected result and analyzing the reason for this behavior is left for future work. In consequence, 
	the method that uses the neural network is the one that obtains the best results in real-world problems in which 
	input noise has been injected in the observed attributes. These problems include data sets extracted 
	from the UCI repository and the MNIST data set.
	
	We have also measured the average time per epoch for each method. Our results show that the method that
	amortizes the variational parameters using a neural network has a training time that is very similar to that
	of a standard multi-class GP classifier. This method scales better to bigger data sets. 
	By contrast, the other two methods lead to a significant increment in the computational cost. 
	This increment is much bigger in the case of the method that uses a linear approximation 
	of the GPs employed in the classifier. The reason for this is that computing the required derivatives is expensive.
	
	We have illustrated the utility of a better predictive distribution by carrying out a
	active learning experiments. In these experiment the uncertainty about the class label 
	of new unseen data is used to choose which data instances should be labeled and 
	introduced into the training set with the goal of improving prediction error the most. 
	Our results show that the method that produces the most accurate predictive distribution (\emph{i.e.}, 
	the one using the neural network to amortize variational parameters) is also the one that identifies the most 
	relevant training points, leading to the largest reduction in the prediction error.
	This specific application shows that a better predictive distribution can also result in 
	better prediction error.
	
	We have also evaluated the proposed methods in a real data set involving astrophysics 
	data with success, both in terms of error rate and negative test log-likelihood. These 
	experiments add empirical evidence to the hypothesis that if we model input noise 
	the results of multi-class classification using GPs can be enhanced. 
	Summing up, our results indicate that if one is interested in obtaining accurate 
	predictive distributions, it is of vital importance to take into account 
	any potential input noise that has contaminated the explaining variables of the
	multi-class classification problem.
	
	We believe that the reason for the prediction error being similar is due to the fact
	that it only depends on the decision boundaries. These boundaries are fully determined
	by the class label with the largest posterior probability, as estimated by the corresponding
	method. Therefore, an accurate decision boundary estimation does not require an accurate
	class posterior probability estimation. However, if one is concerned about the quality of
	the predictive distribution and the uncertainty in the predictions made by the method, an
	accurate class posterior probability estimation is strictly required. The proposed methods
	significantly improve the quality of class posterior probability estimation, while providing
	similar prediction errors.
	
	\acks{
		We would like to thank M. A. S\'anchez-Conde, J. Coronado and V. Gammaldi for pointing our attention to the data set that 
		motivated this work, as well as for the discussions concerning the data extraction. We thank as well 
		E. Fern\'andez-Mart\'inez, A. Su\'arez and C. M. Ala\'iz-Gud\'in for useful discussions and feedback about the work. 
		BZ especially acknowledges the hospitality of the Machine Learning group of UAM during the development of this project.  
		BZ is supported by the Programa Atracci\'on de Talento de la Comunidad de Madrid under grant n. 2017-T2/TIC-5455,
		from the Spanish MINECO’s “Centro de Excelencia Severo Ochoa” Programme via grant SEV-2016-0597, and from the Comunidad de Madrid project SI1-PJI-2019-00294, of which BZ is the P.I. The authors gratefully
		acknowledge the use of the facilities of Centro de Computaci\'on Cient\'ifica (CCC) at
		Universidad Aut\'onoma de Madrid. The authors also acknowledge financial support from Spanish
		Plan Nacional I+D+i, grants TIN2016-76406-P. Finally, the authors acknowledge financial support from PID2019-106827GB-I00 /
		AEI / 10.13039/501100011033.
	}
	
	\appendix 
	
	\section{Stochastic Approximation of the Lower Bound}
	\label{sec:appendix_a}
	
	In this section we describe how to compute an stochastic estimate of the ELBO, described in (\ref{eq:elbo}). The stochasticity
	of the approximation arises from (i) using mini-batches of data to approximate the data-dependent term, and from (ii) approximating
	the corresponding expectations using Monte Carlo sampling. For this, we use the fact that the approximate distribution $q$ is 
	reparametrizable in the sense that it allows to separate, in each random sample, the dependence on the distribution parameters and 
	the randomness. 
	
	Consider the first term in the ELBO and a mini-batch of data instances $\mathcal{B}$. Then,
	\begin{align}
		\sum_{i=1}^N \mathds{E}_q\left[ \log p(y_i|\mathbf{f}_i)\right]
		\approx \frac{|\mathcal{B}|}{N}\sum_{i \in \mathcal{B}}\mathds{E}_q\left[ \log p(y_i|\mathbf{f}_i)\right]
		\label{eq:first_term}
	\end{align}
	results in an un-biased estimate of the first term in (\ref{eq:elbo}). 
	
	Consider now the term $\mathds{E}_q\left[ \log p(y_i|\mathbf{f}_i)\right]$.
	The expectation with respect to the $q$ distribution can be computed analytically in the case of the random variables $\mathbf{U}$. For this,
	we only have to use (\ref{eq:marginalizing_u}). Furthermore, the expectation with respect to the posterior distribution of $\mathbf{f}_i$ can
	be approximated accurately using a one-dimensional quadrature. In particular,
	\begin{align}
		\mathds{E}_{q(\mathbf{f}_i)} [\log p(y_i|\mathbf{f}_i)]  &= 
		(1 - e) \frac{1}{S}\sum_{i=1}^S\int \mathcal{N}(f^{y_i}(\mathbf{x}_i)|m_{y_i}, v_{y_i})
		\prod_{c\neq y_i}\Phi\left(\frac{f^{y_i}(\mathbf{x}_i) - m_c)}{\sqrt{v_c}}\right) d f^{y_i}(\mathbf{x}_i) \nonumber \\ 
		& \quad + \frac{e}{C-1}\,,
		\label{eq:quadrature}
	\end{align}
	where $\Phi(\cdot)$ is the cumulative probability of a standard Gaussian distribution and
	\begin{align*}
		m_c &= \mathbf{k}^c_{\mathbf{x}_i,\mathbf{Z}^c} (\mathbf{K}_{\mathbf{Z}^c,\mathbf{Z}^c})^{-1} \mathbf{m}_c\,, \\
		v_c &= k^c_{\mathbf{x}_i,\mathbf{x}_i} - \mathbf{k}^c_{\mathbf{x}_i,\mathbf{Z}^c} 
		(\mathbf{K}^c_{\mathbf{Z}^c,\mathbf{Z}^c})^{-1} \mathbf{K}_{\mathbf{Z}^c,\mathbf{x}_i} + 
		\mathbf{k}^c_{\mathbf{x}_i,\mathbf{Z}^c} (\mathbf{K}^c_{\mathbf{Z}^c,\mathbf{Z}^c})^{-1}\mathbf{S}_c(\mathbf{K}^c_{\mathbf{Z}^c,\mathbf{Z}^c})^{-1} 
		\mathbf{k}^c_{\mathbf{Z}^c,\mathbf{x}_i}\,,
	\end{align*}
	for $k=1,\ldots,C$ with $k_{\mathbf{x}_i,\mathbf{x}_i}^c$ the variance of
	$f^c(\mathbf{x}_i)$, $\mathbf{k}_{\mathbf{x}_i,{\mathbf{Z}}^c}$ the covariance vector between $f^c(\mathbf{x}_i)$ and $f^c(\cdot)$ evaluated 
	at $\mathbf{Z}^c$, and $\mathbf{K}_{\mathbf{Z}^c,\mathbf{Z}^c}$ the covariance matrix among the values of $f^c(\cdot)$ at $\mathbf{Z}^c$. 
	All these values and matrices can be easily computed given the corresponding covariance functions $\{k_{\theta_c}(\cdot,\cdot)\}_{c=1}^C$.
	Finally, $\mathbf{m}_c$ and $\mathbf{S}_c$ are the mean and covariance parameters of $q(\mathbf{u}^c)$, respectively. See (\ref{eq:post_approx}) for further details.
	
	It remains now to approximate the expectation of $\mathds{E}_{q(\mathbf{f}_i)} [\log p(y_i|\mathbf{f}_i)]$ with respect to $q(\mathbf{x}_i)$. This is done
	by using a Monte Carlo approximation combined with the reparametrization trick \citep{KingmaW13}. More precisely, we generate a single sample from $q(\mathbf{x}_i)$ 
	by separating the randomness and the dependence on the parameters of $q(\mathbf{x}_i)$. Namely,
	\begin{align*}
		\hat{\mathbf{x}}_i &= \mathbf{L}_i \bm{\epsilon} + \bm{\mu}_i^x\,, &  \bm{\epsilon} \sim \mathcal{N}(\mathbf{0},\mathbf{I})\,,
	\end{align*}
	where $\mathbf{L}_i$ is a diagonal matrix whose entries contain the square root of the diagonal entries of $\mathbf{V}_i^x$,
	and $\bm{\mu}_i^x$ is the mean of $\mathbf{x}_i$. See (\ref{eq:post_approx}) for further details. Let us define 
	$\hat{\mathbf{f}}_i=(f^1(\hat{\mathbf{x}}_i), \ldots, f^C(\hat{\mathbf{x}}_i))^\text{T}$. Then,
	\begin{align*}
		\mathds{E}_{q(\mathbf{x}_i)}[\mathds{E}_{q(\mathbf{f}_i)} [\log p(y_i|\mathbf{f}_i)]] & \approx
		\mathds{E}_{q(\hat{\mathbf{f}}_i)} [\log p(y_i|\hat{\mathbf{f}}_i)]\,,
	\end{align*}
	where the right hand side is given by (\ref{eq:quadrature}) in which we have replaced $\mathbf{x}_i$ by $\hat{\mathbf{x}}_i$.
	The consequence is that
	\begin{align}
		\sum_{i=1}^N \mathds{E}_q\left[ \log p(y_i|\mathbf{f}_i)\right]
		\approx \frac{|\mathcal{B}|}{N}\sum_{i \in \mathcal{B}}\mathds{E}_{q(\hat{\mathbf{f}}_i)}\left[ \log p(y_i|\hat{\mathbf{f}}_i)\right]
		\,,
		\label{eq:first_term_monte_carlo}
	\end{align}
	where the right hand side of is an unbiased estimate of the left hand side. 
	
	The second term in (\ref{eq:elbo}) can be approximated using a mini-batch and the corresponding expectation can be computed analytically. In particular,
	\begin{align*}
		\sum_{i=1}^N \mathds{E}_q[\log p(\tilde{\mathbf{x}}_i|\mathbf{x}_i)] & \approx
		\frac{|\mathcal{B}|}{N} \sum_{i \in \mathcal{B}}\mathds{E}_q[\log p(\tilde{\mathbf{x}}_i|\mathbf{x}_i)]  \nonumber \\
		&= \frac{|\mathcal{B}|}{N} \sum_{i \in \mathcal{B}} \left[ -\frac{d}{2}\log 2 \pi - \frac{1}{2} \log |\mathbf{V}_i| 
		\nonumber \right.\\ 
		& \quad \left.
		- \frac{1}{2} \text{trace}\left( \mathbf{V}_i(\mathbf{V}_i^x + \bm{\mu}_i^x (\bm{\mu}_i^x)^\text{T}) \right)
		+ \tilde{\mathbf{x}}_i^\text{T} \mathbf{V}_i \bm{\mu}_i^x
		- \frac{1}{2} \tilde{\mathbf{x}}_i^\text{T} \mathbf{V}_i  \tilde{\mathbf{x}}_i
		\right]
		\label{eq:second_term}
	\end{align*}
	
	The third term in (\ref{eq:elbo}) is the Kullback-Leibler divergence between Gaussian distributions. This is given by
	\begin{align*}
		\sum_{c=1}^C \text{KL}(q(\mathbf{u}^c)|p(\mathbf{u}^c))  &= 
		\sum_{c=1}^C \frac{1}{2} \left[ \text{trace}\left((\mathbf{K}^c_{\mathbf{Z}^c,\mathbf{Z}^c})^{-1} \mathbf{S}_c \right)
		\right. \nonumber \\
		& \quad \left.
		+ \mathbf{m}_c^\text{T}(\mathbf{K}^c_{\mathbf{Z}^c,\mathbf{Z}^c})^{-1} \mathbf{m}_c - M + \log \frac{|\mathbf{K}^c_{\mathbf{Z}^c,\mathbf{Z}^c}|}{|\mathbf{S}_c|}
		\right]\,.
	\end{align*}
	where $\mathbf{m}_c$ and $\mathbf{S}_c$ are the parameters of $q(\mathbf{u}^c)$, and $\mathbf{K}^c_{\mathbf{Z}^c,\mathbf{Z}^c}$
	is the covariance matrix of $p(\mathbf{u}^c)$.
	
	The fourth term in (\ref{eq:elbo}) can be approximated using a mini-batch and the corresponding Kullback-Leibler divergence can be computed analytically. In particular,
	\begin{align*}
		\sum_{i=1}^N \text{KL}(q(\mathbf{x}^i)|p(\mathbf{x}_i)) & \approx
		\frac{|\mathcal{B}|}{N} \sum_{i \in \mathcal{B}} \text{KL}(q(\mathbf{x}^i)|p(\mathbf{x}_i)) \nonumber \\
		& = \frac{|\mathcal{B}|}{N} \sum_{i \in \mathcal{B}} \frac{1}{2} \left[ \text{trace}\left(\mathbf{V}_i^x \mathbf{I}s^{-1}\right)
		+ (\bm{\mu}_i^x)^\text{T} \bm{\mu}_i^x s^{-1} - d + \log \frac{s^d}{|\mathbf{V}_i^x|} \right]
		\,.
	\end{align*}
	
	Note that all these estimates are unbiased. In practice, the stochastic estimate of the lower bound can be easily
	codified in a framework such as Tensorflow \citep{tensorflow2015}, and the corresponding unbiased estimate of the 
	gradient can be computed using automatic differentiation.
	
	\newpage
	
	\section{Extra Experiments}
	\label{sec:appendix_extra_experiments}
	
	In this section we report extra experimental results not covered in the main manuscript.
	
	\subsection{Synthetic Experiments Using Data Augmentation and Sampling}
	
	In this section we compare the original MGP model when we do not 
	take into account the noise in the inputs with two baselines. Namely, the same model, \emph{i.e.}, 
	MGP trained on an augmented data set with samples from the posterior distribution of the observations 
	$p(\tilde{\mathbf{x}}_i|\mathbf{x}_i) = \mathcal{N}(\tilde{\mathbf{x}}_i|\mathbf{x}_i, \mathbf{V}_i)$ 
	assuming uniform prior. The second baseline is a modified version of MGP in which the input attributes 
	of each batch are obtained by sampling from the corresponding posterior distribution under a uniform prior, 
	ignoring label information. For example, if $x_i \sim N(\tilde{x}_i,\sigma^2)$, where $\tilde{x}_i$ is the
	noiseless attribute, then $p(\tilde{x}_i|x_i)=\mathcal{N}(\tilde{x}_i|x_i,\sigma^2)$, assuming a uniform prior
	distribution for $\tilde{x}_i$. When processing each batch of points MGP, we simply sample the input attributes 
	from $p(\tilde{x}_i|x_i)$. The synthetic data set from Section \ref{sec:synt_exp} is considered.
	We generate as many data as twice the amount of data instances present in the training set.
	Note that we cannot sample from the prior distribution of noisy data since we observe noisy observations of 
	the inputs and hence have no access to such a distribution. 
	We report averages over 100 repetitions of the experiments.
	
	\begin{table}[H]
		\centering
		\begin{tabular}{l|r@{$\pm$}lr@{$\pm$}lr@{$\pm$}l}
			\hline 
			& \multicolumn{2}{c}{\bf MGP }&\multicolumn{2}{c}{\bf  MGP  (aug.) } & \multicolumn{2}{c}{\bf  MGP  (sampling)}\\
			\hline
			Noise  0.1 &   0.76  &   0.022 &  0.8  &  0.022 & \bf 0.45 & \bf 00.009\\
			Noise  0.25 &   1.1  &   0.033 &  1.2  &  0.035 & \bf 0.61 & \bf 00.013\\
			Noise  0.5 &   1.5  &   0.042 &  1.6  &  0.045  & \bf 0.74 & \bf 00.014\\
			\hline
		\end{tabular}
		\caption{Average Neg. Test Log Likelihood comparison between MGP, MGP with an augmented data set and MGP using 
			sampling from the posterior distribution of the attributes.}
		\label{tab:syn_augm_nll}
	\end{table}

	\begin{table}[H]
		\centering
		\begin{tabular}{l|r@{$\pm$}lr@{$\pm$}lr@{$\pm$}l}
			\hline 
			& \multicolumn{2}{c}{\bf MGP }&\multicolumn{2}{c}{\bf  MGP (aug.) } & \multicolumn{2}{c}{\bf  MGP (sampling) }\\
			\hline
			Noise  0.1 & \bf  0.11  & \bf  0.0032 &  0.12  &  0.0031 & 0.13 & 0.0035 \\
			Noise  0.25 & \bf  0.16  & \bf  0.0048 &  0.17  &  0.005 & 0.194 & 0.0056 \\
			Noise  0.5 & \bf  0.22  & \bf  0.0058 & \bf  0.22  & \bf  0.0063 & 0.26 & 0.0066 \\
			\hline
		\end{tabular}
		\caption{Average  Test Error comparison between MGP, MGP with an augmented data set and MGP using 
			sampling from the posterior distribution of the attributes.}
		\label{tab:syn_augm_err}
	\end{table}
	
	Tables \ref{tab:syn_augm_nll} and \ref{tab:syn_augm_err} show the average test 
	log-likelihood and average test error for MGP, MGP on the augmented 
	data set and MGP using the sampling scheme described. We observe that augmenting 
	the data set with samples from the posterior distribution does not improve the results.
	The sampling approach leads to worse prediction error results. However, the test log-likelihood improves slightly.
	In any case, the improvements over MGP are smaller than those obtained by the proposed methods 
	NIMGP, $\text{NIMGP}_\text{NN}$ and $\text{NIMGP}_\text{FO}$ in Section \ref{sec:synt_exp}.
	We believe that the reason for the sampling approach not performing very well compared to these methods 
	is that it ignores the information provided by the labels about the potential vales of the noiseless 
	input attributes.
	
	\newpage
	
	\subsection{Synthetic Experiments Varying the Dimensions and the Number of Points}
	
	We repeat the experiments of Section \ref{sec:synt_exp} changing the number of dimensions $d$ and the number of training data $N$.
	We assume the variance of the input noise is given. The results obtained are similar to the 
	ones reported in that Section. Larger $d$ and $N$ lead to better results as there are 
	more explaining attributes and more training data.
	
	\begin{table}[H]
		\scriptsize
		\begin{subtable}{\linewidth}
			\centering
			\begin{tabular}{l|c|r@{$\pm$}l|r@{$\pm$}l|r@{$\pm$}l}
				& Model &  \multicolumn{2}{c}{D = 2} &  \multicolumn{2}{c}{D = 5} &  \multicolumn{2}{c}{D = 10} \\
				\hline
				\multirow{ 4 }{*} {Noise  0.1 }
				&  $\textbf{MGP}$ &  0.556  &  0.018 &  0.528  &  0.015 &  0.273  &  0.019 \\
				&  $\textbf{NIMGP}$ &  0.36  &  0.01 &  0.456  &  0.012 &  0.246  &  0.016 \\
				&  $\textbf{NIMGP}_\textbf{NN}$ &  0.336  &  0.01 &  0.439  &  0.0096 &  0.271  &  0.015 \\
				&  $\textbf{NIMGP}_\textbf{FO}$ &  0.369  &  0.012 &  0.437  &  0.011 &  0.244  &  0.016 \\
				\hline\multirow{ 4 }{*} {Noise  0.25 }
				&  $\textbf{MGP}$ &  0.825  &  0.026 &  0.674  &  0.018 &  0.358  &  0.022 \\
				&  $\textbf{NIMGP}$ &  0.505  &  0.012 &  0.521  &  0.014 &  0.309  &  0.016 \\
				&  $\textbf{NIMGP}_\textbf{NN}$ &  0.455  &  0.012 &  0.501  &  0.012 &  0.328  &  0.016 \\
				&  $\textbf{NIMGP}_\textbf{FO}$ &  0.519  &  0.015 &  0.508  &  0.013 &  0.3  &  0.018 \\
				\hline\multirow{ 4 }{*} {Noise  0.5 }
				&  $\textbf{MGP}$ &  1.07  &  0.033 &  0.828  &  0.026 &  0.419  &  0.026 \\
				&  $\textbf{NIMGP}$ &  0.62  &  0.013 &  0.583  &  0.016 &  0.348  &  0.019 \\
				&  $\textbf{NIMGP}_\textbf{NN}$ &  0.548  &  0.013 &  0.569  &  0.014 &  0.353  &  0.016 \\
				&  $\textbf{NIMGP}_\textbf{FO}$ &  0.655  &  0.017 &  0.565  &  0.016 &  0.323  &  0.02 \\
				\hline
			\end{tabular}
			\caption{N=100}
		\end{subtable}
		
		\begin{subtable}{\linewidth}
			\centering
			\begin{tabular}{l|c|r@{$\pm$}l|r@{$\pm$}l|r@{$\pm$}l}
				& Model &  \multicolumn{2}{c}{D = 2} &  \multicolumn{2}{c}{D = 5} &  \multicolumn{2}{c}{D = 10} \\
				\hline
				\multirow{ 4 }{*} {Noise  0.1 }
				&  $\textbf{MGP}$ &  0.763  &  0.025 &  0.545  &  0.016 &  0.261  &  0.019 \\
				&  $\textbf{NIMGP}$ &  0.27  &  0.0085 &  0.292  &  0.0077 &  0.154  &  0.011 \\
				&  $\textbf{NIMGP}_\textbf{NN}$ &  0.277  &  0.0089 &  0.291  &  0.0076 &  0.153  &  0.01 \\
				&  $\textbf{NIMGP}_\textbf{FO}$ &  0.332  &  0.011 &  0.313  &  0.0078 &  0.166  &  0.011 \\
				\hline\multirow{ 4 }{*} {Noise  0.25 }
				&  $\textbf{MGP}$ &  1.15  &  0.036 &  0.814  &  0.026 &  0.442  &  0.029 \\
				&  $\textbf{NIMGP}$ &  0.391  &  0.011 &  0.383  &  0.011 &  0.222  &  0.014 \\
				&  $\textbf{NIMGP}_\textbf{NN}$ &  0.398  &  0.012 &  0.384  &  0.011 &  0.218  &  0.013 \\
				&  $\textbf{NIMGP}_\textbf{FO}$ &  0.515  &  0.013 &  0.413  &  0.01 &  0.24  &  0.014 \\
				\hline\multirow{ 4 }{*} {Noise  0.5 }
				&  $\textbf{MGP}$ &  1.41  &  0.044 &  1.07  &  0.031 &  0.491  &  0.036 \\
				&  $\textbf{NIMGP}$ &  0.477  &  0.014 &  0.463  &  0.012 &  0.229  &  0.016 \\
				&  $\textbf{NIMGP}_\textbf{NN}$ &  0.5  &  0.016 &  0.468  &  0.012 &  0.235  &  0.015 \\
				&  $\textbf{NIMGP}_\textbf{FO}$ &  0.698  &  0.021 &  0.513  &  0.011 &  0.26  &  0.016 \\
				\hline
			\end{tabular}
			\caption{N=500}
		\end{subtable}
		
		\begin{subtable}{\linewidth}
			\centering
			\begin{tabular}{l|c|r@{$\pm$}l|r@{$\pm$}l|r@{$\pm$}l}
				& Model &  \multicolumn{2}{c}{D = 2} &  \multicolumn{2}{c}{D = 5} &  \multicolumn{2}{c}{D = 10} \\
				\hline
				\multirow{ 4 }{*} {Noise  0.1 }
				&  $\textbf{MGP}$ &  0.802  &  0.021 &  0.553  &  0.02 &  0.287  &  0.02 \\
				&  $\textbf{NIMGP}$ &  0.268  &  0.007 &  0.252  &  0.0083 &  0.141  &  0.009 \\
				&  $\textbf{NIMGP}_\textbf{NN}$ &  0.274  &  0.0077 &  0.255  &  0.008 &  0.139  &  0.0093 \\
				&  $\textbf{NIMGP}_\textbf{FO}$ &  0.34  &  0.009 &  0.275  &  0.0081 &  0.156  &  0.01 \\
				\hline\multirow{ 4 }{*} {Noise  0.25 }
				&  $\textbf{MGP}$ &  1.13  &  0.036 &  0.865  &  0.025 &  0.454  &  0.03 \\
				&  $\textbf{NIMGP}$ &  0.372  &  0.012 &  0.355  &  0.0095 &  0.197  &  0.012 \\
				&  $\textbf{NIMGP}_\textbf{NN}$ &  0.388  &  0.013 &  0.357  &  0.0092 &  0.2  &  0.012 \\
				&  $\textbf{NIMGP}_\textbf{FO}$ &  0.529  &  0.016 &  0.391  &  0.0097 &  0.217  &  0.013 \\
				\hline\multirow{ 4 }{*} {Noise  0.5 }
				&  $\textbf{MGP}$ &  1.49  &  0.042 &  1.12  &  0.037 &  0.507  &  0.04 \\
				&  $\textbf{NIMGP}$ &  0.485  &  0.012 &  0.437  &  0.012 &  0.211  &  0.015 \\
				&  $\textbf{NIMGP}_\textbf{NN}$ &  0.507  &  0.013 &  0.453  &  0.012 &  0.215  &  0.014 \\
				&  $\textbf{NIMGP}_\textbf{FO}$ &  0.76  &  0.021 &  0.507  &  0.011 &  0.238  &  0.016 \\
				\hline
			\end{tabular}
			\caption{N=1000}
		\end{subtable}
		\caption{Average negative test log-likelihood and standard deviations over 100 splits on synthetic data sets varying the number of data points and dimensions.}
		\label{tab:extra_nll}
	\end{table}
	
	\begin{table}[H]
		\scriptsize
		\begin{subtable}{\linewidth}
			\centering
			\begin{tabular}{l|c|r@{$\pm$}l|r@{$\pm$}l|r@{$\pm$}l}
				& Model &  \multicolumn{2}{c}{D = 2} &  \multicolumn{2}{c}{D = 5} &  \multicolumn{2}{c}{D = 10} \\
				\hline
				\multirow{ 4 }{*} {Noise  0.1 }
				&  $\textbf{MGP}$ &  0.146  &  0.0047 &  0.179  &  0.0047 &  0.0965  &  0.0069 \\
				&  $\textbf{NIMGP}$ &  0.144  &  0.0044 &  0.176  &  0.005 &  0.0941  &  0.0067 \\
				&  $\textbf{NIMGP}_\textbf{NN}$ &  0.134  &  0.0044 &  0.173  &  0.0041 &  0.0997  &  0.0069 \\
				&  $\textbf{NIMGP}_\textbf{FO}$ &  0.14  &  0.0046 &  0.171  &  0.0045 &  0.0954  &  0.0072 \\
				\hline\multirow{ 4 }{*} {Noise  0.25 }
				&  $\textbf{MGP}$ &  0.196  &  0.0059 &  0.212  &  0.0061 &  0.119  &  0.0078 \\
				&  $\textbf{NIMGP}$ &  0.205  &  0.0065 &  0.208  &  0.006 &  0.118  &  0.0072 \\
				&  $\textbf{NIMGP}_\textbf{NN}$ &  0.186  &  0.0054 &  0.199  &  0.0056 &  0.122  &  0.0069 \\
				&  $\textbf{NIMGP}_\textbf{FO}$ &  0.188  &  0.0055 &  0.201  &  0.0061 &  0.117  &  0.0075 \\
				\hline\multirow{ 4 }{*} {Noise  0.5 }
				&  $\textbf{MGP}$ &  0.242  &  0.0072 &  0.243  &  0.0079 &  0.133  &  0.0092 \\
				&  $\textbf{NIMGP}$ &  0.251  &  0.0073 &  0.241  &  0.0083 &  0.133  &  0.0092 \\
				&  $\textbf{NIMGP}_\textbf{NN}$ &  0.228  &  0.0066 &  0.23  &  0.007 &  0.129  &  0.0085 \\
				&  $\textbf{NIMGP}_\textbf{FO}$ &  0.229  &  0.0071 &  0.223  &  0.0074 &  0.127  &  0.009 \\
				\hline
			\end{tabular}
			\caption{N=100}
		\end{subtable}
		
		\begin{subtable}{\linewidth}
			\centering
			\begin{tabular}{l|c|r@{$\pm$}l|r@{$\pm$}l|r@{$\pm$}l}
				& Model &  \multicolumn{2}{c}{D = 2} &  \multicolumn{2}{c}{D = 5} &  \multicolumn{2}{c}{D = 10} \\
				\hline
				\multirow{ 4 }{*} {Noise  0.1 }
				&  $\textbf{MGP}$ &  0.124  &  0.0042 &  0.129  &  0.0036 &  0.0663  &  0.0053 \\
				&  $\textbf{NIMGP}$ &  0.117  &  0.0039 &  0.117  &  0.0031 &  0.0598  &  0.0044 \\
				&  $\textbf{NIMGP}_\textbf{NN}$ &  0.117  &  0.0037 &  0.117  &  0.0031 &  0.0587  &  0.0042 \\
				&  $\textbf{NIMGP}_\textbf{FO}$ &  0.118  &  0.0039 &  0.121  &  0.0031 &  0.0642  &  0.005 \\
				\hline\multirow{ 4 }{*} {Noise  0.25 }
				&  $\textbf{MGP}$ &  0.177  &  0.0054 &  0.169  &  0.0051 &  0.099  &  0.0065 \\
				&  $\textbf{NIMGP}$ &  0.168  &  0.005 &  0.154  &  0.0049 &  0.0903  &  0.0059 \\
				&  $\textbf{NIMGP}_\textbf{NN}$ &  0.167  &  0.005 &  0.153  &  0.005 &  0.0883  &  0.0056 \\
				&  $\textbf{NIMGP}_\textbf{FO}$ &  0.169  &  0.0051 &  0.158  &  0.0051 &  0.0934  &  0.0061 \\
				\hline\multirow{ 4 }{*} {Noise  0.5 }
				&  $\textbf{MGP}$ &  0.216  &  0.007 &  0.209  &  0.0059 &  0.102  &  0.0078 \\
				&  $\textbf{NIMGP}$ &  0.204  &  0.0068 &  0.189  &  0.0054 &  0.0914  &  0.0069 \\
				&  $\textbf{NIMGP}_\textbf{NN}$ &  0.205  &  0.0069 &  0.187  &  0.0052 &  0.0902  &  0.0071 \\
				&  $\textbf{NIMGP}_\textbf{FO}$ &  0.204  &  0.0068 &  0.19  &  0.0054 &  0.0945  &  0.0072 \\
				\hline
			\end{tabular}
			\caption{N=500}
		\end{subtable}
		
		\begin{subtable}{\linewidth}
			\centering
			\begin{tabular}{l|c|r@{$\pm$}l|r@{$\pm$}l|r@{$\pm$}l}
				& Model &  \multicolumn{2}{c}{D = 2} &  \multicolumn{2}{c}{D = 5} &  \multicolumn{2}{c}{D = 10} \\
				\hline
				\multirow{ 4 }{*} {Noise  0.1 }
				&  $\textbf{MGP}$ &  0.122  &  0.0032 &  0.115  &  0.0038 &  0.0635  &  0.0043 \\
				&  $\textbf{NIMGP}$ &  0.115  &  0.003 &  0.102  &  0.0035 &  0.0567  &  0.0038 \\
				&  $\textbf{NIMGP}_\textbf{NN}$ &  0.115  &  0.0031 &  0.102  &  0.0033 &  0.056  &  0.0039 \\
				&  $\textbf{NIMGP}_\textbf{FO}$ &  0.116  &  0.0029 &  0.107  &  0.0038 &  0.0607  &  0.004 \\
				\hline\multirow{ 4 }{*} {Noise  0.25 }
				&  $\textbf{MGP}$ &  0.167  &  0.0055 &  0.161  &  0.0044 &  0.0897  &  0.0058 \\
				&  $\textbf{NIMGP}$ &  0.16  &  0.0052 &  0.145  &  0.0043 &  0.0812  &  0.0053 \\
				&  $\textbf{NIMGP}_\textbf{NN}$ &  0.16  &  0.0049 &  0.145  &  0.0041 &  0.0811  &  0.0052 \\
				&  $\textbf{NIMGP}_\textbf{FO}$ &  0.16  &  0.005 &  0.148  &  0.0043 &  0.0844  &  0.0054 \\
				\hline\multirow{ 4 }{*} {Noise  0.5 }
				&  $\textbf{MGP}$ &  0.213  &  0.0058 &  0.193  &  0.0059 &  0.0929  &  0.0074 \\
				&  $\textbf{NIMGP}$ &  0.204  &  0.0057 &  0.178  &  0.0054 &  0.0852  &  0.0067 \\
				&  $\textbf{NIMGP}_\textbf{NN}$ &  0.206  &  0.0058 &  0.181  &  0.0054 &  0.0844  &  0.0066 \\
				&  $\textbf{NIMGP}_\textbf{FO}$ &  0.205  &  0.0055 &  0.179  &  0.0056 &  0.0865  &  0.0064 \\
				\hline
			\end{tabular}
			\caption{N=1000}
		\end{subtable}
		\caption{Average test error and standard deviations over 100 splits on synthetic data sets varying the number of data points and dimensions.}
		\label{tab:extra_err}
	\end{table}

	\subsection{Synthetic Experiments when Varying Number of Classes}
	
	We repeat the experiments of Section \ref{sec:synt_exp} using a different number of classes.
	In these experiments we assume the variance of the input noise is given. The results obtained 
	are similar to the ones reported in that section. In general, when we increase the number of
	class labels the prediction performance is reduced, since the classification problem is more difficult.

	\begin{table}[H]
		\footnotesize
		\centering
		\begin{tabular}{l|c|r@{$\pm$}l|r@{$\pm$}l|r@{$\pm$}l|r@{$\pm$}l}
			& Model &  \multicolumn{2}{c}{C = 4} &  \multicolumn{2}{c}{C = 5} &  \multicolumn{2}{c}{C = 6} &  \multicolumn{2}{c}{C = 7} \\
			\hline
			\multirow{ 4 }{*} {Noise  0.1 }
			&  $\textbf{MGP}$ &  1.11  &  0.022 &  1.28  &  0.021 &  1.45  &  0.02 &  1.58  &  0.02 \\
			&  $\textbf{NIMGP}$ &  0.361  &  0.0064 &  0.419  &  0.0075 &  0.47  &  0.0067 &  0.512  &  0.006 \\
			&  $\textbf{NIMGP}_\textbf{NN}$ &  0.37  &  0.0069 &  0.424  &  0.007 &  0.482  &  0.0074 &  0.527  &  0.0064 \\
			&  $\textbf{NIMGP}_\textbf{FO}$ &  0.45  &  0.0093 &  0.512  &  0.0097 &  0.583  &  0.0098 &  0.636  &  0.0095 \\
			\hline\multirow{ 4 }{*} {Noise  0.25 }
			&  $\textbf{MGP}$ &  1.63  &  0.027 &  1.92  &  0.029 &  2.13  &  0.028 &  2.34  &  0.027 \\
			&  $\textbf{NIMGP}$ &  0.529  &  0.0087 &  0.612  &  0.0081 &  0.688  &  0.0087 &  0.748  &  0.0088 \\
			&  $\textbf{NIMGP}_\textbf{NN}$ &  0.553  &  0.01 &  0.637  &  0.011 &  0.72  &  0.01 &  0.792  &  0.011 \\
			&  $\textbf{NIMGP}_\textbf{FO}$ &  0.757  &  0.014 &  0.882  &  0.015 &  0.976  &  0.015 &  1.06  &  0.015 \\
			\hline\multirow{ 4 }{*} {Noise  0.5 }
			&  $\textbf{MGP}$ &  2.13  &  0.035 &  2.47  &  0.04 &  2.77  &  0.031 &  3.06  &  0.035 \\
			&  $\textbf{NIMGP}$ &  0.685  &  0.011 &  0.787  &  0.01 &  0.891  &  0.009 &  0.991  &  0.01 \\
			&  $\textbf{NIMGP}_\textbf{NN}$ &  0.735  &  0.015 &  0.842  &  0.015 &  0.94  &  0.013 &  1.07  &  0.016 \\
			&  $\textbf{NIMGP}_\textbf{FO}$ &  1.08  &  0.019 &  1.25  &  0.02 &  1.37  &  0.02 &  1.54  &  0.017 \\
			\hline
		\end{tabular}
		\caption{Average negative test log-likelihood error and standard deviations over 100 splits on synthetic data sets varying the number of classes.}
		\label{tab:extra_classes_nll}
	\end{table}
	
	\begin{table}[H]
		\footnotesize
		\centering
		\begin{tabular}{l|c|r@{$\pm$}l|r@{$\pm$}l|r@{$\pm$}l|r@{$\pm$}l}
			& Model &  \multicolumn{2}{c}{C = 4} &  \multicolumn{2}{c}{C = 5} &  \multicolumn{2}{c}{C = 6} &  \multicolumn{2}{c}{C = 7} \\
			\hline
			\multirow{ 4 }{*} {Noise  0.1 }
			&  $\textbf{MGP}$ &  0.161  &  0.0029 &  0.183  &  0.0031 &  0.203  &  0.0029 &  0.221  &  0.0026 \\
			&  $\textbf{NIMGP}$ &  0.152  &  0.003 &  0.174  &  0.0029 &  0.192  &  0.0029 &  0.21  &  0.0028 \\
			&  $\textbf{NIMGP}_\textbf{NN}$ &  0.153  &  0.0028 &  0.174  &  0.0028 &  0.192  &  0.0029 &  0.211  &  0.0027 \\
			&  $\textbf{NIMGP}_\textbf{FO}$ &  0.153  &  0.003 &  0.175  &  0.0028 &  0.195  &  0.0028 &  0.213  &  0.0026 \\
			\hline\multirow{ 4 }{*} {Noise  0.25 }
			&  $\textbf{MGP}$ &  0.228  &  0.004 &  0.261  &  0.0037 &  0.286  &  0.0038 &  0.308  &  0.0034 \\
			&  $\textbf{NIMGP}$ &  0.219  &  0.0039 &  0.251  &  0.0036 &  0.277  &  0.0038 &  0.297  &  0.0033 \\
			&  $\textbf{NIMGP}_\textbf{NN}$ &  0.221  &  0.004 &  0.254  &  0.0039 &  0.278  &  0.0037 &  0.299  &  0.0035 \\
			&  $\textbf{NIMGP}_\textbf{FO}$ &  0.221  &  0.0038 &  0.252  &  0.0036 &  0.279  &  0.0038 &  0.299  &  0.0035 \\
			\hline\multirow{ 4 }{*} {Noise  0.5 }
			&  $\textbf{MGP}$ &  0.293  &  0.005 &  0.33  &  0.0052 &  0.366  &  0.0044 &  0.397  &  0.0046 \\
			&  $\textbf{NIMGP}$ &  0.282  &  0.0047 &  0.32  &  0.0048 &  0.355  &  0.0043 &  0.387  &  0.0045 \\
			&  $\textbf{NIMGP}_\textbf{NN}$ &  0.284  &  0.0052 &  0.322  &  0.0052 &  0.354  &  0.0046 &  0.384  &  0.0044 \\
			&  $\textbf{NIMGP}_\textbf{FO}$ &  0.281  &  0.0047 &  0.318  &  0.0049 &  0.354  &  0.0041 &  0.385  &  0.0044 \\
			\hline
		\end{tabular}
		\caption{Average test error and standard deviations over 100 splits on synthetic data sets varying the number of classes.}
		\label{tab:extra_classes_err}
	\end{table}

	\subsection{Astrophysics data set without Detection and Curvature Significance}
	\label{sec:appendix_b_extra_astrophisics}
	
	This section reports the results obtained on the Astrophysics data set without using the two attributes described.
	The results obtained are slightly worse than the ones reported in Section \ref{sec:astrophysics} which indicates
	that these attributes contain useful information.

	\begin{table}[H]
		\begin{center}
			\centering
			\begin{tabular}{l|r@{$\pm$}lr@{$\pm$}lr@{$\pm$}lr@{$\pm$}l}
				\hline
				& \multicolumn{2}{c}{\bf MGP}  
				& \multicolumn{2}{c}{\bf NIMGP}  
				& \multicolumn{2}{c}{\bf $\text{NIMGP}_\text{NN}$} 
				& \multicolumn{2}{c}{\bf $\text{NIMGP}_\text{FO}$} \\
				\hline
				NLL & 0.381 & 0.021 & \bf{0.268} & \bf{0.013} & 0.28 & 0.012 & 0.309 & 0.015 \\
				Test error & \bf{0.077} & \bf{0.0039} & 0.0811 & 0.0043 & 0.0885 & 0.0047 & 0.0867 & 0.0043 \\
				\hline
			\end{tabular}
			\caption{Average test error and neg. test log-likelihood (NLL) of each method on the astrophysics data set without using detection and curvature significance attributes.}
			\label{tab:astro-data_less}
		\end{center}
	\end{table}

	\begin{table}[H]
		\begin{center}
			\centering
			\begin{tabular}{c|c||r@{$\pm$}l|r@{$\pm$}l|r@{$\pm$}l}
				& Model & \multicolumn{2}{c}{\tt bll} (true) & \multicolumn{2}{c}{\tt fsrq} (true) & \multicolumn{2}{c}{\tt psr} (true) \\
				\hline
				\multirow{4}{*}{{\tt bll} (pred.)}
				& {\bf MGP} & 17.52  &  0.32 &  1.2  &  0.11 &  0.07  &  0.026 \\
				& {\bf NIMGP} & 17.59  &  0.33 &  1.45  &  0.13 &  0.02  &  0.014 \\
				& $\textbf{NIMGP}_\textbf{NN}$ & 17.43  &  0.31 &  1.55  &  0.13 &  0.06  &  0.028 \\
				& $\textbf{NIMGP}_\textbf{FO}$ & 17.65  &  0.32 &  1.87  &  0.16 &  0.02  &  0.014 \\
				\hline
				\multirow{4}{*}{{\tt fsrq} (pred.)}
				& {\bf MGP} & 1.59  &  0.12 &  15.26  &  0.29 &  0.37  &  0.058 \\
				& {\bf NIMGP} & 1.52  &  0.12 &  15.08  &  0.29 &  0.49  &  0.07 \\
				& $\textbf{NIMGP}_\textbf{NN}$ & 1.65  &  0.14 &  14.78  &  0.28 &  0.43  &  0.071\\
				& $\textbf{NIMGP}_\textbf{FO}$ & 1.49  &  0.11 &  14.63  &  0.29 &  0.37  &  0.06 \\
				\hline
				\multirow{4}{*}{{\tt psr} (pred.)}
				& {\bf MGP} & 0.03  &  0.017 &  0.3  &  0.054 &  9.66  &  0.27 \\
				& {\bf NIMGP} & 0.03  &  0.017 &  0.23  &  0.049 &  9.59  &  0.27 \\
				& $\textbf{NIMGP}_\textbf{NN}$  & 0.06  &  0.024 &  0.43  &  0.07 &  9.61  &  0.28 \\
				& $\textbf{NIMGP}_\textbf{FO}$  & 0.00  &  0.00 &  0.26  &  0.05 &  9.71  &  0.28 \\
				\hline
				
			\end{tabular}
			\caption{Average confusion matrix for the astrophysics data set for the different methods considered in this work without using detection and curvature significance attributes.}
			\label{tab:confusion-matrix_astro_less}
		\end{center}
	\end{table}
	
	\section{Summary of the Parameters of Each Method}
	\label{sec:appendix_b_params}
	
	We provide a table where we show the parameters used by each of the described methods. We also report the number
	of parameters employed. All methods share the typical GP hyper-parameters, including the length-scales, amplitudes,
	noise and inducing points locations. Since this parameters are the same for each method they are not included here.
	Similarly, all methods share the parameters for the variational distribution $q(\mathbf{u}^c)$ modeling the
	process values at the inducing points for class $c$. These parameters are not included here either.
	NIMGP, $\text{NIMGP}_\text{NN}$ and $\text{NIMGP}_\text{FO}$ share the parameters of the variance of the input noise
	$\mathbf{V}_i$ associated to each training instance. Let $N$ and $d$ be respectively the number of training
	points and the dimensionality of each method. Last, let $M$ the note the number of parameters (weights and biases) 
	of the neural network considered in $\text{NIMGP}_\text{NN}$. 
	
	\begin{table}[H]
		\centering
		\begin{tabular}{lcccc}
			\hline 
			&
			{\bf MGP }&
			{\bf  NIMGP }&
			{\bf $\text{NIMGP}_\text{NN}$ }&
			{\bf $\text{NIMGP}_\text{FO}$ }\\
			\hline
			Parameters & None & $\boldsymbol{\mu}_i^x, \textbf{V}_i^x$, $\mathbf{V}_i$ & $\theta$, $\mathbf{V}_i$ & $\mathbf{V}_i$ \\
			\# Param. & None & $N \times(d+d+d)$ & $M+N\times d$ & $N\times d$ \\
			\hline
		\end{tabular}
		\caption{Parameters and number of parameters of each method}
		\label{tab:parameters}
	\end{table}
	
	Recall that $\boldsymbol{\mu}_i^x$ and $\textbf{V}_i^x$ are the variational parameters modeling the posterior distribution
	(means and variances, respectively) of the attributes of each noiseless data instance. Similarly, $\theta$ is the set of parameters of the 
	neural network that predicts $\boldsymbol{\mu}_i^x$ and $\textbf{V}_i^x$ in $\text{NIMGP}_\text{NN}$. Finally, 
	$\mathbf{V}_i$ are the input noise variance parameters associated to the $i$-th data instance. Depending on the 
	number data instances, the methods with the largest number of parameters are either NIMGP or $\text{NIMGP}_\text{NN}$. 
	In large data sets, however, it is expected that $M \ll N$. 
	
	\bibliography{references}
\end{document}